\documentclass[journal]{IEEEtran}

\usepackage{times}
\usepackage{epsfig}
\usepackage{graphicx}
\usepackage{amsmath}
\usepackage{amssymb}
\usepackage{enumitem}
\usepackage{tabu}
\usepackage{booktabs}
\usepackage{multirow}
\usepackage{array}
\usepackage{adjustbox}
\usepackage{xcolor}
\usepackage{cite}
\usepackage{url}
\usepackage{pifont}
 \usepackage[implicit=false]{hyperref}

\definecolor{XH_color}{rgb}{1, 0.45, 0}

\begin{document}
%
\title{PSCC-Net: Progressive Spatio-Channel Correlation Network for Image Manipulation Detection and Localization}
%
%
%

\author{Xiaohong~Liu,~
Yaojie Liu,~
Jun~Chen,~\IEEEmembership{Senior Member,~IEEE},~
Xiaoming Liu,~\IEEEmembership{Senior Member,~IEEE}
\thanks{X. Liu is with the John Hopcroft Center, Shanghai Jiao Tong University, Shanghai, 200240, China (e-mail: xiaohongliu@sjtu.edu.cn).}
\thanks{J. Chen is with the Department of Electrical and Computer Engineering, McMaster University, Hamilton, ON L8S 4K1, Canada (e-mail:  chenjun@mcmaster.ca). }	
\thanks{Y. Liu and X. Liu are with the Department of Computer Science and Engineering, Michigan State University, East Lansing, MI 48824, USA (e-mail: \{liuyaoj1, liuxm\}@msu.edu). }
\thanks{Most of the work are conducted when Xiaohong Liu was a visiting scholar at MSU. This material is based upon work partially supported by the Defense Advanced Research Projects Agency (DARPA) under Agreement No.~HR00112090131.}}
\maketitle

\begin{abstract}
To defend against manipulation of image content, such as splicing, copy-move, and removal, we develop a Progressive Spatio-Channel Correlation Network (PSCC-Net) to detect and localize image manipulations. 
PSCC-Net processes the image in a two-path procedure: a top-down path that extracts local and global features and a bottom-up path that detects whether the input image is manipulated, and estimates its manipulation masks at multiple scales, where each mask is conditioned on the previous one. Different from the conventional encoder-decoder and no-pooling structures, PSCC-Net leverages features at different scales with dense cross-connections to produce manipulation masks in a coarse-to-fine fashion.
Moreover, a Spatio-Channel Correlation Module (SCCM) captures both spatial and channel-wise correlations in the bottom-up path, which endows features with holistic cues, enabling the network to cope with a wide range of manipulation attacks. Thanks to the light-weight backbone and progressive mechanism, PSCC-Net can process $1,080$P images at $50+$ FPS. Extensive experiments demonstrate the superiority of PSCC-Net over the state-of-the-art methods on both detection and localization. Codes and models are available at \url{https://github.com/proteus1991/PSCC-Net}.

\end{abstract}



\section{Introduction}

\textit{Seeing is believing?} Not anymore. Recent advances on image manipulation techniques~\cite{li2020manigan, lee2020maskgan, dhamo2020semantic, liu2020open} enable easy editing of raw images, such as removing unwanted objects~\cite{li2019progressive, li2020recurrent, lahiri2020prior, zeng2020high}, face swapping~\cite{lee2020maskgan}, attribute changing~\cite{shen2020interpreting}, \textit{etc}.
Although such techniques are neutral, malicious attackers may utilize them to create deceitful content to propagate false information, \textit{e.g.}, fake news~\cite{huh2018fighting}, insurance fraud~\cite{wu2019mantra}, and Deepfake~\cite{tolosana2020deepfakes, on-the-detection-of-digital-face-manipulation,yang2021msta,hu2021detecting}. 
Thus, concerns of the adverse impact on social media and even real-world systems have been raised~\cite{indaily, cnnnew}.
To alleviate the concerns, it is crucial to develop reliable models to expose the manipulated images. 
While being used in machine and systems, the model is required to, at a minimal, distinguish manipulated images from pristine ones, where the objective is to \textit{detect}.
While being used for human's viewing, the model is further required to estimate tampered areas in forged images, where the objective is to \textit{localize}.
\begin{figure}[t]
	\centering
	\begin{minipage}[h]{0.24\linewidth}
		\centering
		\includegraphics[width=\linewidth]{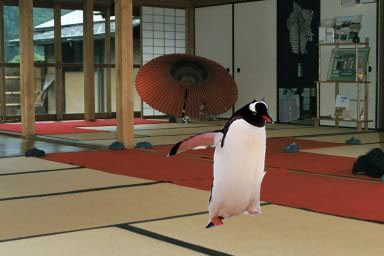}
	\end{minipage}
	\vspace{1mm}
	\begin{minipage}[h]{0.24\linewidth}
		\centering
		\includegraphics[width=\linewidth]{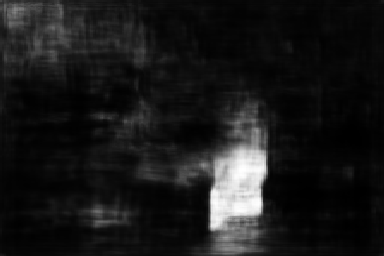}
	\end{minipage}
	\begin{minipage}[h]{0.24\linewidth}
		\centering
		\includegraphics[width=\linewidth]{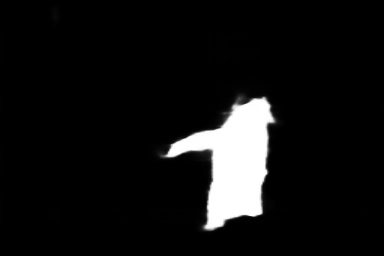}
	\end{minipage}
	\begin{minipage}[h]{0.24\linewidth}
		\centering
		\includegraphics[width=\linewidth]{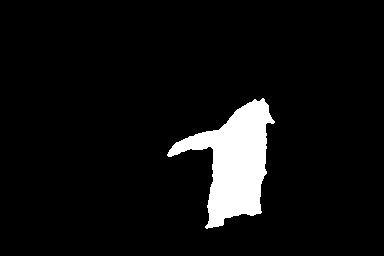}
	\end{minipage}
	\begin{minipage}[h]{0.24\linewidth}
		\centering
		\includegraphics[width=\linewidth]{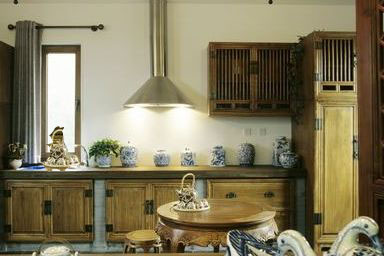}
	\end{minipage}
	\vspace{1mm}
	\begin{minipage}[h]{0.24\linewidth}
		\centering
		\includegraphics[width=\linewidth]{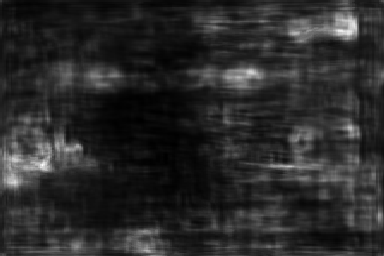}
	\end{minipage}
	\begin{minipage}[h]{0.24\linewidth}
		\centering
		\includegraphics[width=\linewidth]{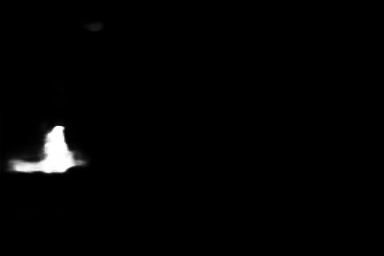}
	\end{minipage}
	\begin{minipage}[h]{0.24\linewidth}
		\centering
		\includegraphics[width=\linewidth]{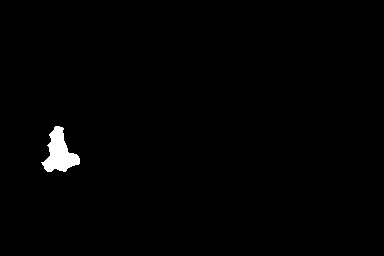}
	\end{minipage}
	\begin{minipage}[h]{0.24\linewidth}
		\centering
		\includegraphics[width=\linewidth]{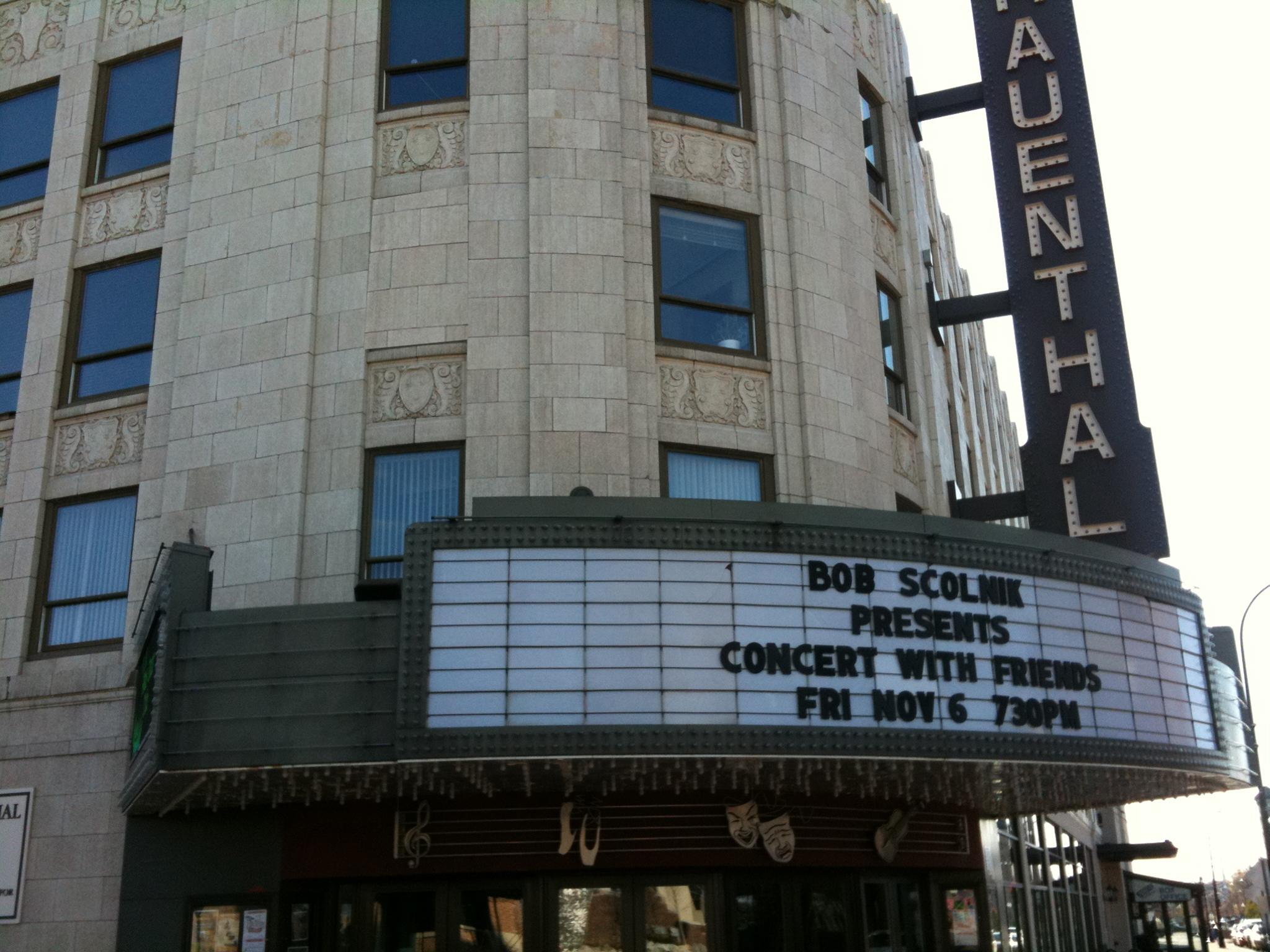}
		\scriptsize{(a) Manipulated}
	\end{minipage}
	\begin{minipage}[h]{0.24\linewidth}
		\centering
		\includegraphics[width=\linewidth]{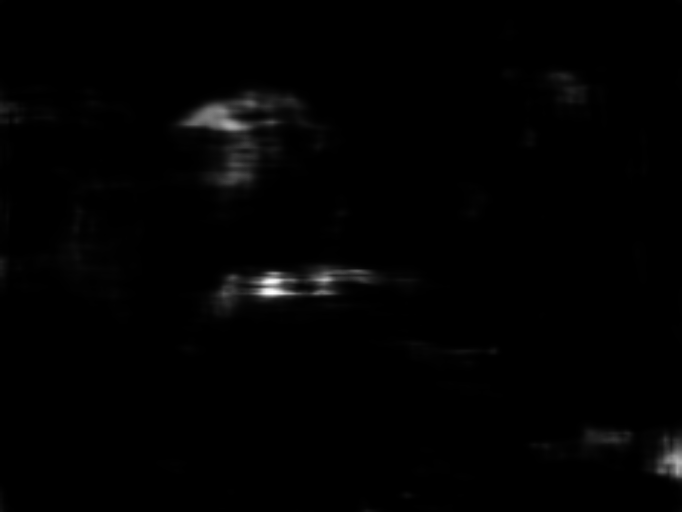}
		\scriptsize{(b) SPAN~\cite{hu2020span}}
	\end{minipage}
	\begin{minipage}[h]{0.24\linewidth}
		\centering
		\includegraphics[width=\linewidth]{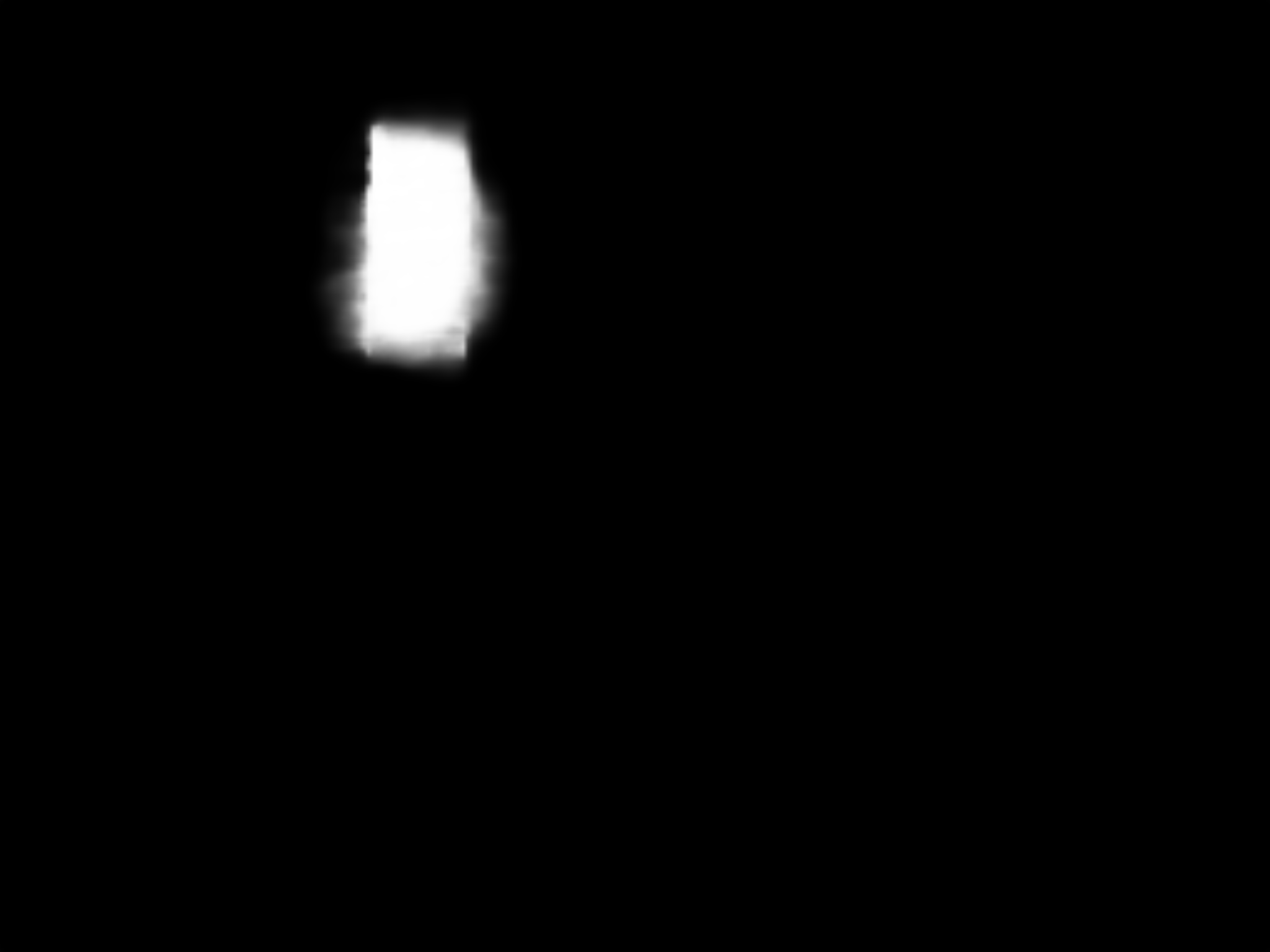}
		\scriptsize{(c) PSCC-Net}
	\end{minipage}
	\begin{minipage}[h]{0.24\linewidth}
		\centering
		\includegraphics[width=\linewidth]{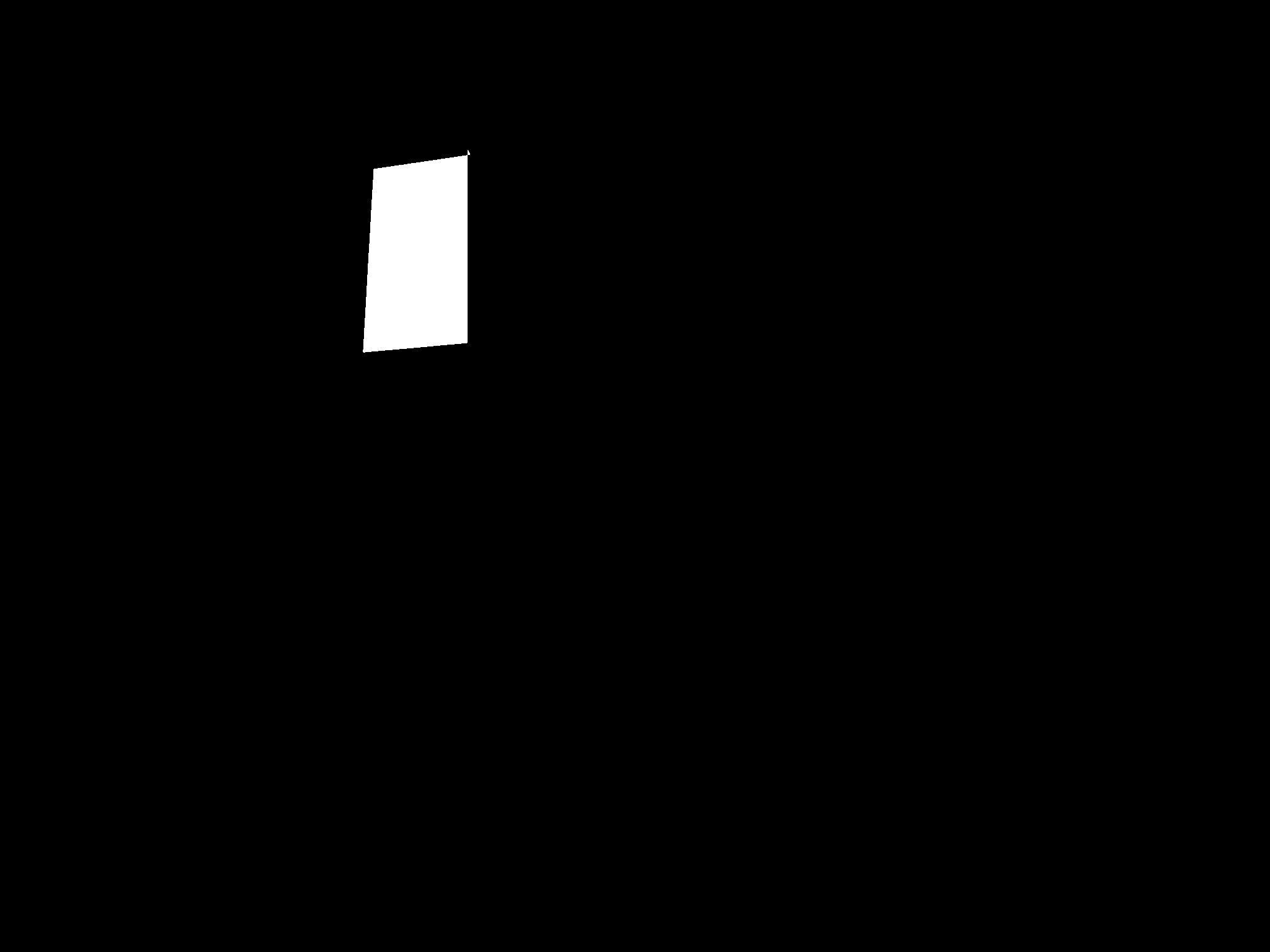}
		\scriptsize{(d) GT}
	\end{minipage}
	\caption{Examples of image manipulation localization. Three examples are splicing, copy-move, and removal manipulations respectively. With novel designs of progressive mechanism and correlation module, our method demonstrates robust and accurate estimation at different scales and types.}
	\label{fig:firstpage}
\end{figure}

Generally, image manipulation consists of the content-dependent process and content-independent process. 
The former includes splicing, copy-move, and removal, as shown in Fig.~\ref{fig:firstpage}. 
Both splicing and copy-move are content-copying forgeries, where the splicing content is from a different donor image while the copy-move content is from the target image \textit{per se}. 
Removal takes out certain objects from the target image and performs refilling via inpainting.
Often, the content-dependent process follows the semantic arrangement in the target image, \textit{e.g.}, placing a car on the road and replacing one face with another, which makes the resulting image visually ``authentic" and indistinguishable from the pristine one.
However, based on image/camera trace analysis~\cite{cozzolino2019noiseprint, chen2020camera}, subtle patterns can still be revealed to indicate the manipulation.
On the other hand, the content-independent process includes global modifications such as brightness/contrast change, blurring, noising and image compression. 
They barely create any disinformation, but their resultant noise may undermine the analysis of image/camera traces and potentially hide the discrepancy between the manipulated and pristine areas.


To defend against manipulations, many image manipulation detection and localization (IMDL) methods have been proposed in the past. 
In the early stages, methods are designed to handle a single type of manipulation.
In recent years, works~\cite{bappy2017exploiting, salloum2018image, zhou2018learning, bappy2019hybrid, wu2019mantra, hu2020span,on-the-detection-of-digital-face-manipulation, proactive-image-manipulation-detection,https://doi.org/10.48550/arxiv.2106.07873} are proposed to build generic IMDL models for {\it multiple} manipulation types.
However, there are still $3$ major unsolved problems for IMDL: 

\subsubsection{Scale variation} The forged area varies in sizes. Most prior works neglect the importance of scale variations and encounter difficulty when detecting forged areas of different sizes. Both the conventional encoder-decoder~\cite{zhou2018learning, bappy2019hybrid} and no-pooling~\cite{wu2019mantra, hu2020span} structures have difficulties in leveraging local and global features jointly, thus can only handle a limited scale variation.

\subsubsection{Image correlation}
Manipulated regions can best be determined when compared to pristine regions, especially for splicing attacks. A naive learning of mapping from the manipulated image to manipulation mask may lead to an overfitting to the specific attack type in training.
In contrast, considering the image spatial correlation can lead to a more generalized localization solution.
Yet, such correlation is mostly neglected in prior works.

\subsubsection{Detection} 
In principle, manipulation detection and localization are highly relevant tasks, where the detection score can be simply derived from the response of the predicted manipulation mask, {\it i.e.}, at least one part of the forged image has high response while no part of the pristine one does. 
However, most prior works assume the {\it existence of manipulation} in all input images. As a result, this could cause many false alarms on pristine images and make the detection unreliable.

\begin{figure*}[t]
	\centering
	\includegraphics[width=\linewidth]{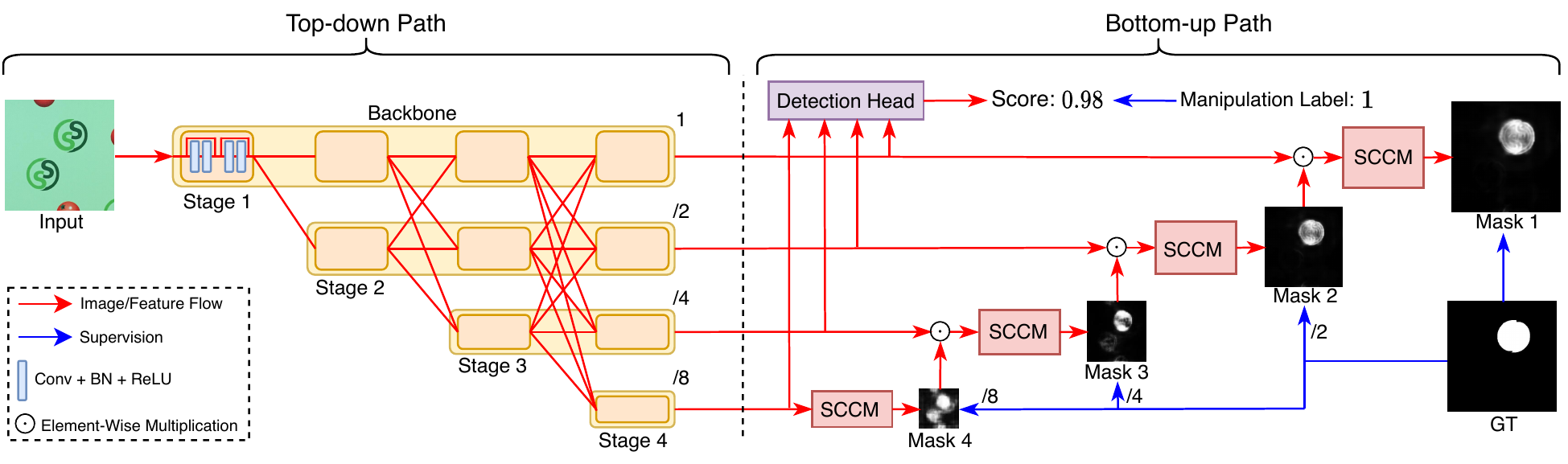}
	\caption{The architecture of the proposed PSCC-Net. The  detection score predicted by the detection head indicates if the input is manipulated or not. The accuracy of manipulation localization from \textit{Mask 4} to \textit{Mask 1} is gradually improved, {\it e.g.}, the prediction of \textit{Mask 4} confuses the pasted (forged) region with the pristine (copied) one, while \textit{Mask 1} effectively fixes it.}
	\label{fig:PDANet}
\end{figure*}

To address the above issues, we propose a novel Progressive Spatio-Channel Correlation Network (PSCC-Net), as in Fig.~\ref{fig:PDANet}. 
PSCC-Net consists of a top-down path and a bottom-up path.
In the top-down path, a backbone encoder first extracts the local and global features from an input image.
We adopt the network structure of~\cite{wang2020deep} as our encoder, whose dense connections among different scales facilitate information exchange.
In the bottom-up path, we leverage the learned features to estimate $4$ manipulation masks from small scales to large ones, where each mask serves as a prior in the next-scale estimation. Thanks to such a design, the final mask is estimated in a coarse-to-fine fashion, harvesting both the local and global information. This design enables a potential speed-up by terminating the bottom-up mask estimation, if the intermediate mask is satisfactory.
Moreover, rather than investigating the response of predicted manipulation masks, we feed the learned features into a detection head to produce the score for binary classification.

To exploit image correlation, we propose a Spatio-Channel Correlation Module (SCCM) that grasps both spatial and channel-wise correlations at each bottom-up step.
The spatial correlation aggregates the global context among local features. As the response from different channels might be associated with the same class (\textit{e.g.}, manipulated or pristine), the channel-wise correlation computes the similarity among feature maps to enhance the representation in interest areas.
Given the light-weight design of the encoder, PSCC-Net can process $1,080$P at $50+$ FPS.
Our proposed approach demonstrates a superior manipulation localization on several benchmarks. In addition, we show that the recent IMDL methods encounter difficulty in distinguishing manipulated images from pristine ones. By explicitly introducing a detection head, our method achieves the state of the art (SOTA) on manipulation detection.

We summarize the contributions of this work as follows:

\begin{itemize}
\item A new PSCC-Net is proposed that performs favorably on manipulation detection and enables progressive improvement of  manipulation localization in a coarse-to-fine fashion;

\item A novel SCCM module is designed to capture the spatial and channel-wise correlations for better generalization. SCCM avoids the use of massive annotated data to pre-train our feature extractor;

\item The SOTA results for both image manipulation detection and localization are successively achieved.
\end{itemize}
\section{Related Work}
\subsection{Image Manipulation Detection} Image manipulation detection aims to distinguish manipulated images from pristine ones via image-level binary classification. There are two major approaches for this detection: the implicit manner~\cite{wu2018busternet, huh2018fighting} and the explicit manner~\cite{islam2020doa}. The former obtains the detection score by the statistics (\textit{e.g.}, average~\cite{huh2018fighting} or maximum~\cite{wu2018busternet} value) of the predicted manipulation mask, and the latter explicitly outputs the score from a dedicated classification module. Recent works~\cite{wu2019mantra, hu2020span} 
focus on pixel-level manipulation localization but neglect the importance of image-level detection. 
Instead, this work leverages both manipulated and pristine images in training and jointly considers detection and localization of image manipulation.

\subsection{Image Manipulation Localization} Early works propose to localize the manipulation of one specific type,  \textit{e.g.,} splicing~\cite{lyu2014exposing, cozzolino2015splicebuster, amerini2017localization, bondi2017tampering, wu2017deep, huh2018fighting, cozzolino2019noiseprint, kniaz2019point,zhang2021multi}, copy-move~\cite{cozzolino2015efficient, wen2016coverage, wu2018busternet, wu2018image, d2018patchmatch, islam2020doa}, removal~\cite{zhu2018deep,aloraini2020sequential,yang2020spatiotemporal,wu2021iid}, and the content-preserved process~\cite{bappy2019hybrid,joseph2015literature}.
Although most methods perform well on detecting that specific forgery type, they fall short in handling real-world cases, where usually the forgery type is unknown in advance and various types of forgery might be utilized in manipulation.
In the related problem of face anti-spoofing, researchers also study how to localize the facial pixels covered with various spoof mediums~\cite{on-disentangling-spoof-traces-for-generic-face-anti-spoofing}.


Recent works attempt to tackle multiple forgeries in one model.
J-LSTM~\cite{bappy2017exploiting} and H-LSTM~\cite{bappy2019hybrid} integrate the LSTM and CNN to capture the boundary-discriminative features.
However, due to the patch-based design, both methods are time-consuming, and the size of detectable regions is limited by the preset patch size.
RGB-N~\cite{zhou2018learning} adopts the steganalysis rich model~\cite{fridrich2012rich} and Faster R-CNN~\cite{ren2015faster}, but it can only provide bounding boxes instead of segmentation masks.
Later, ManTra-Net~\cite{wu2019mantra} learns features to distinguish $385$ known manipulation types and treats the problem as anomaly detection. To learn the distinguishable features, auxiliary labeled data, such as camera sensors, are used. 
SPAN~\cite{hu2020span} extends ManTra-Net to further model the spatial correlation via local self-attention blocks and pyramid propagation. 
However, as the correlation is only considered in the local region,  ManTra-Net and SPAN fail to take full advantage of the spatial correlation and consequently have limited generalizability.
In this work, our PSCC-Net utilizes a progressive mechanism to improve the multi-scale feature representation and SCCM modules to better explore spatial and channel-wise correlations.

\subsection{Progressive Mechanism} 
Progressive mechanism tackles a challenging task in a coarse-to-fine fashion. It has been widely adopted in many low-level and high-level vision tasks, such as denoising~\cite{ren2018gated,li2019progressive}, inpainting~\cite{yi2019progressive}, super-resolution~\cite{jiang2020multi,fsrnet-end-to-end-learning-face-super-resolution-with-facial-priors}, and object detection~\cite{zhang2018progressive,zhuprogressface,song2020progressive,pedestrian-detection-with-autoregressive-network-phases}. The pyramid structure is commonly utilized to build multi-scale features. 
In this work, we propose a densely connected pyramid structure that progressively refines the manipulation mask from small scales to large ones, where each predicted mask serves as a prior for the next-scale estimation.

\subsection{Attention Mechanism} The pioneer work~\cite{vaswani2017attention} proposes an attention mechanism to improve the feature representation with relatively low cost, which has been widely employed in various vision tasks~\cite{hu2018squeeze, wang2018non, liu2019griddehazenet, isobe2020video, islam2020doa, on-the-detection-of-digital-face-manipulation,mitigating-face-recognition-bias-via-group-adaptive-classifier}.
According to the applied domain, the attention mechanism can be divided into two types: spatial attention~\cite{wang2018non} and channel-wise attention~\cite{hu2018squeeze}. Recent works~\cite{park2018bam, woo2018cbam,fu2019dual} take the benefit of both types to further improve the representation capability of DNN. 
These methods adopt separate schemes to explore the spatial and channel-wise attentions and thus require additional efforts to fuse them. In addition, due to memory limit, they can only apply to high-level features where the spatial size is small. 
In this work, a unified SCCM jointly explores the image correlation and discrepancy in both spatial domain and feature channels on the same features. Besides, owing to the dimensional reduction design, SCCM is able to adapt both low-level and high-level features with arbitrary sizes.

\section{PSCC-Net}


Our PSCC-Net enables the detection and localization of various types of manipulations.
As compared to the image-level detection, the pixel-level localization is more difficult. Therefore, PSCC-Net pays special attention to tackling the localization problem. 
Indeed, since the features for detection and localization are jointly learned, improving the localization performance will naturally benefit detection.

\subsection{Network Architecture}\label{sec:PM}
\subsubsection{Top-Down Path}
Most prior works use the conventional encoder-decoder~\cite{zhou2018learning, bappy2019hybrid} and no-pooling structures~\cite{wu2019mantra, hu2020span} to extract features. Since forged areas have various sizes, it is important to fuse local and global features to handle the scale variation. However, both structures extract features in a sequential pipeline and neglect feature fusion among different scales, and thus can only handle a limited scale variation. 
To address this issue, we adopt a light-weight backbone in~\cite{wang2020deep}, named HRNetV2p-W18. Following its default setting, the stage down-scaling ratio $s$ is set to $2$, and there are totally $4$  stages. 

Compared to encoder-decoder and no-pooling structures, the benefits of our backbone are two-fold. 
First, features from different scales are computed in parallel. Hence, dense connections among different scales enable effective information exchange, which is beneficial for handling scale variations. 
Second, since the local and global feature fusion is performed for every scale, each feature contains sufficient information to predict a manipulation mask at the corresponding scale. 
Therefore, this backbone is in line with our progressive mechanism, where the prediction of each mask should rely on all local and global features to improve its accuracy. Indeed, except the predicted mask on the last scale, the others serve as a prior for the next-scale mask prediction.
After the top-down path, the manipulated features on $4$ scales are extracted. Then, we use the bottom-up path to perform manipulation detection and localization.

\subsubsection{Bottom-Up Path}
The bottom-up path in PSCC-Net estimates the detection score and the manipulation mask. Specifically, the detection score is predicted based on the extracted features from the top-down-path via a detection head~\cite{wang2020deep}, then the manipulation mask is generated through a  progressive mechanism with full supervision.
In particular, the coarse-to-fine progressive mechanism mimics how  human tackles complicated problems in daily life.

We denote the input image as $\mathbf{I}\in\mathbb{R}^{H\times W\times 3}$. The extracted features at $4$ scales are  $\mathbf{F}_1\in\mathbb{R}^{H\times W\times C}$, $\mathbf{F}_2\in\mathbb{R}^{H/s\times W/s\times sC}$, $\mathbf{F}_3\in\mathbb{R}^{H/s^2\times W/s^2\times s^2C}$ and $\mathbf{F}_4\in\mathbb{R}^{H/s^3\times W/s^3\times s^3C}$, and their corresponding masks are denoted as $\mathbf{M}_1\in\mathbb{R}^{H\times W}$, $\mathbf{M}_2\in\mathbb{R}^{H/s\times W/s}$, $\mathbf{M}_3\in\mathbb{R}^{H/s^2\times W/s^2}$ and $\mathbf{M}_4\in\mathbb{R}^{H/s^3\times W/s^3}$. Here $H$, $W$, and $C$ are the height, width, and channel number of the image/feature respectively.
Formally, we have
\begin{equation} \label{equ:PM}
\mathbf{M}_{n-1} = f_{n-1}(\tau(\mathbf{M}_n) \cdot \mathbf{F}_{n-1}), \quad n=2,3,4,
\end{equation}
where $f_n$ denotes the SCCM on the $n$th scale, and $\tau$ is the upsampling operation (\textit{e.g.}, the bilinear interpolation). Since $\mathbf{M}_4$ is the mask on the last scale, it can be directly expressed as $\mathbf{M}_4 = f_4(\mathbf{F}_4)$. For Scales $1$-$3$,  the feature on the current scale is associated with the upsampled mask from the previous scale for feature modulation. Then, the modulated feature is fed into SCCM to produce a manipulation mask.

To reduce the prediction difficulty, the proposed progressive mechanism avoids generating the mask at the finest scale directly. Instead, the mask on the coarsest scale is first predicted to locate the regions that are potentially forged based on current available information. The subsequent prediction on the finer scale can leverage the previous mask and pay more attention to those selected regions. This process continues until the generation of
the manipulation mask at the finest scale, which serves as the final prediction.
However, without explicit supervision on each scale, the intermediate masks might not follow the coarse-to-fine order. Therefore, full supervisions are applied on all scales to guide the  mask estimation.


\subsection{Spatio-Channel Correlation Module} 
\label{sec:DAB}

Attention mechanisms are commonly used to modulate learned features according to their relative significance. As the final manipulation mask is binary, the localization can be considered as a pixel-level binary classification. Ideally, we expect the learned features on forged regions are similar to each other but distinct from those in pristine regions. 
In this case, a fundamental clustering method may suffice to produce an effective mask. Therefore, to better tackle manipulation localization, we propose a SCCM that employs the spatial attention to aggregate the pixel-level features based on their contextual correlations, and the channel-wise attention to consolidate the feature maps based on their channel correlations.

\begin{figure}[t]
	\centering
	\includegraphics[width=\linewidth]{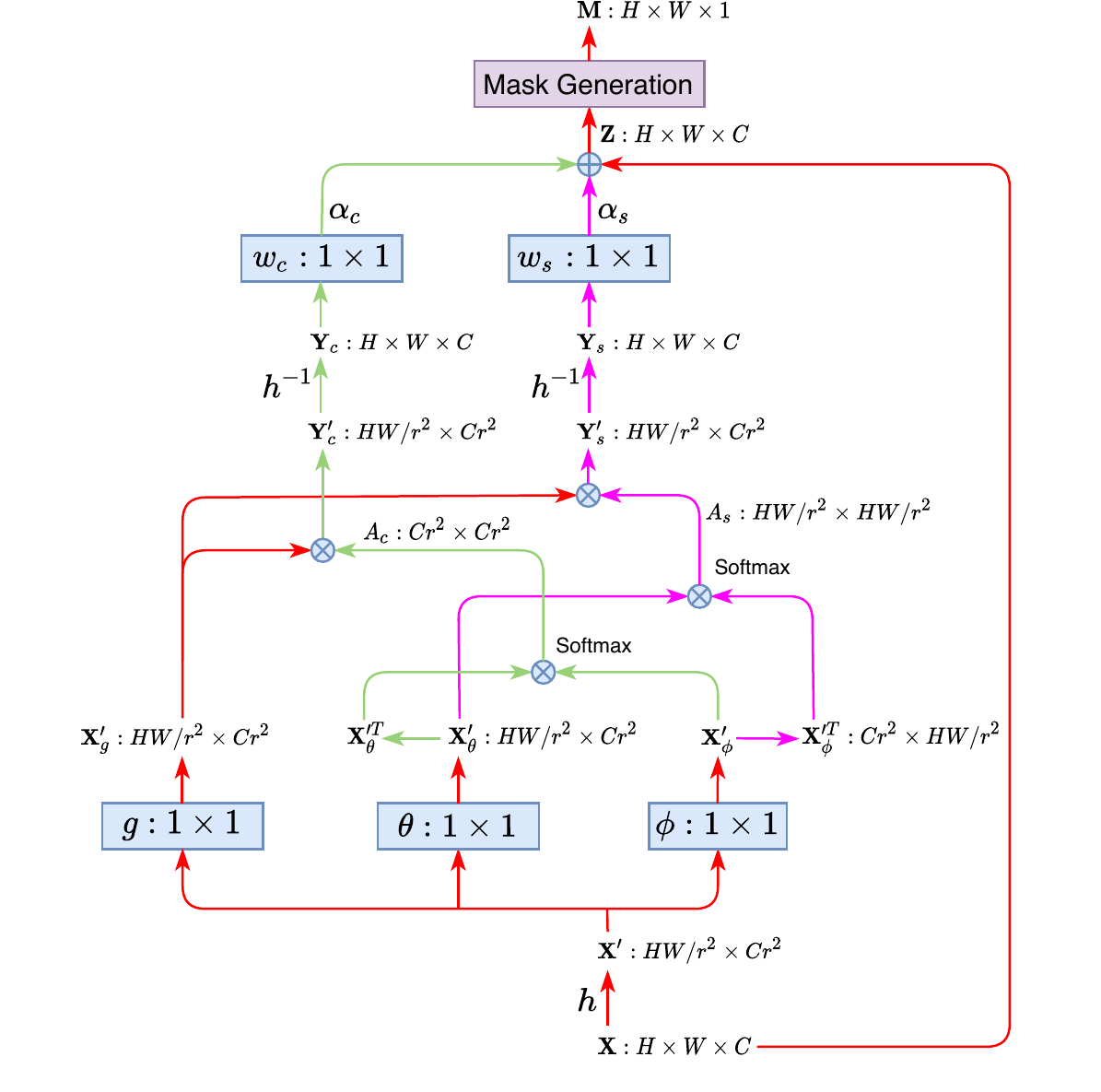}
	\caption{The structure of SCCM. Here $\otimes$ represents the matrix multiplication and $\oplus$ the element-wise addition; the red arrow shows the common feature flows; the pink  and green arrows show the feature flows of spatial and channel-wise attentions respectively.}
	\label{fig:DAB}
\end{figure}

We illustrate the detailed structure of SCCM in Fig.~\ref{fig:DAB}, where the input feature $\mathbf{X}$ is of size ${H\times W\times C}$. Note that even though $\mathbf{X}$ is small ($256 \times 256$), the size of its spatial correlation can be enormous ($65,536 \times 65,536$), easily exceeding the memory limit. Therefore, we use function $h$ to reshape the input $\mathbf{X}\in\mathbb{R}^{H \times W \times C}$ to $\mathbf{X}'\in\mathbb{R}^{HW/r^2 \times Cr^2}$, where each feature map is flattened to form a vector based on SCCM down-scaling ratio $r$. For instance, with $r=4$, the size of spatial correlation is $4,096 \times 4,096$ instead of $65,536 \times 65,536$.
Therefore, this operation preserves all feature information and avoids modeling the spatial correlation of potentially large size $HW \times HW$.

To build the spatial and channel-wise correlations, one may directly leverage $\mathbf{X}'$. However, additional flexibility could be achieved by introducing the embedded Gaussian function~\cite{wang2018non}. 
Therefore, we use the $1\times1$ convolution to build different functions $g$, $\theta$, and $\phi$ to transform  $\mathbf{X}'$ into new linear embeddings as $\mathbf{X}_g'=g(\mathbf{X}')$, $\mathbf{X_{\theta}'}=\theta(\mathbf{X}')$, and $\mathbf{X_{\phi}'}=\phi(\mathbf{X}')$, all with the same size as $\mathbf{X}'$. Subsequently, the spatial and channel-wise correlations (denoted as $\mathbf{A}_s\in\mathbb{R}^{HW/r^2\times HW/r^2}$ and $\mathbf{A}_c\in\mathbb{R}^{Cr^2\times Cr^2}$) of embedded features $\mathbf{X_{\theta}'}$ and $\mathbf{X_{\phi}'}$ are computed, and the Gaussian operation is implemented by Softmax function. In the end, the spatial and channel-wise attentions are realized by performing matrix multiplications  $\mathbf{A}_s\mathbf{X}_g'$ and $\mathbf{X}_g'\mathbf{A}_c$, respectively.
Unlike prior methods~\cite{park2018bam,woo2018cbam,fu2019dual} that employ two attentions on \textit{different} features, we apply both to the \textit{same} linear embedding for mutual accommodation. Indeed, applying attentions in this way reduces the difficulty of subsequent fusion process, and also saves computational operations in SCCM.
Specifically, the spatial attention can be formulated as:
\begin{equation} \label{equ:SA}
\mathbf{Y}_s' = \mathbf{A}_s\mathbf{X}_g' = \text{softmax}(\mathbf{X}'_\theta \mathbf{X}_\phi'^T)\mathbf{X}_g',
\end{equation}
where $\mathbf{Y}'_s\in\mathbb{R}^{HW/r^2 \times Cr^2}$ is the feature resulting from the application of spatial attention, and $\text{softmax}(\cdot)$ denotes the Softmax function. The element ($i, j$) in $\mathbf{A}_s$ indicates the similarity between the feature vectors in the $i$th row of $\mathbf{X}'_\theta$  and $j$th row of $\mathbf{X}'_\phi$. 
The more similar they are, the higher correlation they have. 
This helps the network to learn feature representations for distinguishing  forged regions from  pristine ones and avoid overfitting to a specific attack type in training. Similarly, the channel-wise attention is expressed as:
\begin{equation} \label{equ:CA}
\mathbf{Y}'_c = \mathbf{X}'_g\mathbf{A}_c = \mathbf{X}'_g\text{softmax}(\mathbf{X}_\theta '^T\mathbf{X}'_\phi),
\end{equation}
where $\mathbf{Y}'_c\in\mathbb{R}^{HW/r^2 \times Cr^2}$ is the feature resulting from the application of channel-wise attention. The element ($i, j$) in $\mathbf{A}_c$ measures the similarity between the channel maps in the $i$th column of $\mathbf{X}'_\theta$ and $j$th column of $\mathbf{X}'_\phi$. Since the response from different channels might be associated with the same class, {\it e.g.}, manipulated or pristine, the channel-wise correlation aggregates feature maps based on their similarities to enhance the representation in forged regions.

We use $h^{-1}$ to reshape $\mathbf{Y}'_s$ and $\mathbf{Y}'_c$ respectively back to  $\mathbf{Y}_s$ and $\mathbf{Y}_c$ of size  $H \times W \times C$. Further, two functions $\omega_s$ and $\omega_c$ are built by $1\times1$ convolution to improve their feature representations. The output features from $\omega_s$ and $\omega_c$ are complement to each other. As it is non-trivial to determine their relative significance, two learnable parameters $\alpha_s$ and $\alpha_c$, both initialized as $1$, are used for trade-off. The learned values of $\alpha_s$ and $\alpha_c$ can be found in supplementary. We also adopt the residual learning~\cite{he2016deep} to express the feature $\mathbf{Z}$ as:
\begin{equation} \label{equ:z}
\mathbf{Z} = \mathbf{X} + \alpha_s\cdot\omega_s(\mathbf{Y}_s) + \alpha_c\cdot\omega_c(\mathbf{Y}_c).
\end{equation}

The final output of SCCM is a predicted mask with only one channel. To reduce the channel number in $\mathbf{Z}$, we employ a mask generation block with the sequential order of \textit{Conv-ReLU-Conv-Sigmoid}, where \textit{Conv} is a $3\times3$ convolution. 

\subsection{Loss Function}
To train the PSCC-Net, we adopt the binary cross-entropy loss ($L_{bce}$) for both detection and localization tasks. The predicted detection score ($s_d$) is supervised by the ground-truth (GT) label ($l_d$) with $0$ standing for pristine image and $1$ for  forged image. Moreover, full supervisions are applied on each predicted mask by downsampling the GT mask $\mathbf{G}_1$ to $\mathbf{G}_2$, $\mathbf{G}_3$, and $\mathbf{G}_4$ according to their corresponding sizes, with $0$ standing for pristine pixel and $1$ for forged pixel. The masks predicted through the progressive mechanism at different scales  are considered to be of equal importance. Therefore, our final loss function $\hat{L}$ can be expressed as:
\begin{equation} \label{equ:loss}
\hat{L} = L_{bce}(s_d, l_d) + \frac{1}{4}\sum\nolimits_{m=1}^{4}L_{bce}(\mathbf{M}_m, \mathbf{G}_m).
\end{equation}


\subsection{Training Data Synthesis}
Since there is no standard IMDL dataset for training, a synthetic dataset is built to train and validate our PSCC-Net. This dataset includes four categories 1) splicing, 2) copy-move, 3) removal, and 4) pristine classes. For splicing, following~\cite{wu2017deep, liu2019adversarial}, we use the MS COCO~\cite{lin2014microsoft} to generate spliced images, where one annotated region is randomly selected per image, and pasted into a different image after several transformations. We adopt the same transformation as~\cite{wu2017deep} including the scale, rotation, shift and luminance changes. Since the spliced region is not necessarily an object, we use the Bezier curve~\cite{mortenson1999mathematics} to generate random contours, then fill them to produce splicing masks. We follow the same processes above but randomly select donor and target images in KCMI~\cite{KCMI}, VISION~\cite{shullani2017vision}, and Dresden~\cite{gloe2010dresden} that are commonly used to identify camera source~\cite{chen2020camera}, to generate additional spliced images as supplementary. For copy-move, the dataset from~\cite{wu2018busternet} is adopted. For removal, inspired by~\cite{wu2017deep, liu2019adversarial}, we adopt the SOTA inpainting method~\cite{li2020recurrent} to fill one annotated region that is randomly removed from each chosen MS COCO image. As to the pristine class, we simply select images from the MS COCO dataset. 

\begin{figure}[t]
	\centering
	\begin{minipage}[t]{0.24\linewidth}
		\centering
		\includegraphics[width=\linewidth]{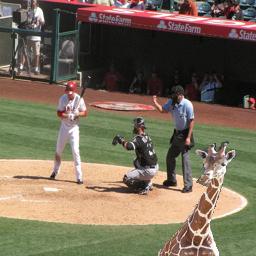}
	\end{minipage}
	\begin{minipage}[t]{0.24\linewidth}
		\centering
		\includegraphics[width=\linewidth]{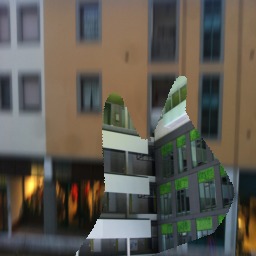}
	\end{minipage}
	\begin{minipage}[t]{0.24\linewidth}
		\centering
		\includegraphics[width=\linewidth]{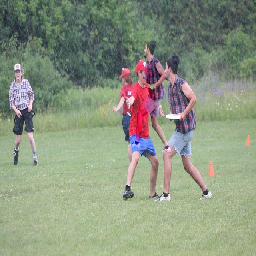}
	\end{minipage}
	\begin{minipage}[t]{0.24\linewidth}
		\centering
		\includegraphics[width=\linewidth]{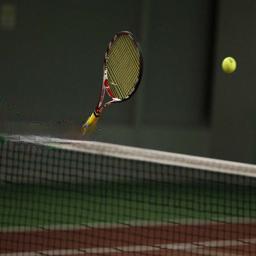}
	\end{minipage} \\
	\vspace{1mm}
	\begin{minipage}[t]{0.24\linewidth}
		\centering
		\includegraphics[width=\linewidth]{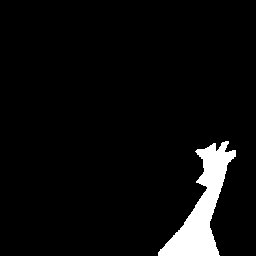}
		\scriptsize{(a) Content-aware splicing}
	\end{minipage}
	\begin{minipage}[t]{0.24\linewidth}
		\centering
		\includegraphics[width=\linewidth]{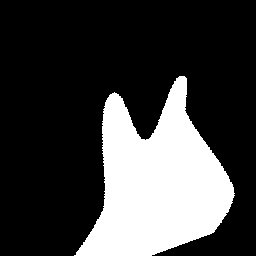}
		\scriptsize{(b) Random-mask splicing}
	\end{minipage}
	\begin{minipage}[t]{0.24\linewidth}
		\centering
		\includegraphics[width=\linewidth]{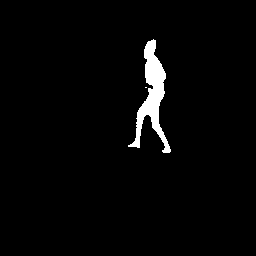}
		\scriptsize{(c) Copy-move}
	\end{minipage}
	\begin{minipage}[t]{0.24\linewidth}
		\centering
		\includegraphics[width=\linewidth]{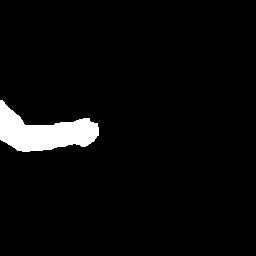}
		\scriptsize{(d) Removal}
	\end{minipage}
	\caption{Examples from our synthetic dataset. The generated images of different manipulation types and their ground-truth masks are demonstrated.}
	\label{fig:dataset_visualization}
\end{figure}

In summary, we have $116,583$ images in splicing class, $100,000$ images in copy-move class, $78,246$ images in removal class, and $81,910$ images in pristine class, thus $\sim$$0.38$M in total. Examples of different manipulation types in our synthetic dataset are demonstrated in Fig~\ref{fig:dataset_visualization}. It should be emphasized that our training dataset is much smaller than that of MantraNet and SPAN, where massive annotated data ($1.25$M) is used to train their feature extractor, not to mention the large number of synthesized manipulations for training the rest of their networks.

As it is inefficient to train all manipulated images in one epoch, we uniformly sample $0.025$M images per class to form a $0.1$M dataset on-the-fly for training in each epoch. In addition, we also build a validation set that contains $4 \times 100$ images. The size of synthetic images are all set to $256\times 256$.

\section{Experiments}

\begin{table}[t]
	\centering
	\caption{Summary of test datasets for our pre-trained and fine-tuned models ($\#$ stands for the number of images. \ding{52} and \ding{55} indicate whether or not the manipulation type is involved).}
	\renewcommand{\arraystretch}{1.4}
	\begin{adjustbox}{width=\linewidth}
		\begin{tabu}{lccccccc}
			\toprule
			\multirow{2}{*}{Dataset} & \multicolumn{1}{c}{Pre-trained} && \multicolumn{2}{c}{Fine-tuned} & \multirow{2}{*}{Splicing} & \multirow{2}{*}{Copy-move} & \multirow{2}{*}{Removal} \\
			\cline{2-2} \cline{4-5} & $\#$ Test  && $\#$ Train & $\#$ Test \\ \hline
			Columbia & $180$ && $-$ & $-$ & \ding{52} & \ding{55} & \ding{55} \\ 
			Coverage & $100$ && $75$ & $25$ & \ding{55} & \ding{52} & \ding{55} \\
			CASIA & $6,044$ && $5,123$ & $921$ & \ding{52} & \ding{52} & \ding{55} \\
			NIST & $564$ && $404$ & $160$ & \ding{52} & \ding{52} & \ding{52} \\
			IMD20 & $2,010$ && $-$ & $-$ & \ding{52} & \ding{52} & \ding{52} \\
			\bottomrule
		\end{tabu}
	\end{adjustbox}
	\label{table:summary}
\end{table}

\subsection{Experimental Setup}
\subsubsection{Test data} 
We evaluate the manipulation localization on four standard test datasets: Columbia~\cite{ng2009columbia}, Coverage~\cite{wen2016coverage}, CASIA~\cite{dong2013casia} and NIST16~\cite{NIST}, and one real-world dataset: IMD20~\cite{novozamsky2020imd2020}. To finetune PSCC-Net, we follow the same training/testing split on Coverage, CASIA, and NIST16 as in~\cite{zhou2018learning, hu2020span} for fair comparisons.
Specifically, Columbia~\cite{ng2009columbia} is a splicing dataset of $180$ images.
Coverage~\cite{wen2016coverage} is a copy-move dataset of $100$ images; for fine-tuning, it is split into $75/25$ for training and testing.
CASIA~\cite{dong2013casia} (v$1.0$ + v$2.0$) includes both splicing and copy-move;
for fine-tuning, $5,123$ images from v2.0 is adopted for training, and $921$ images from v1.0 is for testing.
NIST16~\cite{NIST} has $564$ images, involving all three manipulations; for fine-tuning, $404$ images are used for training and $160$ for testing. IMD20~\cite{novozamsky2020imd2020} consists of $2,010$ real-life manipulated images collected from Internet, and involves all three manipulations as well. We summarize the manipulation types for each test dataset and the number of images for evaluating our pre-trained and fine-tuned models  in Tab.~\ref{table:summary}.

As the manipulation detection is not considered by recent works, there is no standard dataset for benchmarking. 
Since CASIA is the only test dataset in here that corresponds each manipulated image to its pristine image, we use both forged and pristine images and define an evaluation protocol for detection. This dataset is named CASIA-D and consists of $1,842$ images with $50\%$ forged and $50\%$ pristine.



\subsubsection{Metrics} 
To quantify the localization performance, following previous works~\cite{wu2019mantra, hu2020span}, we use pixel-level Area Under Curve (AUC) and F$1$ score on manipulation masks.
To evaluate the detection performance, we use image-level AUC and F$1$ score, Equal Error Rate (EER), and True Positive Rate at $1\%$ false positive rate (TPR{$_{1\%}$}). Since binary masks and detection scores are required to compute F$1$ scores, we adopt the EER threshold to binarize them.


\subsubsection{Implementation details} 
PSCC-Net is end-to-end trainable and light-weighted. Its top-down path and bottom-up path have $2.0$ and $1.6$ Million (M) parameters. In the bottom-up path, the detection head has $0.9$ M and the rest part (for localization) has only $0.7$ M parameters. In comparison, the ManTra-Net~\cite{wu2019mantra} and SPAN~\cite{hu2020span} have $3.8$ and $3.7$ M parameters, respectively. 
Implemented by PyTorch, our model is trained with GeForce GTX $1080$Ti.
We initialize our backbone with ImageNet pre-trained weights, and optimize the whole model by Adam~\cite{kingma2014adam} with a batch size of $10$ and an initial learning rate of $2e$-$4$. 
The learning rate is halved every $5$ epochs and the total training period is $25$ epochs. 

Our network can take arbitrary-size images as input. To avoid performance degradation caused by size mismatch between training (\textit{e.g.}, $256 \times 256$) and testing data (\textit{e.g.}, $4,000 \times 3,000$), at the end of top-down path, we resample the extracted features from the first to the last scales respectively into  fixed sizes $256\times256$, $128\times128$, $64\times64$, and $32\times32$, where the ratio $r$ in SCCM is set to $4$, $2$, $2$, and $1$ respectively to reduce the computational burden. The produced masks are resampled back to the same size as the input image for localization evaluation.


\subsection{Comparisons on  Localization}

The compared IMDL methods include J-LSTM~\cite{bappy2017exploiting}, H-LSTM~\cite{bappy2019hybrid}, RGB-N~\cite{zhou2018learning}, ManTra-Net~\cite{wu2019mantra}, and SPAN~\cite{hu2020span} where SPAN has reported the SOTA performance on localization.
Following the evaluation protocol defined in SPAN~\cite{hu2020span}, we compare the localization performance using two models: 1) the pre-trained model is trained on the synthetic dataset and evaluated on the \textit{full} test datasets, and 2) the fine-tuned model is the pre-trained model further fine-tuned on the training split of test datasets and evaluated on their \textit{test split}. The pre-trained model is to show the generalization ability of each method, and the fine-tuned model is to manifest their localization performance while the domain discrepancy has been greatly alleviated. Note that the reported results of all compared methods are either from their original papers or by running their public codes.

\begin{table}[t!]
	\centering
	\caption{Localization AUC ($\%$) of pre-trained models.}
	\renewcommand{\arraystretch}{1.4}
	\begin{adjustbox}{width=\linewidth}
		\begin{tabu}{llllll}
			\toprule
			Method & Columbia & Coverage & CASIA & NIST16 & IMD20 \\ \hline
			ManTra-Net~\cite{wu2019mantra} & $82.4$ & $81.9$ & $81.7$ & $79.5$ & $74.8$\\ 
			SPAN~\cite{hu2020span} & $93.6$ & $\mathbf{92.2}$ & $79.7$ & $84.0$ & $ 75.0 $\\ \hline
			PSCC-Net & $\mathbf{98.2}$ & $84.7$ & $\mathbf{82.9}$ & $\mathbf{85.5}$ & $\mathbf{80.6}$\\
			\bottomrule
		\end{tabu}
	\end{adjustbox}
	\label{table:pretrain}
\end{table}

\begin{table}[t!]
	\centering
	\footnotesize
	\caption{Evaluation of the fine-tuned models. Localization AUC/F$1$s are reported (in $\%$). ManTra-Net is not shown here as it has only developed the pre-trained model.}
	\renewcommand{\arraystretch}{1.4}
	\begin{adjustbox}{width=0.85\linewidth}
		\begin{tabu}{llll}
			\toprule
			Method & Coverage & CASIA & NIST16\\\midrule
			J-LSTM~\cite{bappy2017exploiting} & $61.4$ / - & - / - & $76.4$ / - \\
			H-LSTM~\cite{bappy2019hybrid}  & $71.2$ / - & - / - & $79.4$ / - \\
			RGB-N~\cite{zhou2018learning}  & $81.7$ / $43.7$ & $79.5$ / $40.8$ & $93.7$ / $72.2$ \\
			SPAN~\cite{hu2020span} & $93.7$ / $55.8$ & $83.8$ / $38.2$ & $96.1$ / $58.2$ \\ \hline
			PSCC-Net & $\mathbf{94.1}$ / $\mathbf{72.3}$ & $\mathbf{87.5}$ / $\mathbf{55.4}$ & $\mathbf{99.1}$ / $\mathbf{74.2}$ \\
			\bottomrule
		\end{tabu}
	\end{adjustbox}
	\label{table:finetune}
\end{table}

\subsubsection{Pre-trained model}
We choose the best pre-trained model based on the performance on our validation set.
Tab.~\ref{table:pretrain} shows the localization performance of pre-trained models for different methods on four standard datasets and one real-world dataset under pixel-level AUC.
The pre-trained PSCC-Net achieves the best localization performance on Columbia, CASIA, NIST16, and IMD20, and ranks the second on Coverage. The most significant performance gain is achieved while tackling real-life manipulated images ($5.6\%\uparrow$). This validates that the PSCC-Net has the best generalization ability as compared to the others.
We fail to achieve the best performance on Coverage, despite surpassing ManTra-Net $2.8\%$ under AUC. The reason might be the imperfection of our training data for the case, where the copied object is intentionally moved to cover a pristine object with similar appearance. 
Indeed, by fine-tuning the pre-trained model on Coverage, PSCC-Net achieves the $0.4\%$ gain over SPAN under AUC (Tab.~\ref{table:finetune}).

%

\begin{figure}[t!]
	\centering
	\begin{minipage}[h]{0.18\linewidth}
		\centering
		\includegraphics[width=\linewidth]{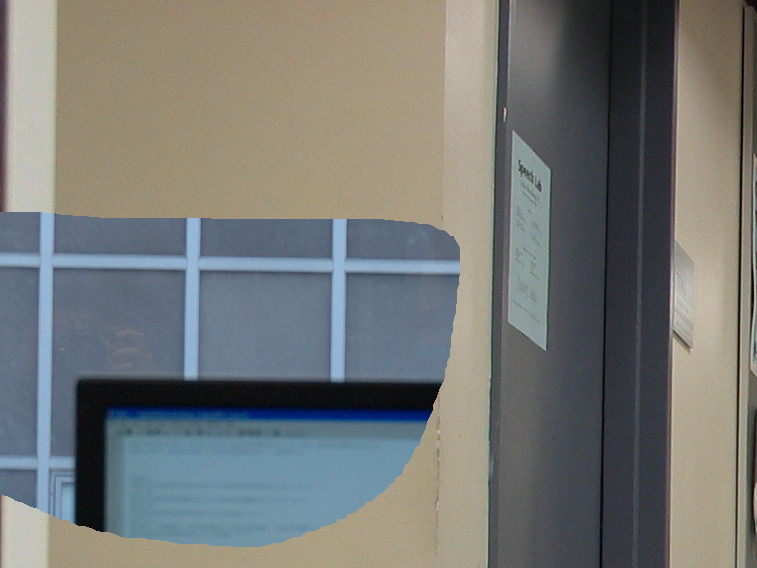}
	\end{minipage}
	\vspace{1mm}
	\begin{minipage}[h]{0.18\linewidth}
		\centering
		\includegraphics[width=\linewidth]{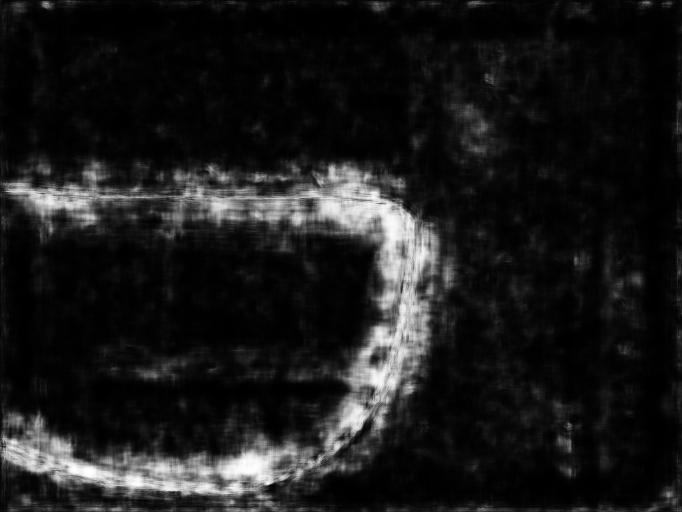}
	\end{minipage}
	\begin{minipage}[h]{0.18\linewidth}
		\centering
		\includegraphics[width=\linewidth]{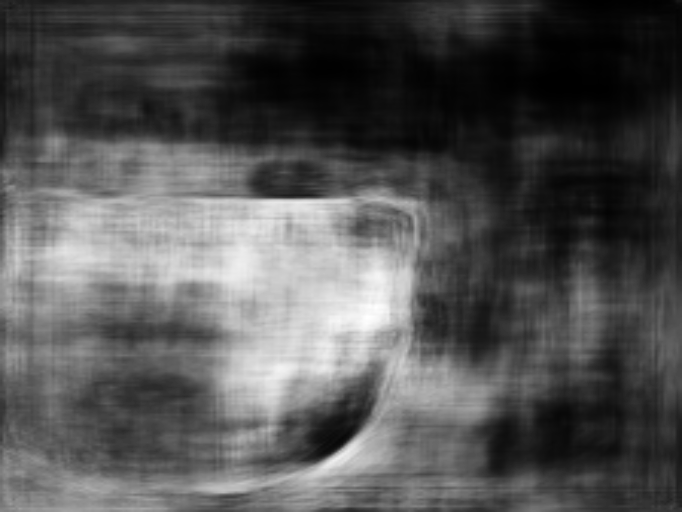}
	\end{minipage}
	\begin{minipage}[h]{0.18\linewidth}
		\centering
		\includegraphics[width=\linewidth]{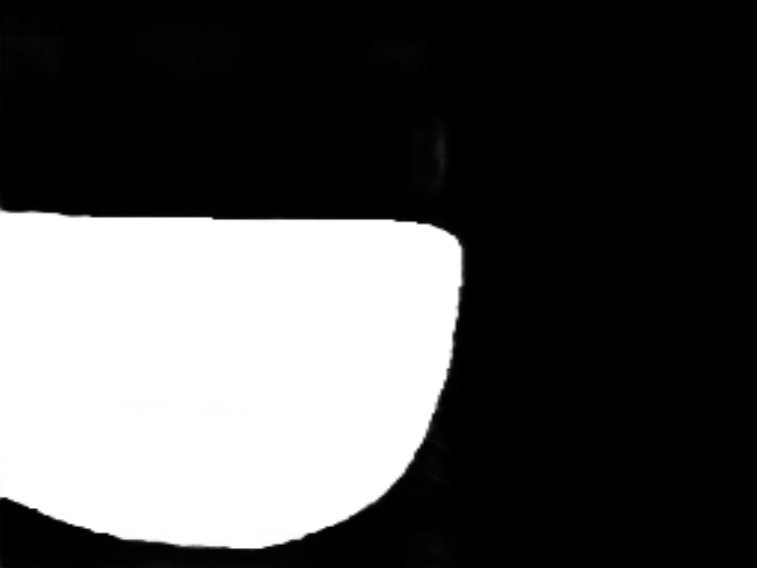}
	\end{minipage}
	\begin{minipage}[h]{0.18\linewidth}
		\centering
		\includegraphics[width=\linewidth]{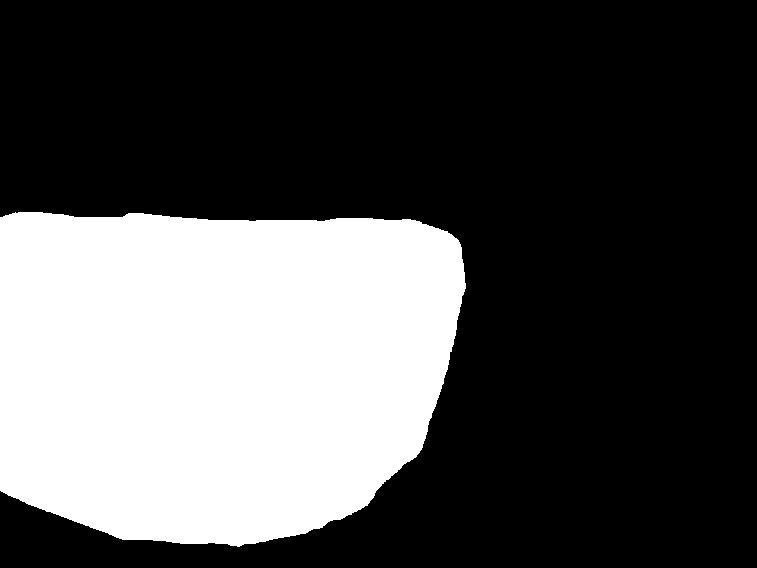}
	\end{minipage}
	\begin{minipage}[h]{0.18\linewidth}
		\centering
		\includegraphics[width=\linewidth]{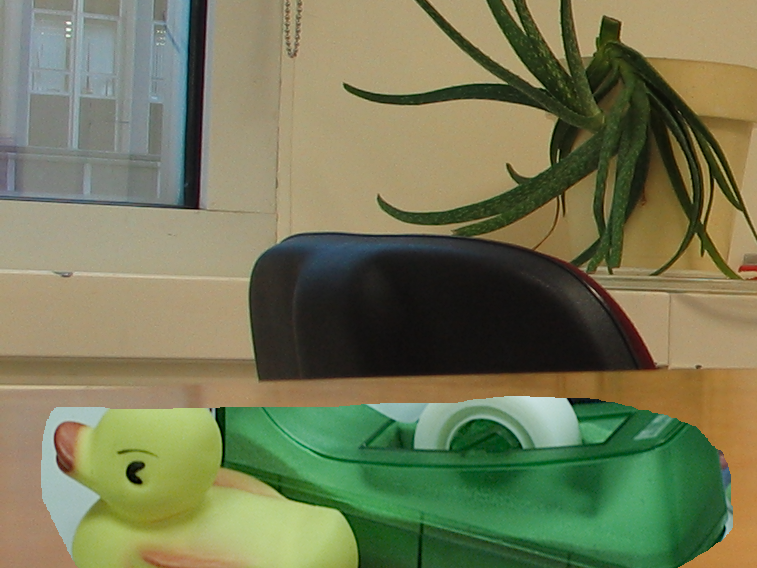}
	\end{minipage}
	\vspace{1mm}
	\begin{minipage}[h]{0.18\linewidth}
		\centering
		\includegraphics[width=\linewidth]{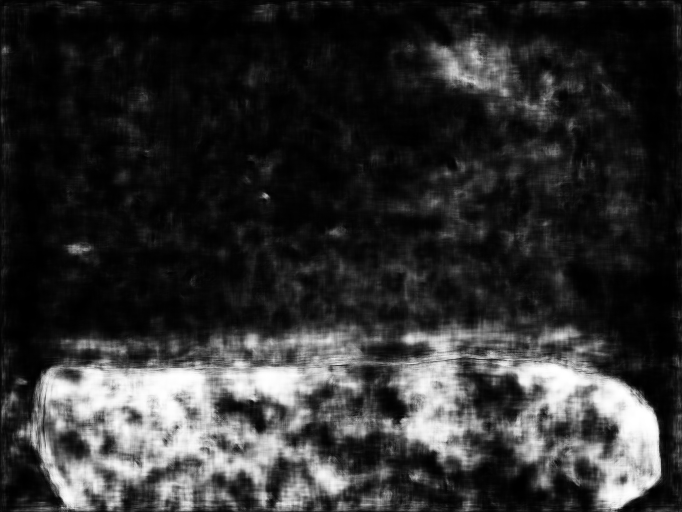}
	\end{minipage}
	\begin{minipage}[h]{0.18\linewidth}
		\centering
		\includegraphics[width=\linewidth]{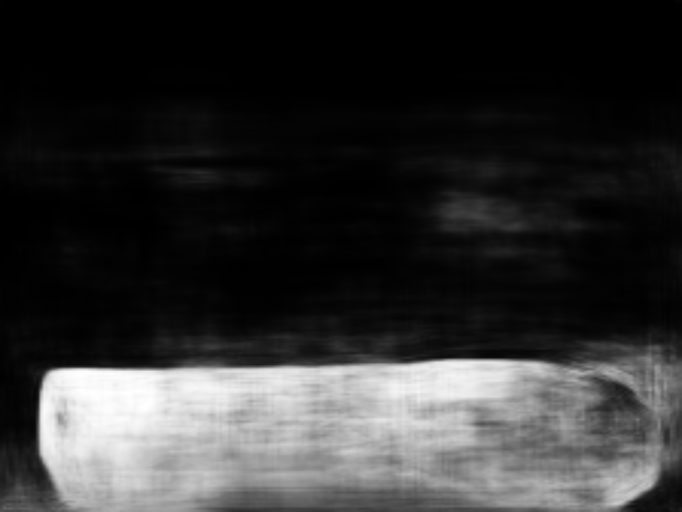}
	\end{minipage}
	\begin{minipage}[h]{0.18\linewidth}
		\centering
		\includegraphics[width=\linewidth]{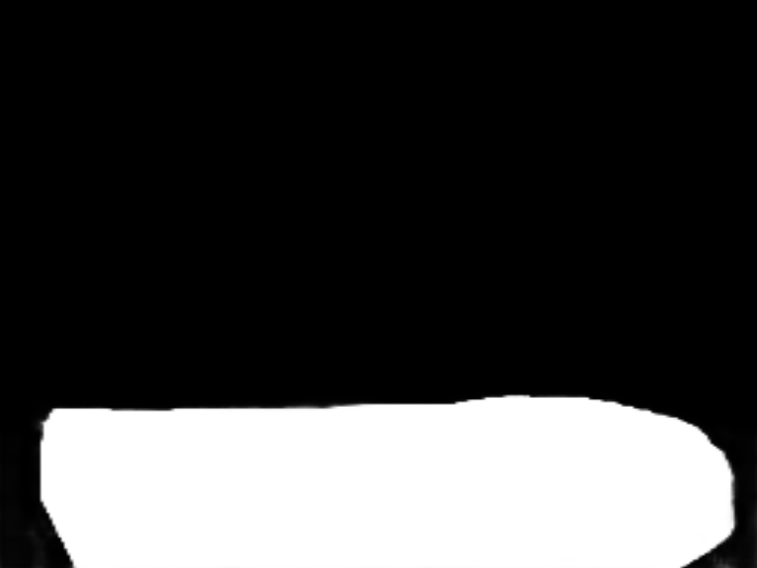}
	\end{minipage}
	\begin{minipage}[h]{0.18\linewidth}
		\centering
		\includegraphics[width=\linewidth]{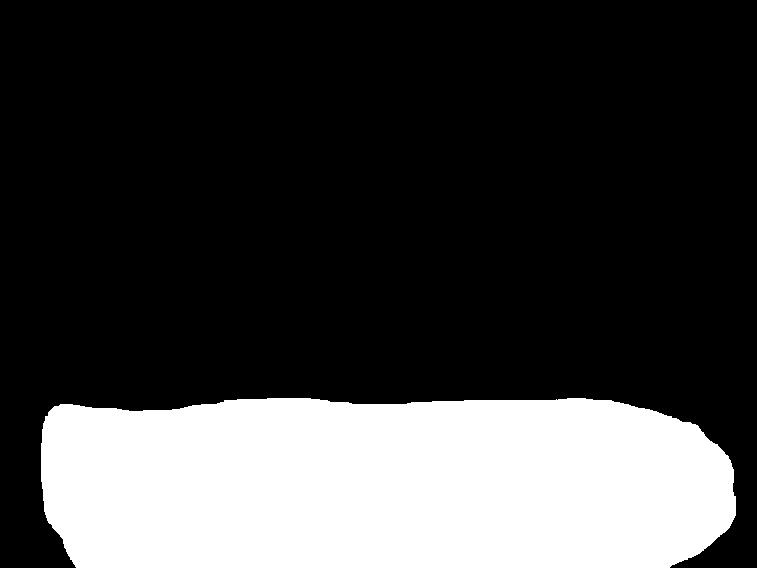}
	\end{minipage}
	\begin{minipage}[h]{0.18\linewidth}
	\centering
	\includegraphics[width=\linewidth]{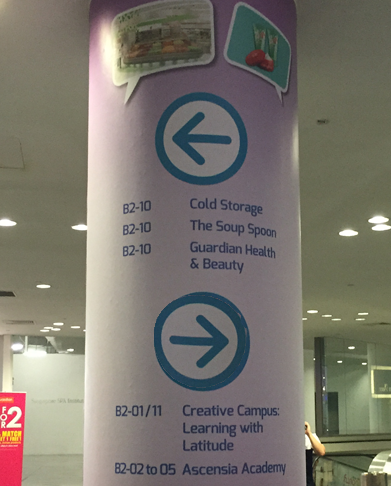}
	\end{minipage}
	\vspace{1mm}
	\begin{minipage}[h]{0.18\linewidth}
		\centering
		\includegraphics[width=\linewidth]{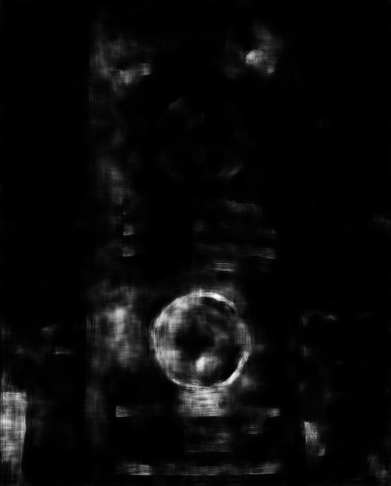}
	\end{minipage}
	\begin{minipage}[h]{0.18\linewidth}
		\centering
		\includegraphics[width=\linewidth]{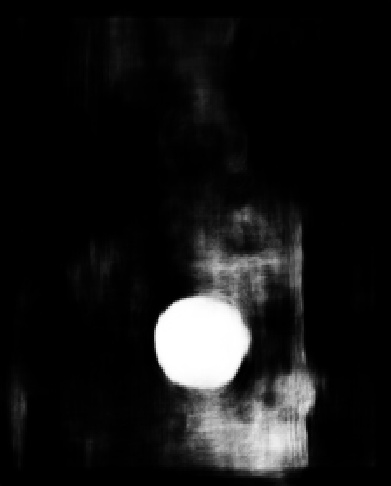}
	\end{minipage}
	\begin{minipage}[h]{0.18\linewidth}
		\centering
		\includegraphics[width=\linewidth]{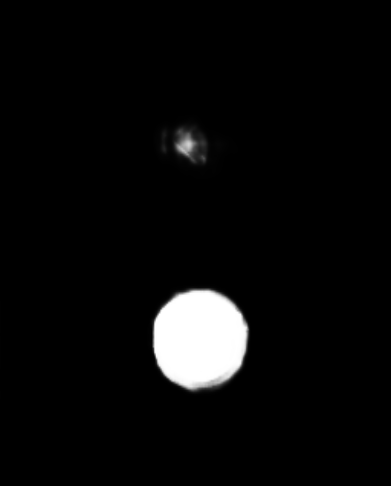}
	\end{minipage}
	\begin{minipage}[h]{0.18\linewidth}
		\centering
		\includegraphics[width=\linewidth]{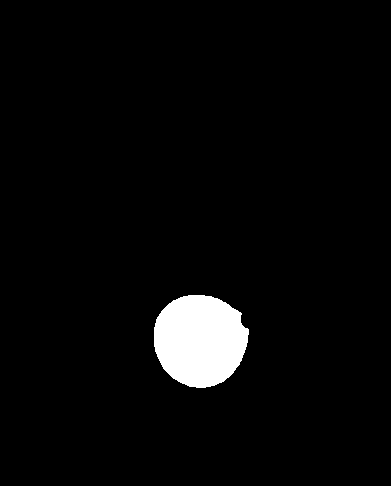}
	\end{minipage}
	\begin{minipage}[h]{0.18\linewidth}
	\centering
	\includegraphics[width=\linewidth]{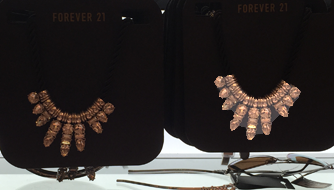}
	\end{minipage}
	\vspace{1mm}
	\begin{minipage}[h]{0.18\linewidth}
		\centering
		\includegraphics[width=\linewidth]{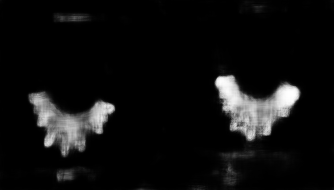}
	\end{minipage}
	\begin{minipage}[h]{0.18\linewidth}
		\centering
		\includegraphics[width=\linewidth]{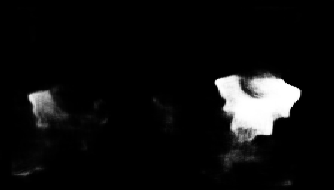}
	\end{minipage}
	\begin{minipage}[h]{0.18\linewidth}
		\centering
		\includegraphics[width=\linewidth]{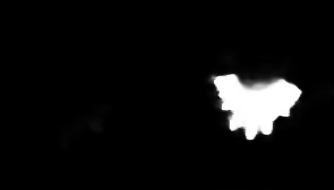}
	\end{minipage}
	\begin{minipage}[h]{0.18\linewidth}
		\centering
		\includegraphics[width=\linewidth]{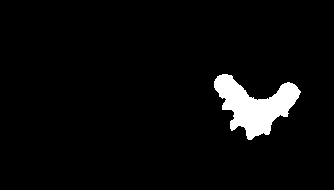}
	\end{minipage}
	\begin{minipage}[h]{0.18\linewidth}
		\centering
		\includegraphics[width=\linewidth]{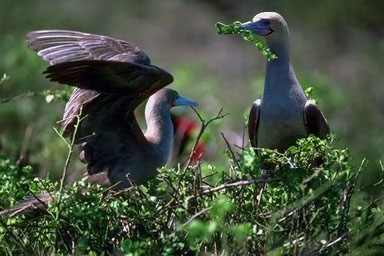}
	\end{minipage}
	\vspace{1mm}
	\begin{minipage}[h]{0.18\linewidth}
		\centering
		\includegraphics[width=\linewidth]{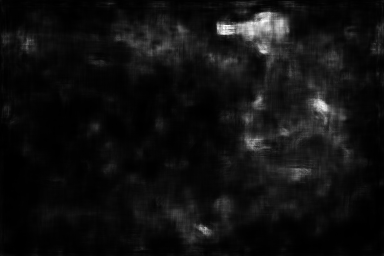}
	\end{minipage}
	\begin{minipage}[h]{0.18\linewidth}
		\centering
		\includegraphics[width=\linewidth]{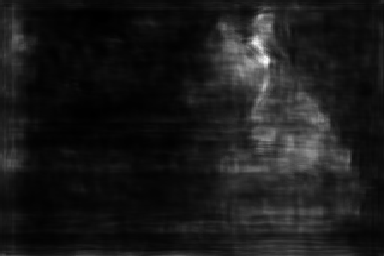}
	\end{minipage}
	\begin{minipage}[h]{0.18\linewidth}
		\centering
		\includegraphics[width=\linewidth]{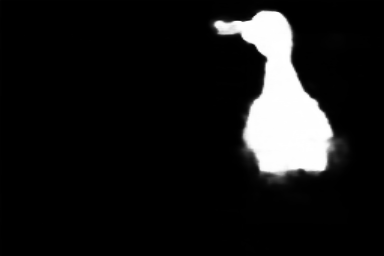}
	\end{minipage}
	\begin{minipage}[h]{0.18\linewidth}
		\centering
		\includegraphics[width=\linewidth]{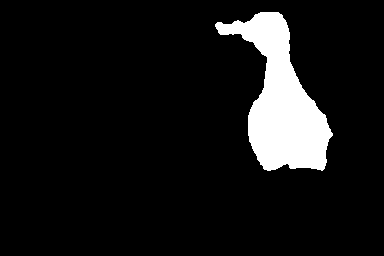}
	\end{minipage}
	\begin{minipage}[h]{0.18\linewidth}
		\centering
		\includegraphics[width=\linewidth]{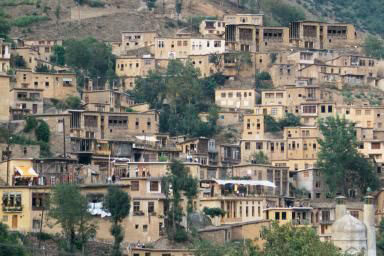}
	\end{minipage}
	\vspace{1mm}
	\begin{minipage}[h]{0.18\linewidth}
		\centering
		\includegraphics[width=\linewidth]{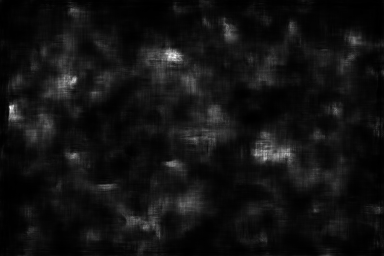}
	\end{minipage}
	\begin{minipage}[h]{0.18\linewidth}
		\centering
		\includegraphics[width=\linewidth]{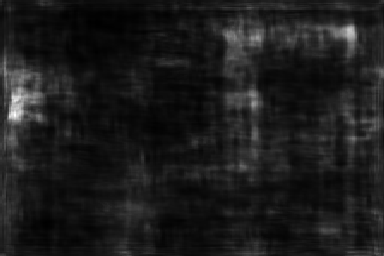}
	\end{minipage}
	\begin{minipage}[h]{0.18\linewidth}
		\centering
		\includegraphics[width=\linewidth]{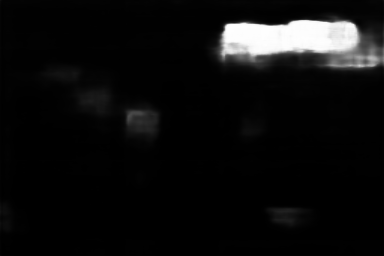}
	\end{minipage}
	\begin{minipage}[h]{0.18\linewidth}
		\centering
		\includegraphics[width=\linewidth]{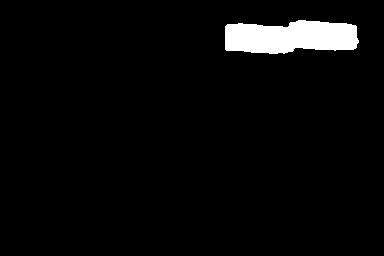}
	\end{minipage}
	\begin{minipage}[h]{0.18\linewidth}
		\centering
		\includegraphics[width=\linewidth]{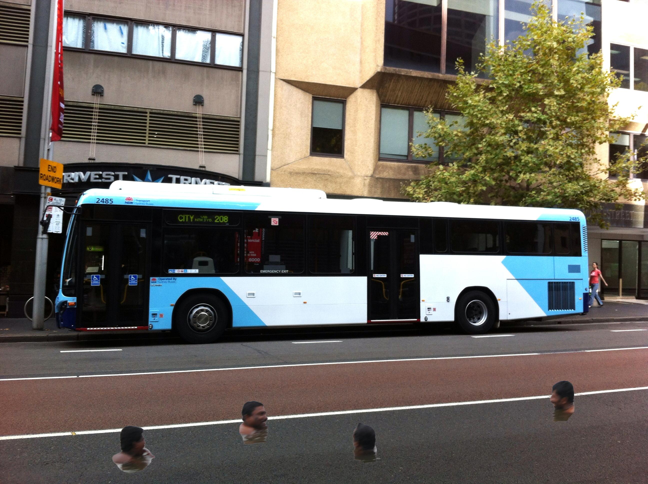}
	\end{minipage}
	\vspace{1mm}
	\begin{minipage}[h]{0.18\linewidth}
		\centering
		\includegraphics[width=\linewidth]{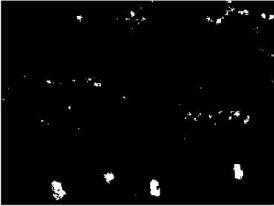}
	\end{minipage}
	\begin{minipage}[h]{0.18\linewidth}
		\centering
		\includegraphics[width=\linewidth]{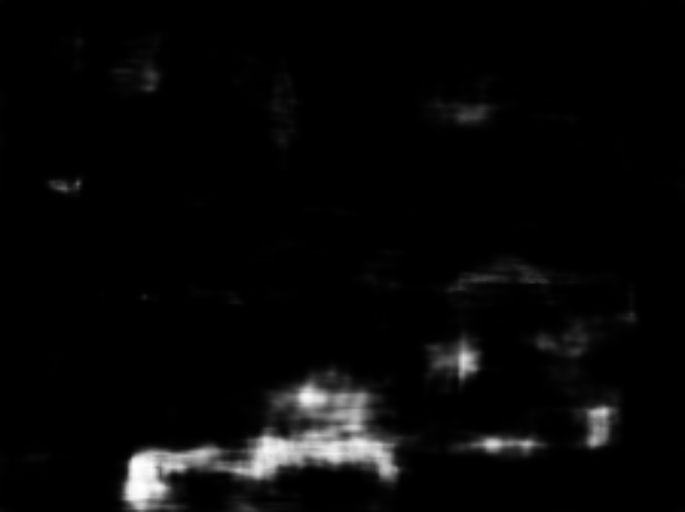}
	\end{minipage}
	\begin{minipage}[h]{0.18\linewidth}
		\centering
		\includegraphics[width=\linewidth]{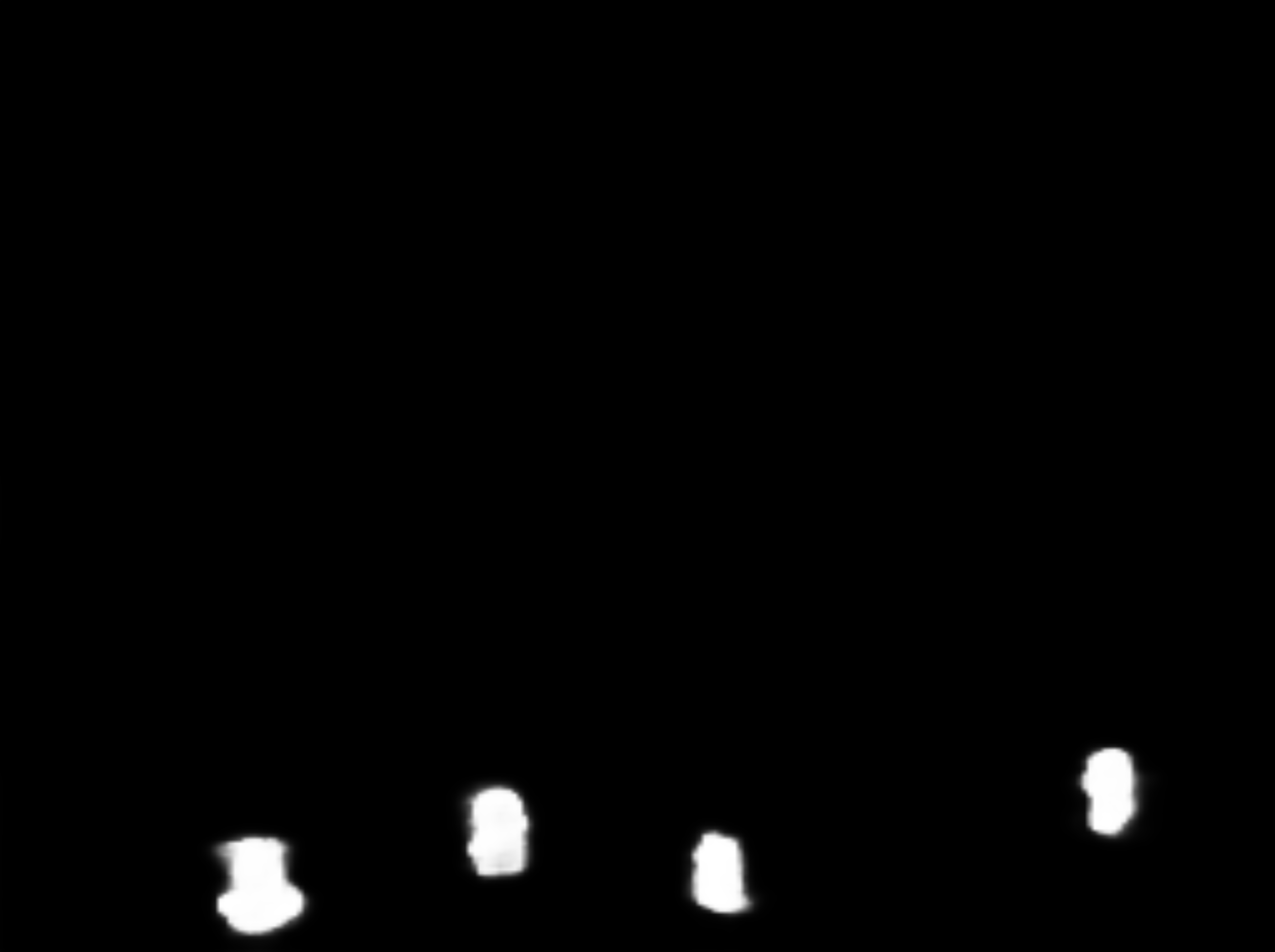}
	\end{minipage}
	\begin{minipage}[h]{0.18\linewidth}
		\centering
		\includegraphics[width=\linewidth]{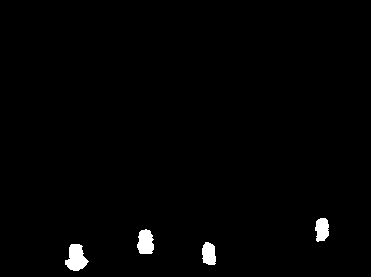}
	\end{minipage}
	\begin{minipage}[h]{0.18\linewidth}
	\centering
	\includegraphics[width=\linewidth]{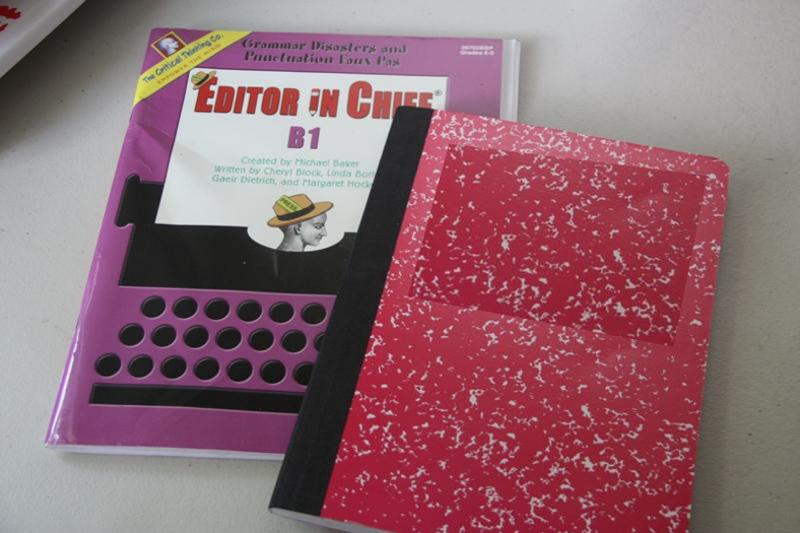}
	\scriptsize{(a) Manipulated}
	\end{minipage}
	\begin{minipage}[h]{0.18\linewidth}
		\centering
		\includegraphics[width=\linewidth]{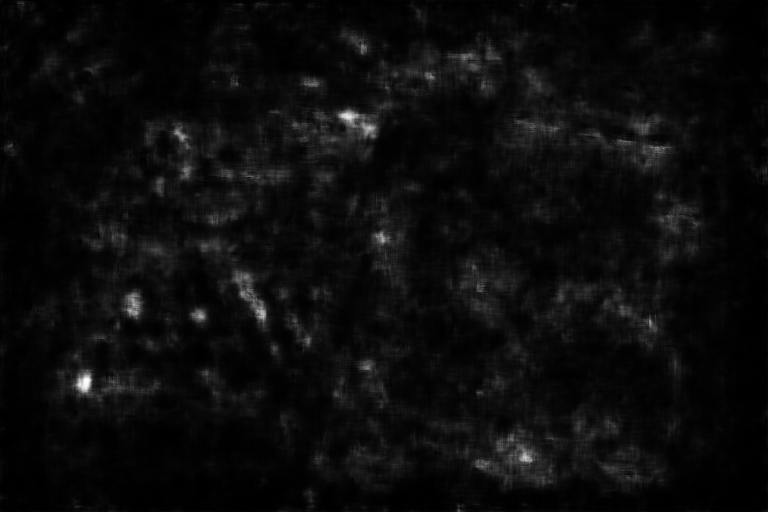}
	\scriptsize{(b) ManTra-Net}
	\end{minipage}
	\begin{minipage}[h]{0.18\linewidth}
		\centering
		\includegraphics[width=\linewidth]{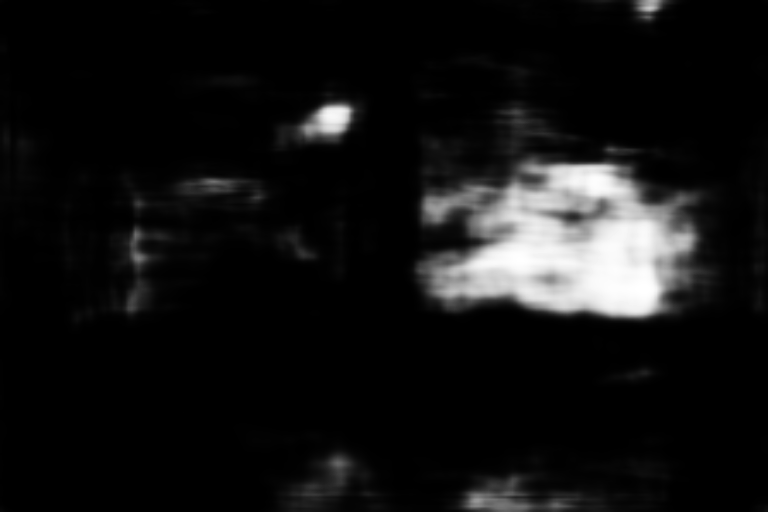}
	\scriptsize{(c) SPAN}
	\end{minipage}
	\begin{minipage}[h]{0.18\linewidth}
		\centering
		\includegraphics[width=\linewidth]{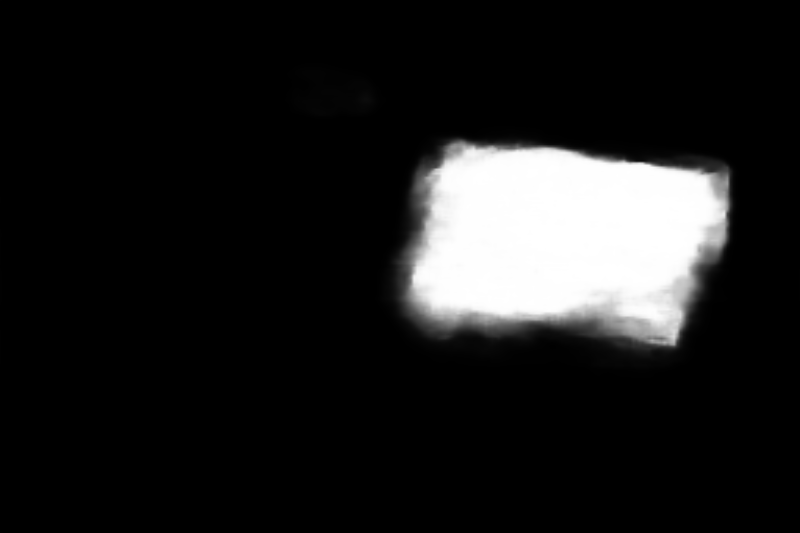}
	\scriptsize{(d) PSCC-Net}
	\end{minipage}
	\begin{minipage}[h]{0.18\linewidth}
		\centering
		\includegraphics[width=\linewidth]{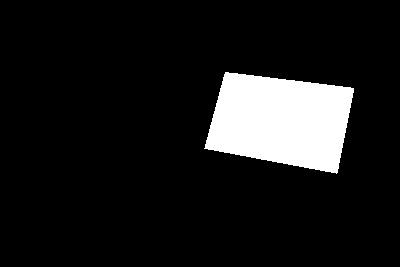}
	\scriptsize{(e) GT}
	\end{minipage}
	\caption{Qualitative localization evaluations on four standard test datasets. From top to bottom, our PSCC-Net is compared to SOTAs on Columbia, Coverage, CASIA, and NIST16 datasets respectively, each with two images. }
	\label{fig:experiments}
\end{figure}

\begin{figure}[t!]
	\centering
	\begin{minipage}[h]{0.18\linewidth}
		\centering
		\includegraphics[width=\linewidth]{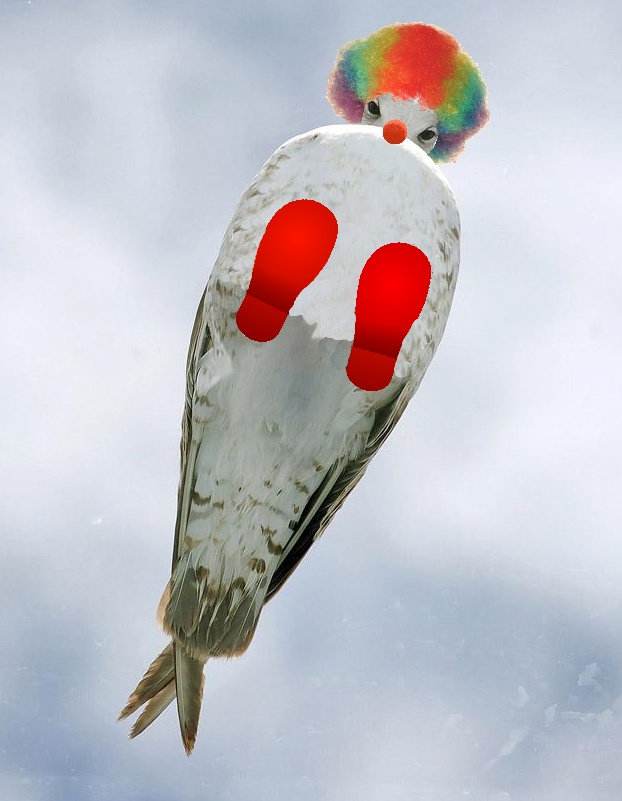}
	\end{minipage}
	\vspace{1mm}
	\begin{minipage}[h]{0.18\linewidth}
		\centering
		\includegraphics[width=\linewidth]{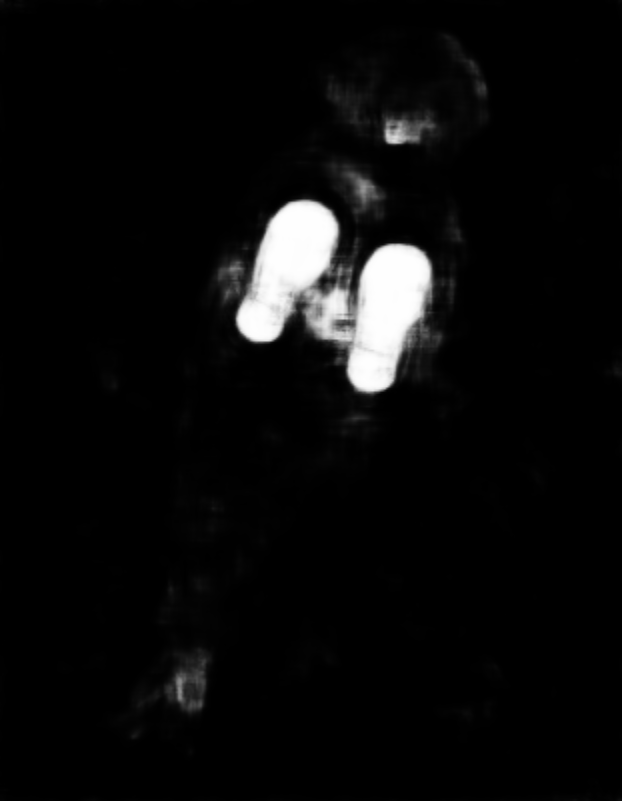}
	\end{minipage}
	\begin{minipage}[h]{0.18\linewidth}
		\centering
		\includegraphics[width=\linewidth]{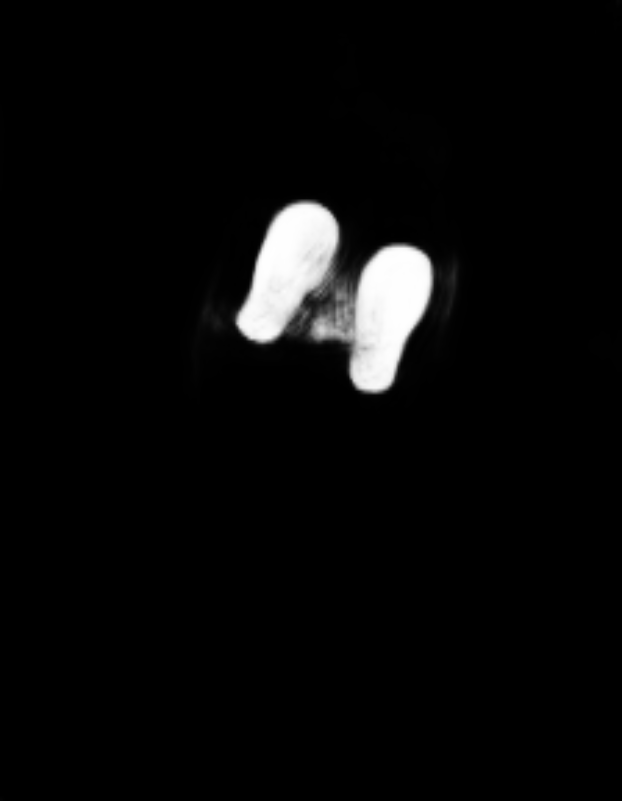}
	\end{minipage}
	\begin{minipage}[h]{0.18\linewidth}
		\centering
		\includegraphics[width=\linewidth]{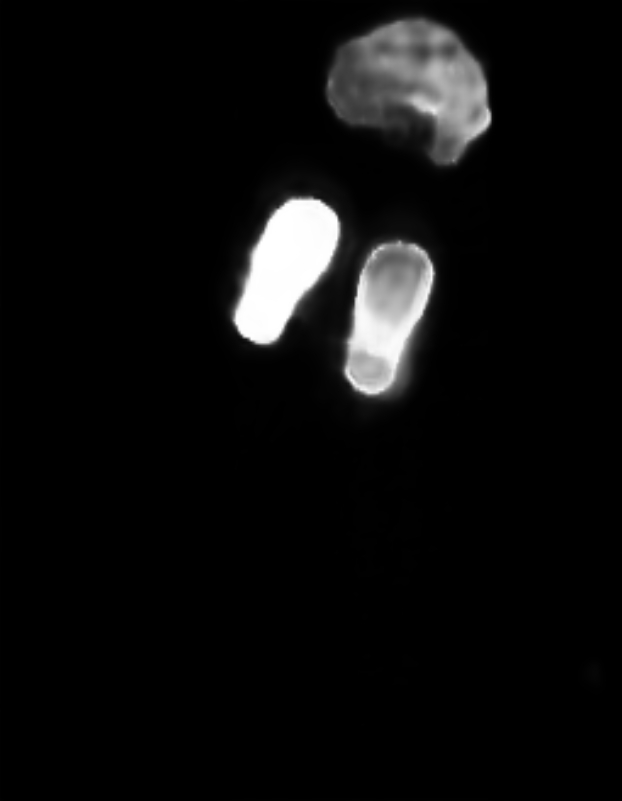}
	\end{minipage}
	\begin{minipage}[h]{0.18\linewidth}
		\centering
		\includegraphics[width=\linewidth]{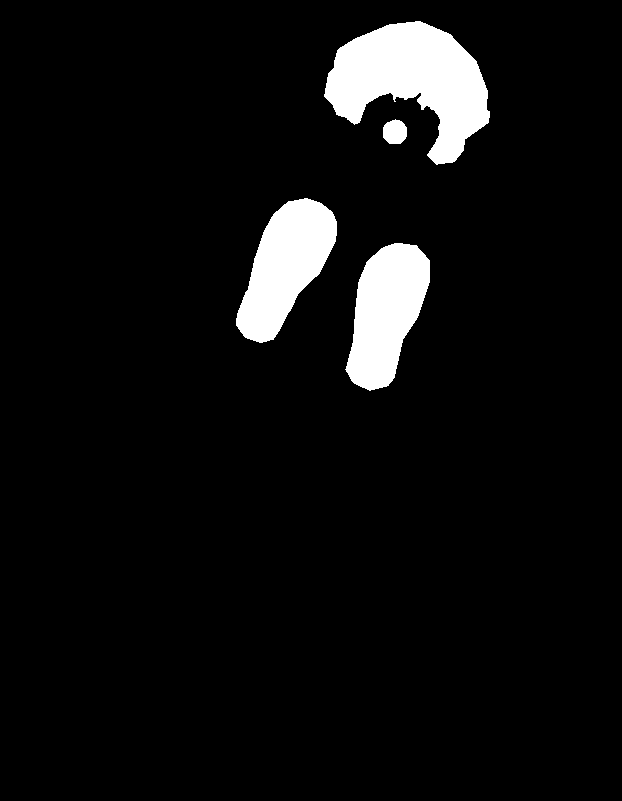}
	\end{minipage}
	\begin{minipage}[h]{0.18\linewidth}
		\centering
		\includegraphics[width=\linewidth]{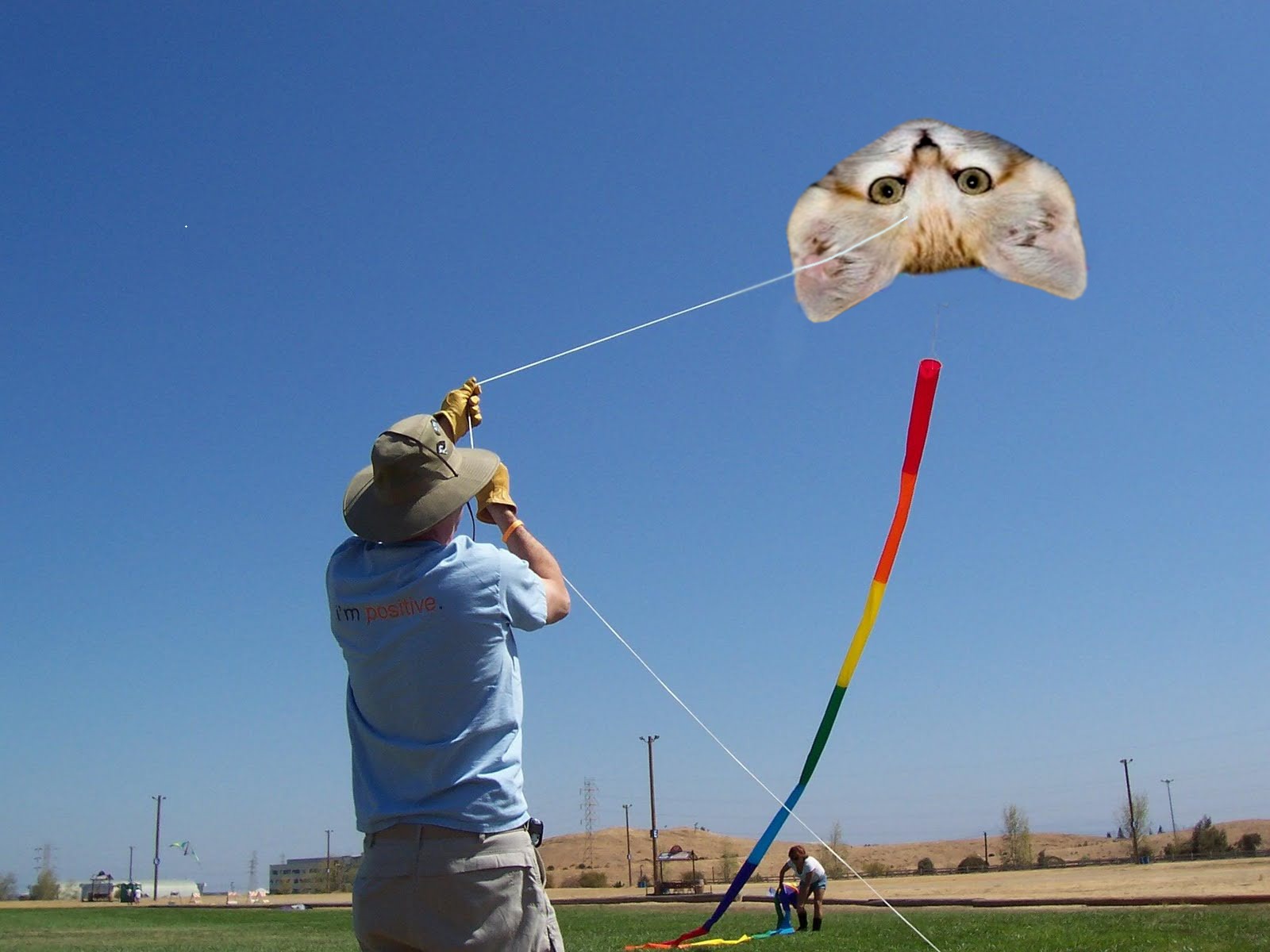}
	\end{minipage}
	\vspace{1mm}
	\begin{minipage}[h]{0.18\linewidth}
		\centering
		\includegraphics[width=\linewidth]{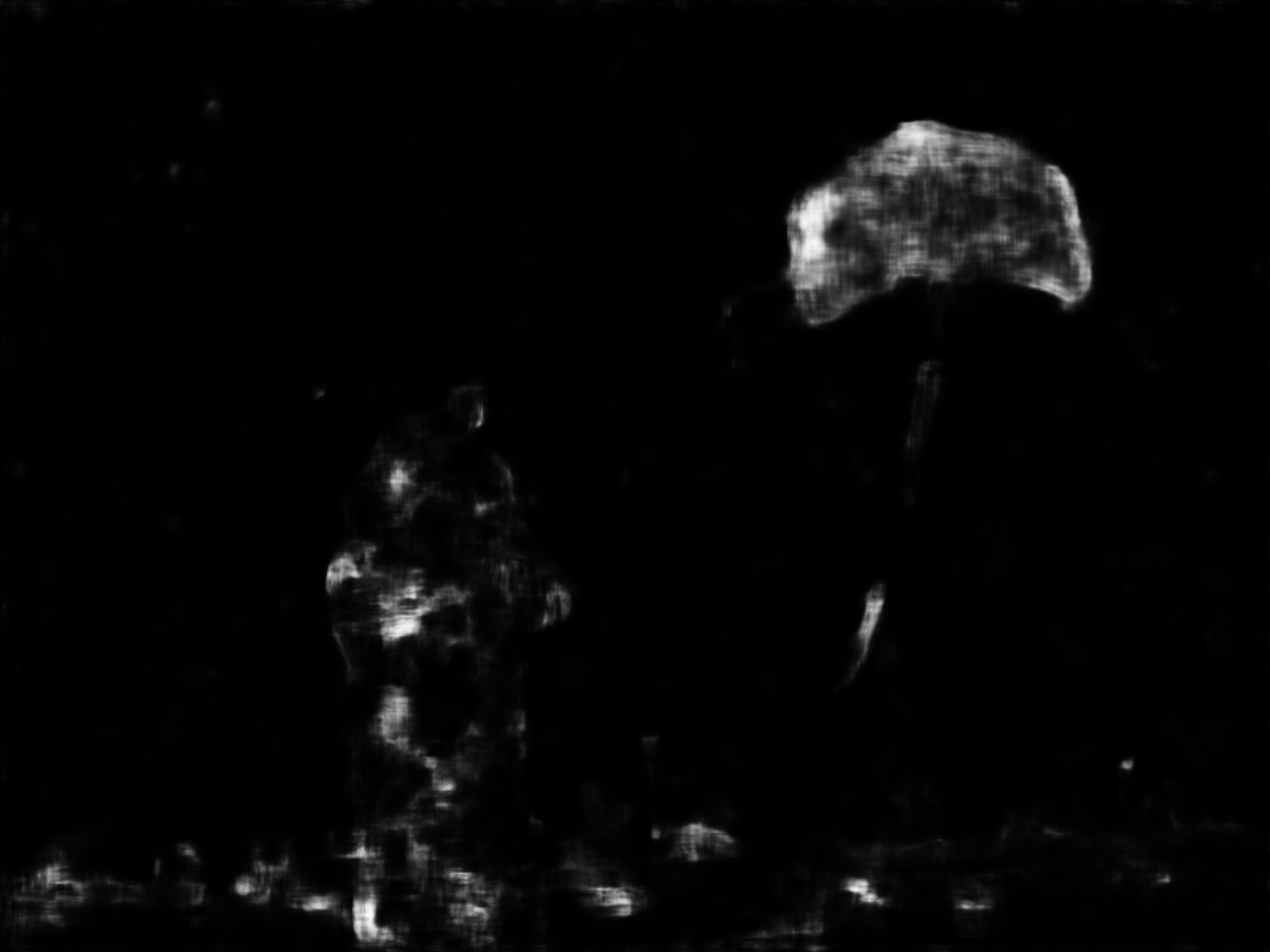}
	\end{minipage}
	\begin{minipage}[h]{0.18\linewidth}
		\centering
		\includegraphics[width=\linewidth]{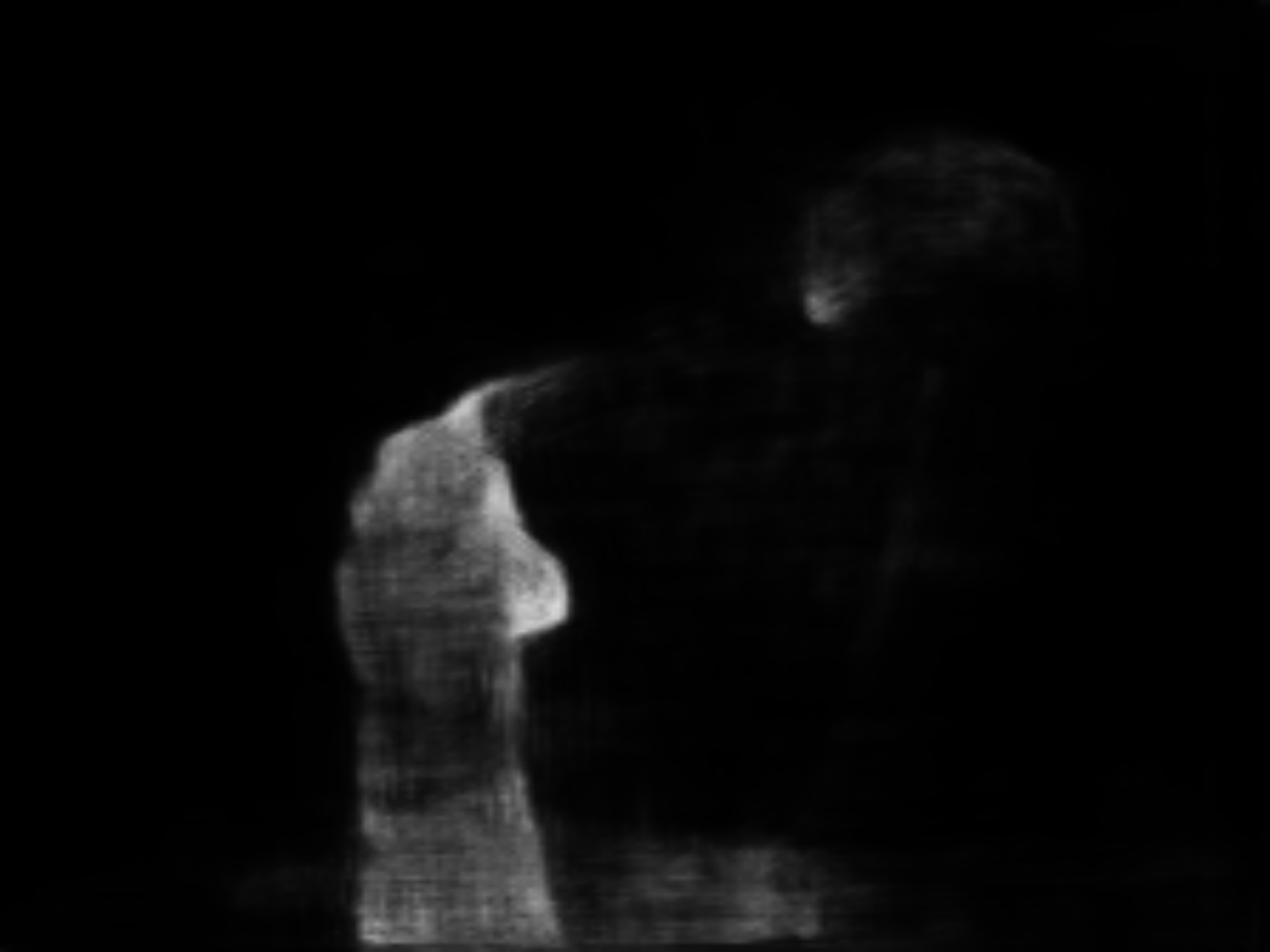}
	\end{minipage}
	\begin{minipage}[h]{0.18\linewidth}
		\centering
		\includegraphics[width=\linewidth]{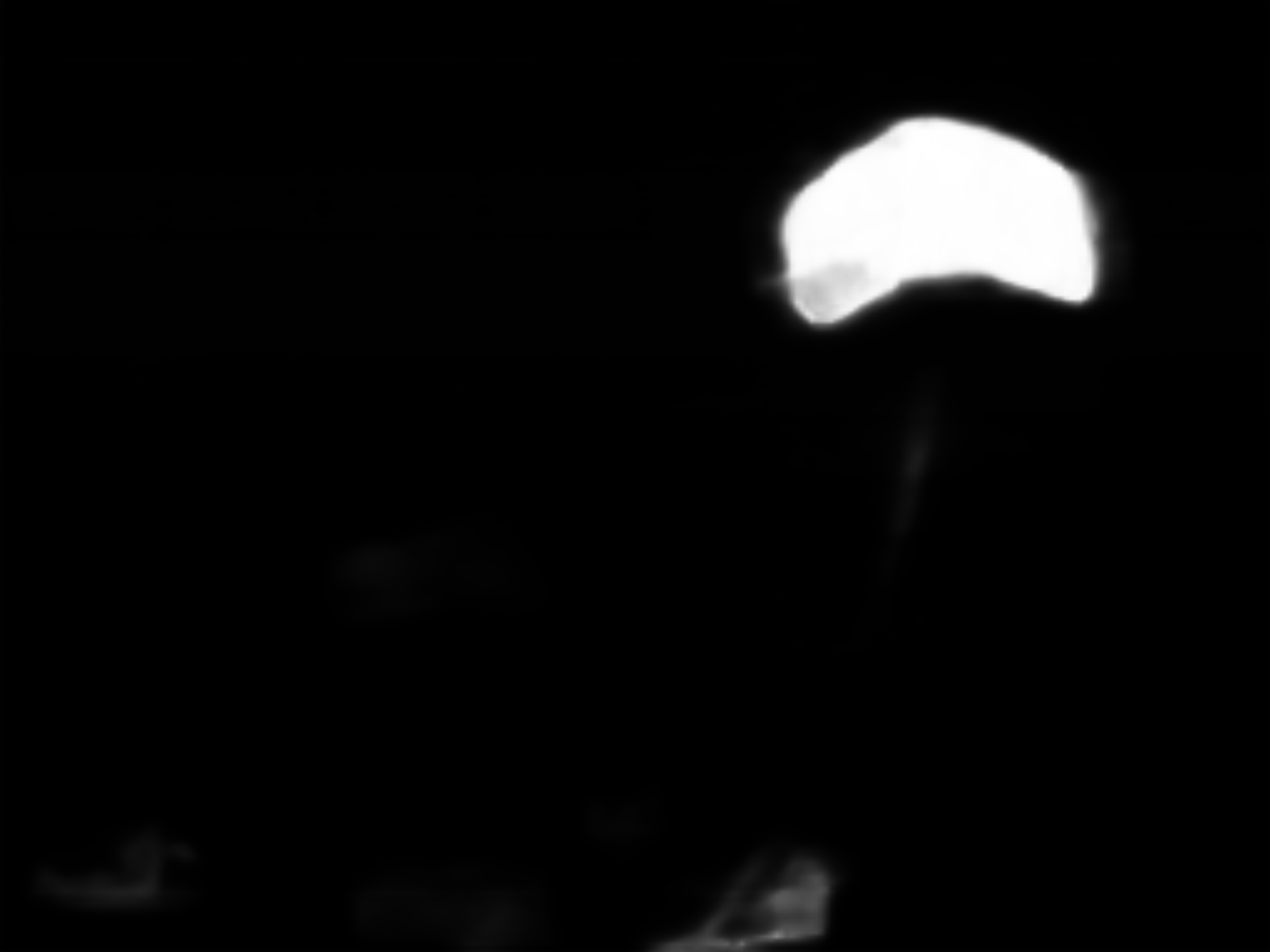}
	\end{minipage}
	\begin{minipage}[h]{0.18\linewidth}
		\centering
		\includegraphics[width=\linewidth]{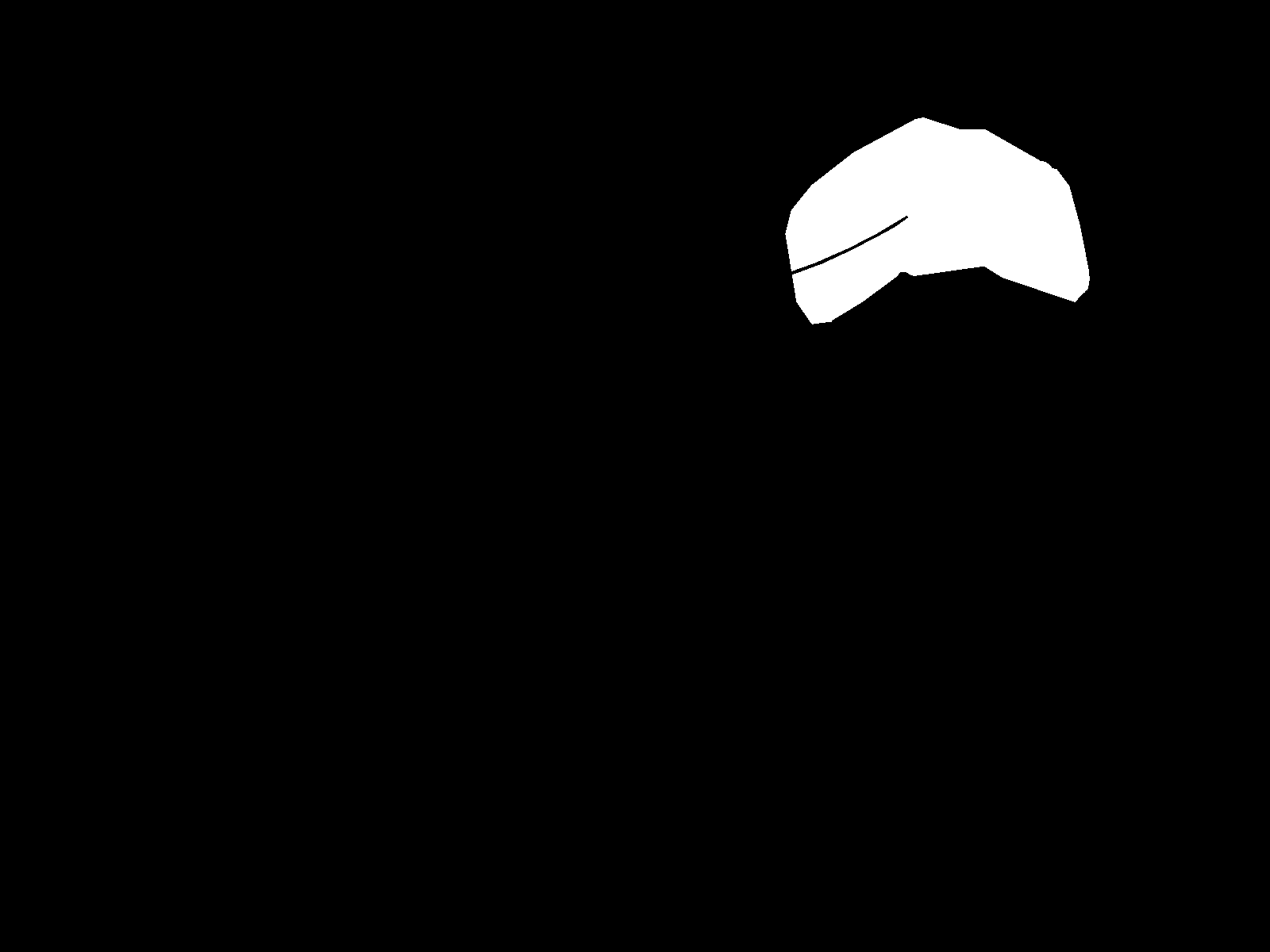}
	\end{minipage}
	\begin{minipage}[h]{0.18\linewidth}
		\centering
		\includegraphics[width=\linewidth]{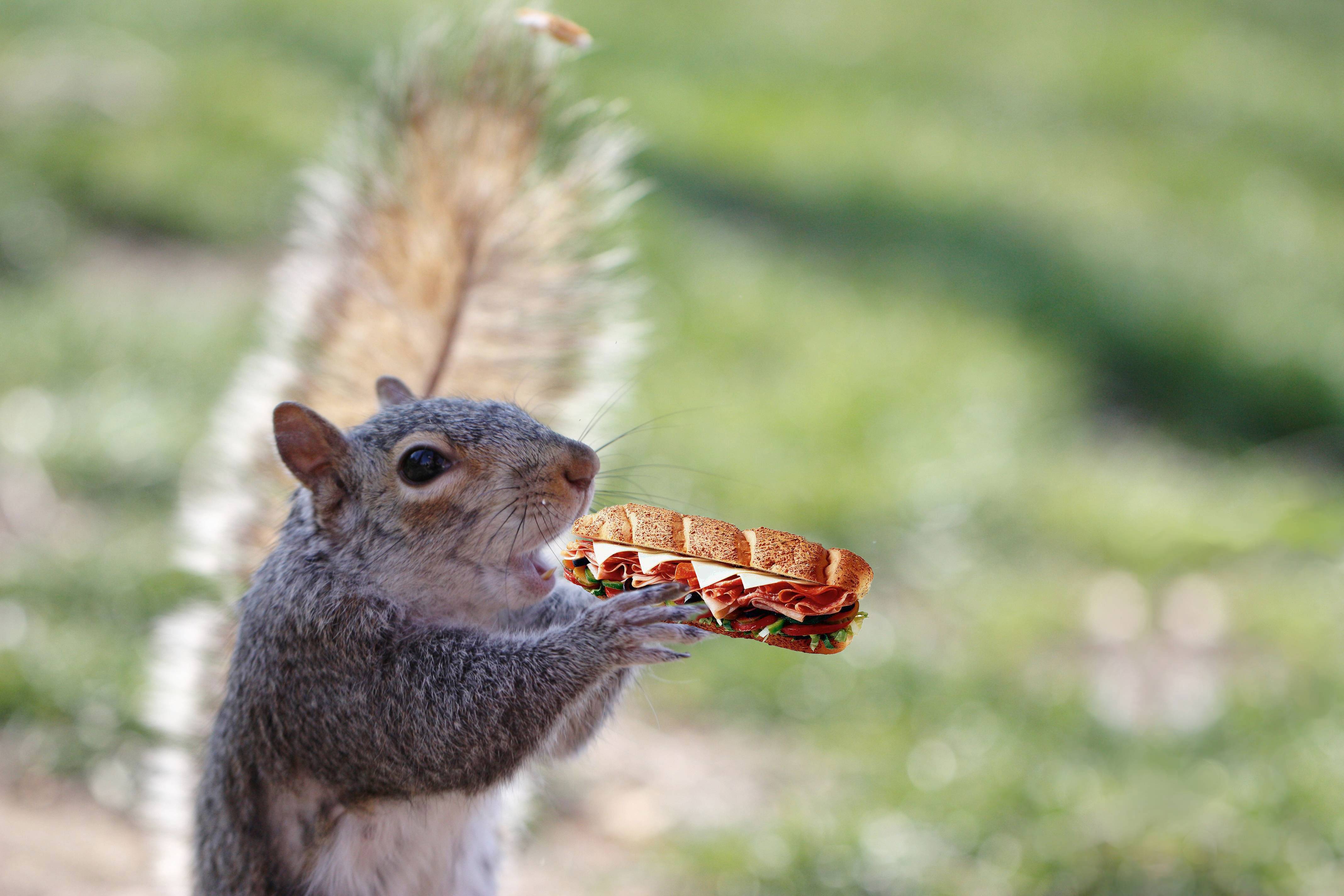}
	\end{minipage}
	\vspace{1mm}
	\begin{minipage}[h]{0.18\linewidth}
		\centering
		\includegraphics[width=\linewidth]{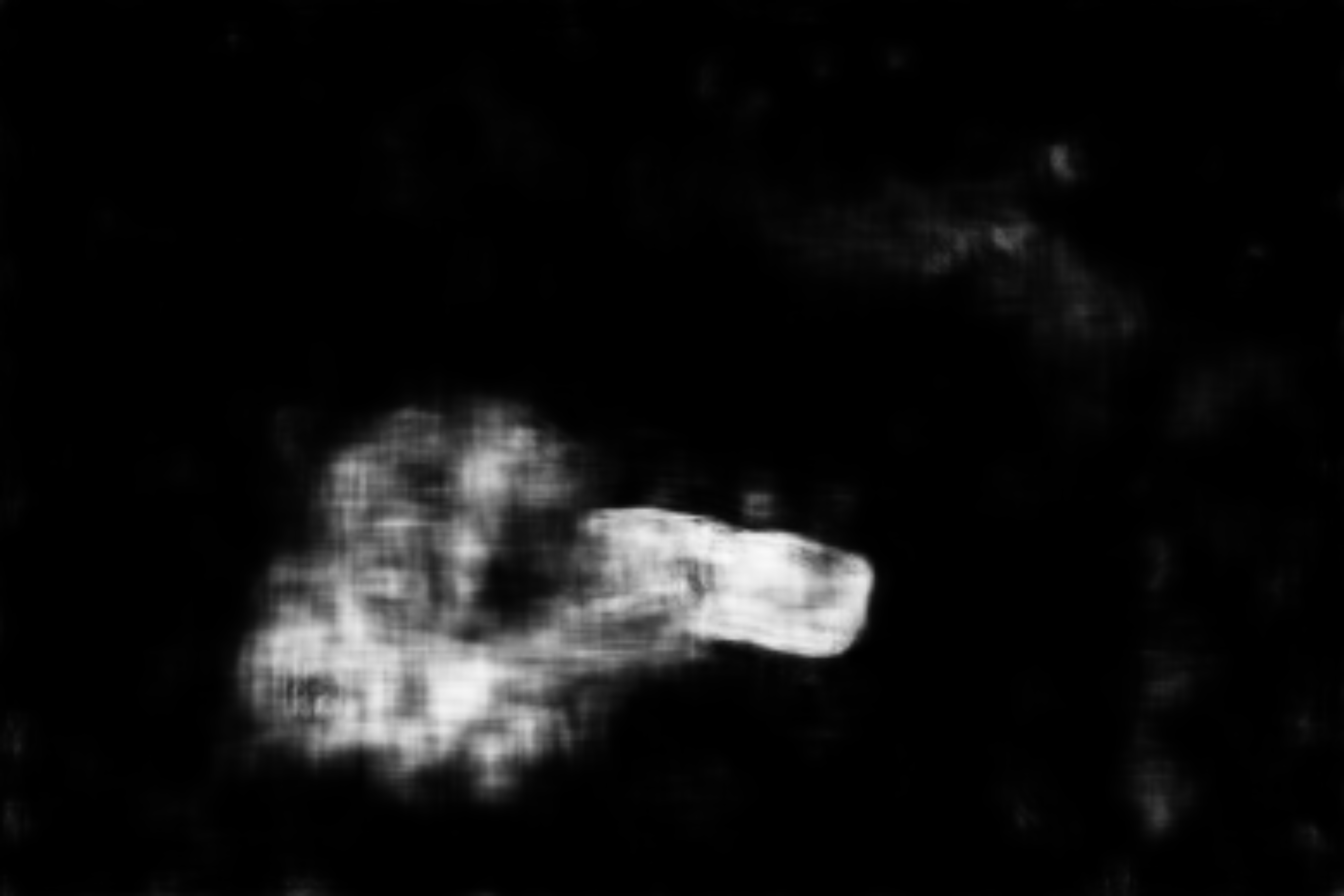}
	\end{minipage}
	\begin{minipage}[h]{0.18\linewidth}
		\centering
		\includegraphics[width=\linewidth]{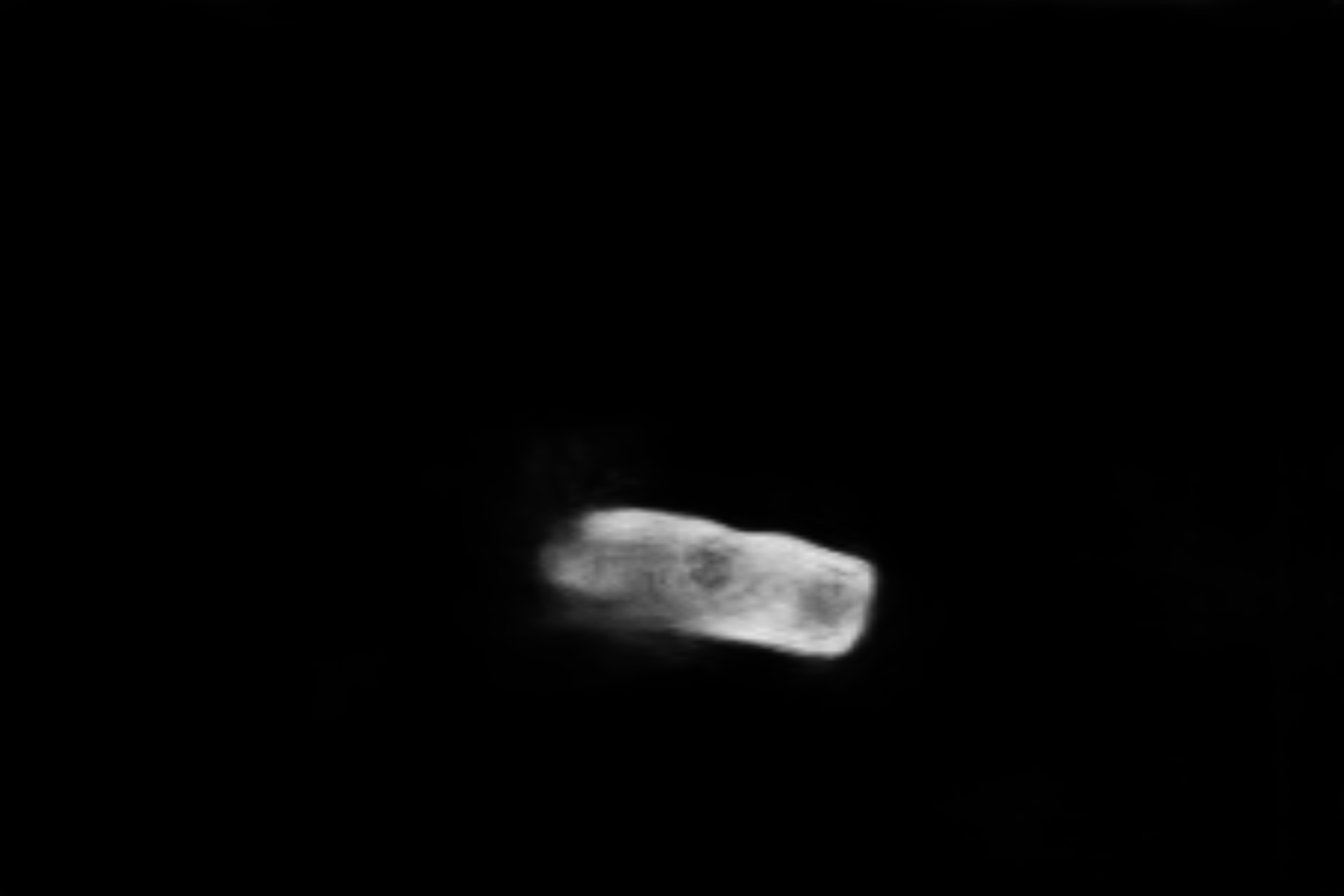}
	\end{minipage}
	\begin{minipage}[h]{0.18\linewidth}
		\centering
		\includegraphics[width=\linewidth]{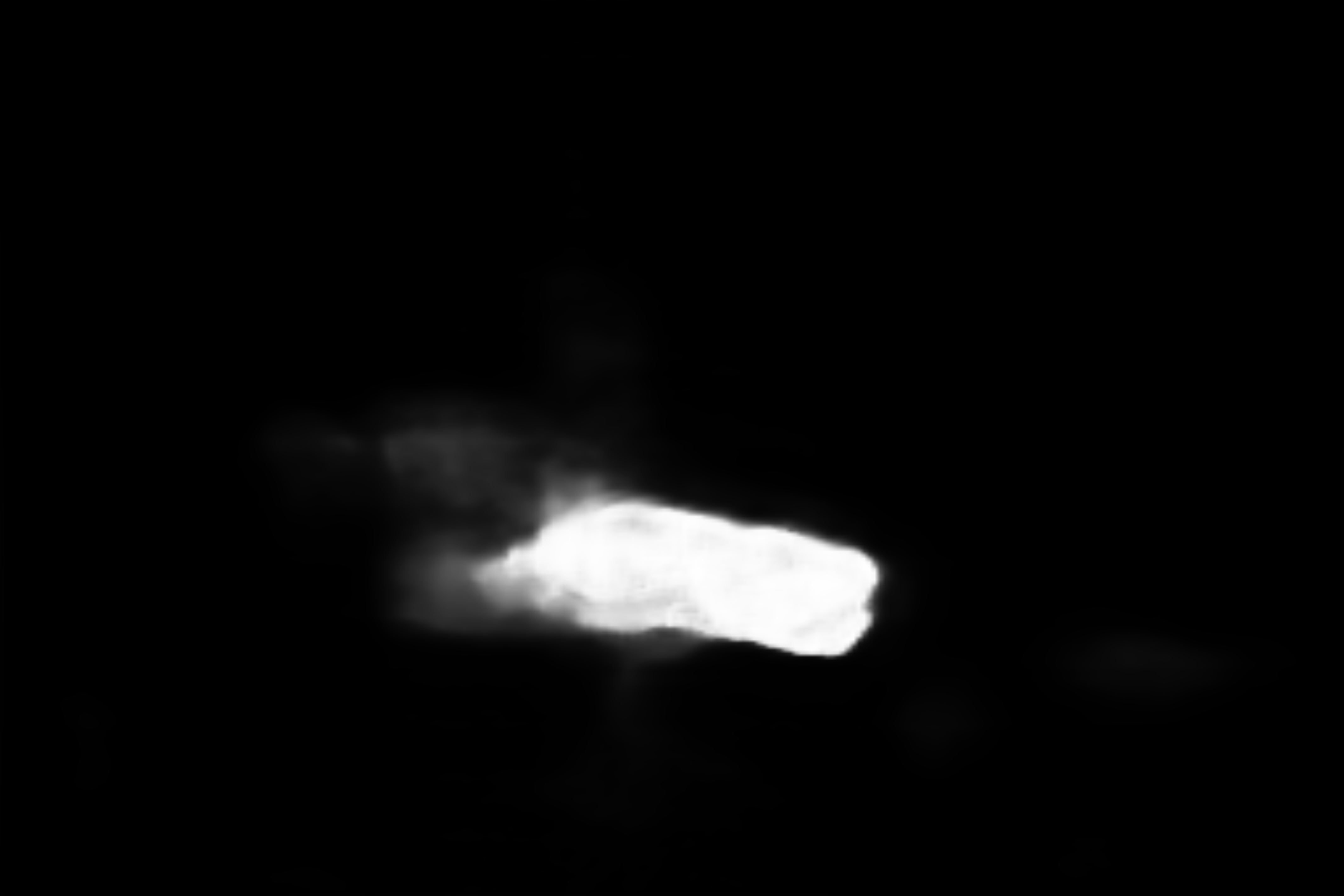}
	\end{minipage}
	\begin{minipage}[h]{0.18\linewidth}
		\centering
		\includegraphics[width=\linewidth]{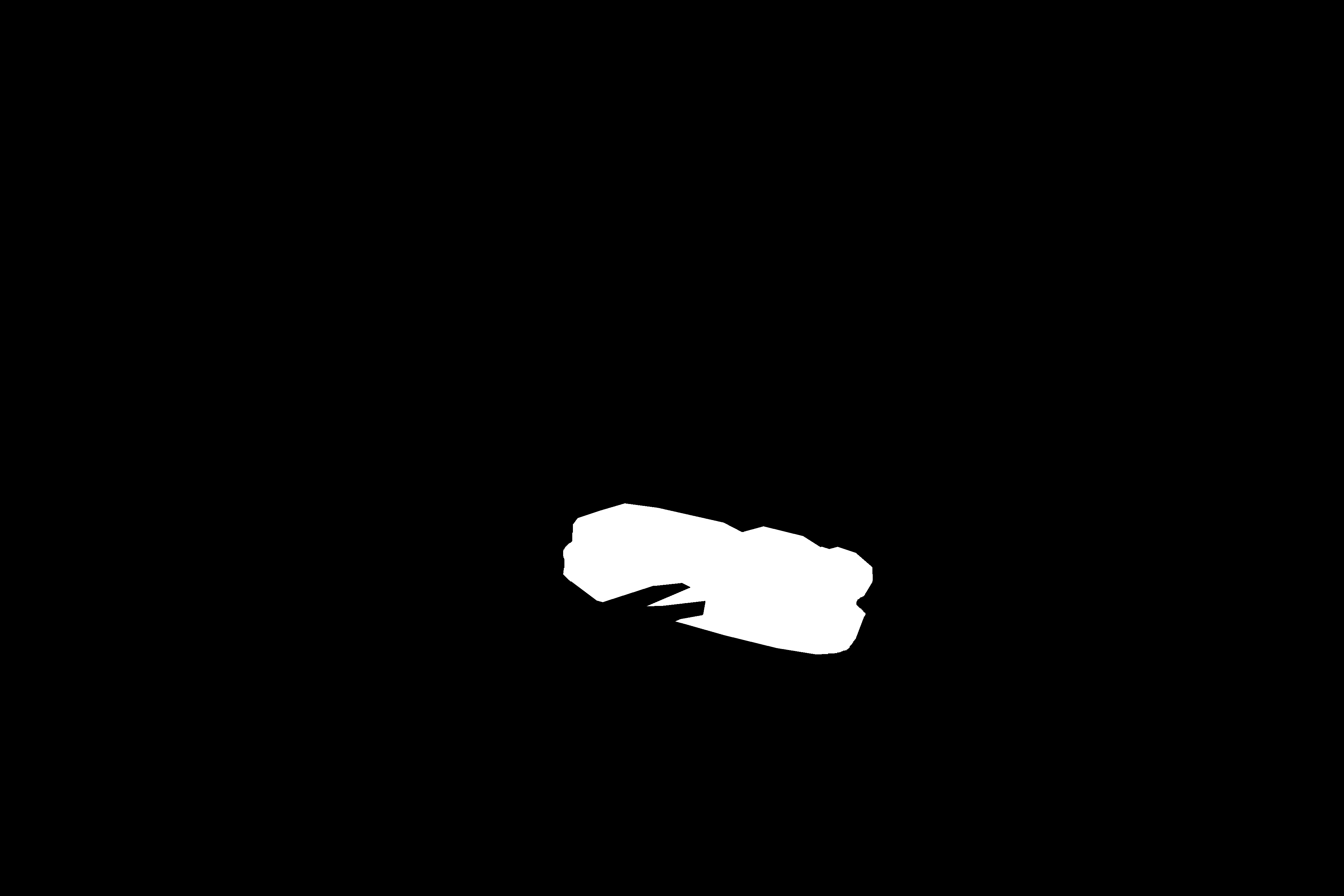}
	\end{minipage}
	\begin{minipage}[h]{0.18\linewidth}
	\centering
	\includegraphics[width=\linewidth]{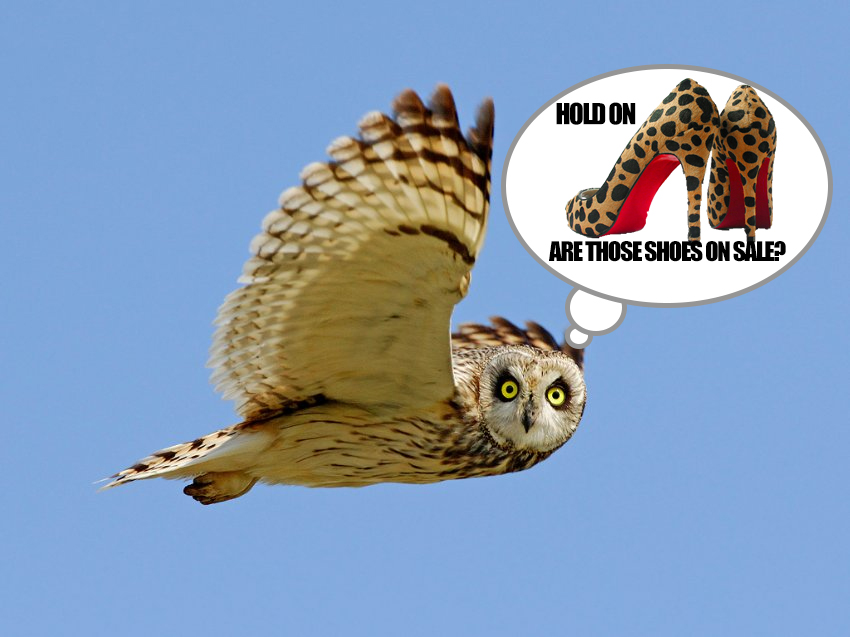}
	\scriptsize{(a) Manipulated}
	\end{minipage}
	\begin{minipage}[h]{0.18\linewidth}
		\centering
		\includegraphics[width=\linewidth]{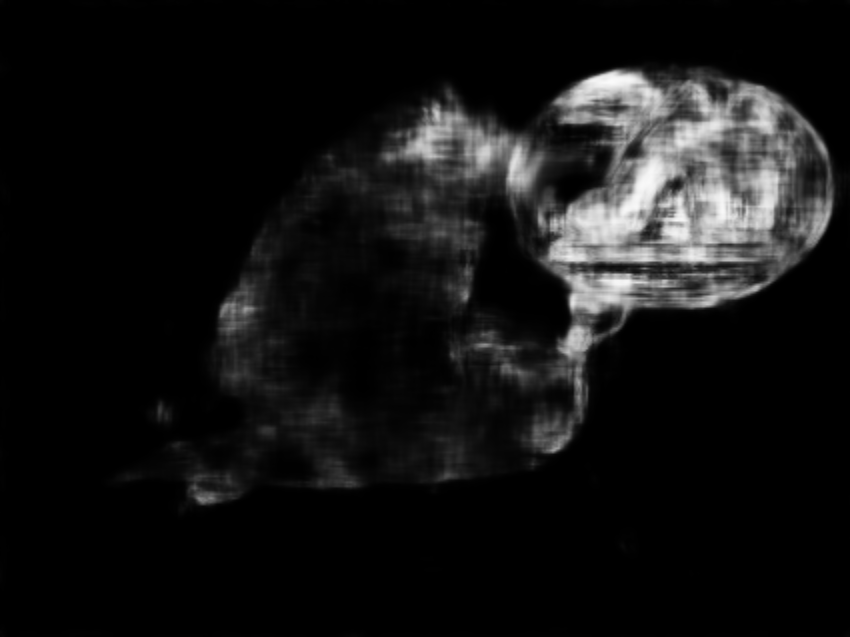}
	\scriptsize{(b) ManTra-Net}
	\end{minipage}
	\begin{minipage}[h]{0.18\linewidth}
		\centering
		\includegraphics[width=\linewidth]{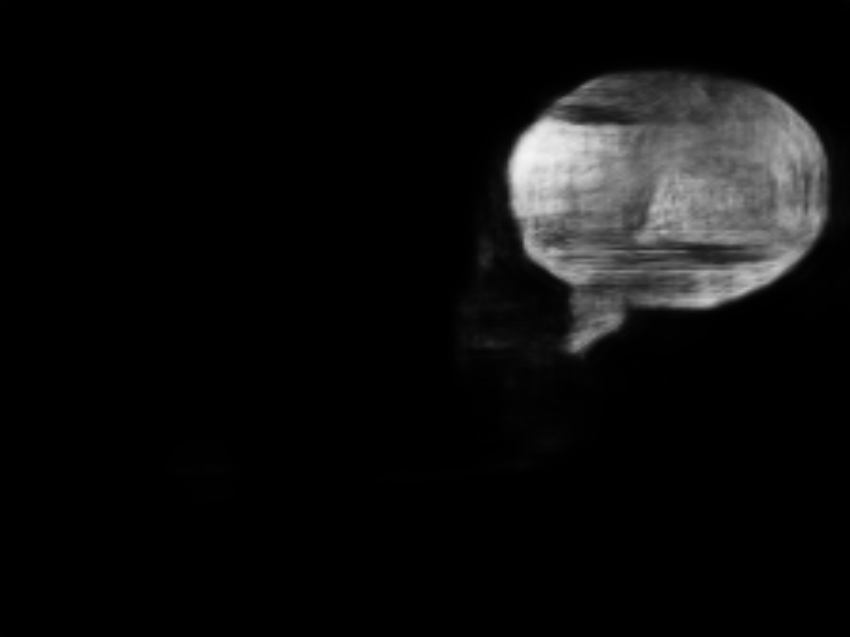}
	\scriptsize{(c) SPAN}
	\end{minipage}
	\begin{minipage}[h]{0.18\linewidth}
		\centering
		\includegraphics[width=\linewidth]{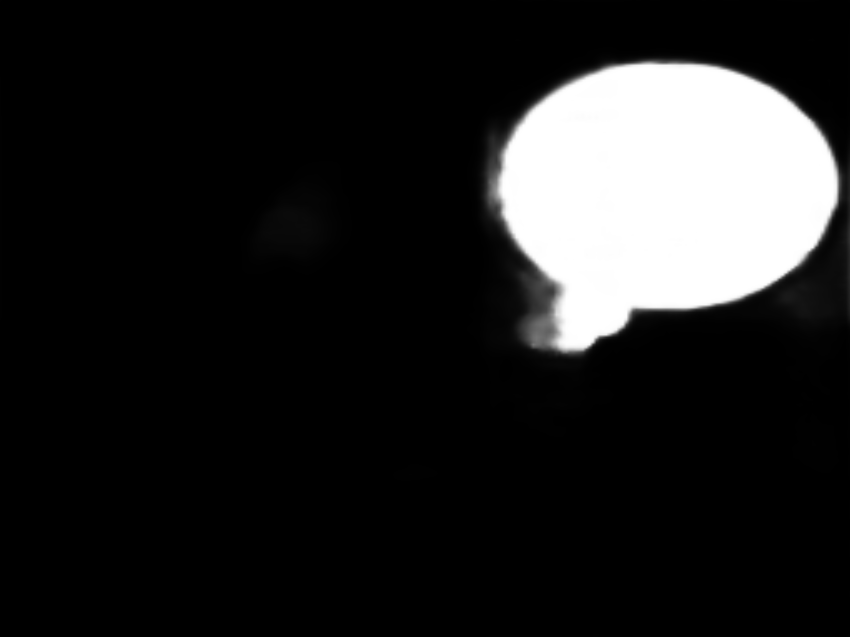}
	\scriptsize{(d) PSCC-Net}
	\end{minipage}
	\begin{minipage}[h]{0.18\linewidth}
		\centering
		\includegraphics[width=\linewidth]{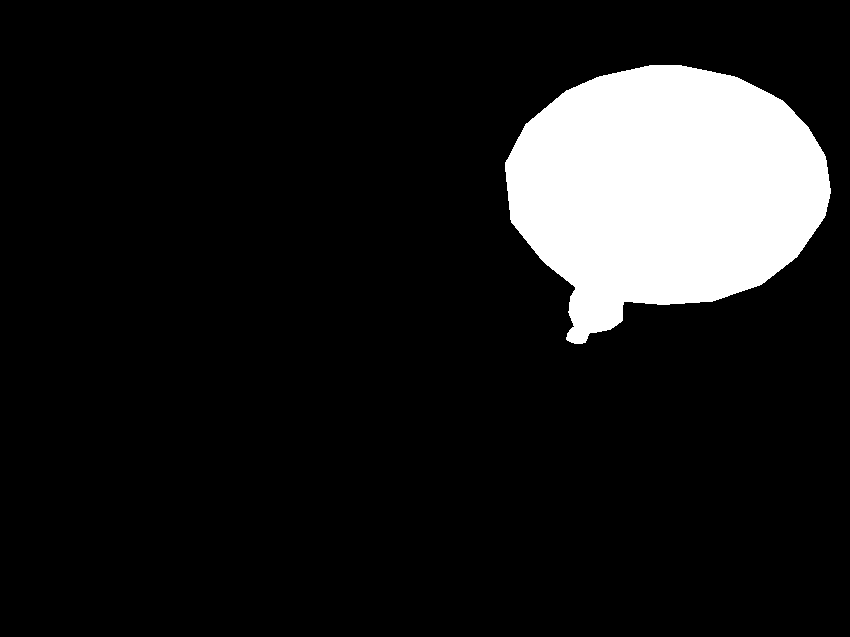}
	\scriptsize{(e) GT}
	\end{minipage}
	\caption{Qualitative localization evaluations on the IMD20 real-life dataset.}
	\label{fig:experiments_real}
\end{figure}

\subsubsection{Fine-tuned model}
The network weights of the pre-trained model are used to initiate the fine-tuned models that will be trained on the \textit{training split} of Coverage, CASIA, and NIST16 datasets, respectively.
The training strategy for fine-tuned models is the same as the one for pre-trained model, except setting the initial learning rate to $1e$-$4$. We evaluate the fine-tuned models of different methods in Tab.~\ref{table:finetune}. For AUC, PSCC-Net surpasses baselines in all cases (over $2.4\%$ to SPAN on average). As for F1 score, our model outperforms them with a large margin (over $16.6\%$ to SPAN on average). This validates the effectiveness of our overall network design.


\subsubsection{Qualitative comparisons}
We provide qualitative evaluations of manipulation localization on four standard test datasets and one real-life dataset shown in Fig.~\ref{fig:experiments} and Fig.~\ref{fig:experiments_real}, respectively, where the best available model for each method is used to produce manipulation masks. Compared to ManTra-Net~\cite{wu2019mantra} and SPAN~\cite{hu2020span}, the predicted masks from our PSCC-Net achieve the best performance in terms of higher prediction accuracy (\textit{e.g.}, the $1st$ row in Fig.~\ref{fig:experiments}) and fewer false alarms (\textit{e.g.}, the $6$th row in Fig.~\ref{fig:experiments}). In addition, the proposed method is less sensitive to the scale variation. Both large (\textit{e.g.}, the $5$th row in Fig.~\ref{fig:experiments}) and small (\textit{e.g.}, the $7$th row in Fig.~\ref{fig:experiments}) manipulations can be localized effectively. On the real-life dataset, PSCC-Net still performs much better than the other two (\textit{e.g.}, the $2$th row in Fig.~\ref{fig:experiments_real}), which demonstrates its good generalization ability.


\begin{table}[t!]
	\centering
	\caption{Detection evaluation on CASIA-D, all reported in $\%$.} 
	\renewcommand{\arraystretch}{1.4}
	\begin{adjustbox}{width=0.9\linewidth}
		\begin{tabu}{lllll}
			\toprule
			Method & AUC $\uparrow$ & F1 $\uparrow$ & EER $\downarrow$ & TPR$_{1\%}$ $\uparrow$\\ \hline
			ManTra-Net~\cite{wu2019mantra} & $59.94$ & $56.69$ & $43.21$ & $5.43$\\ 
			SPAN~\cite{hu2020span} & $67.33$ & $63.48$ & $36.47$ & $ 5.54 $\\ 
			PSCC-Net$^\dagger$ & $74.40$ & $66.88$ & $33.21$ & $28.37$\\
			PSCC-Net & $\mathbf{99.65}$ & $\mathbf{97.12}$ & $\mathbf{2.83}$ & $\mathbf{95.65}$\\
			\bottomrule
		\end{tabu}
	\end{adjustbox}
	\label{table:detection}
\end{table}

\begin{figure}[t!]
	\small
	\centering
	\includegraphics[width=0.9\linewidth]{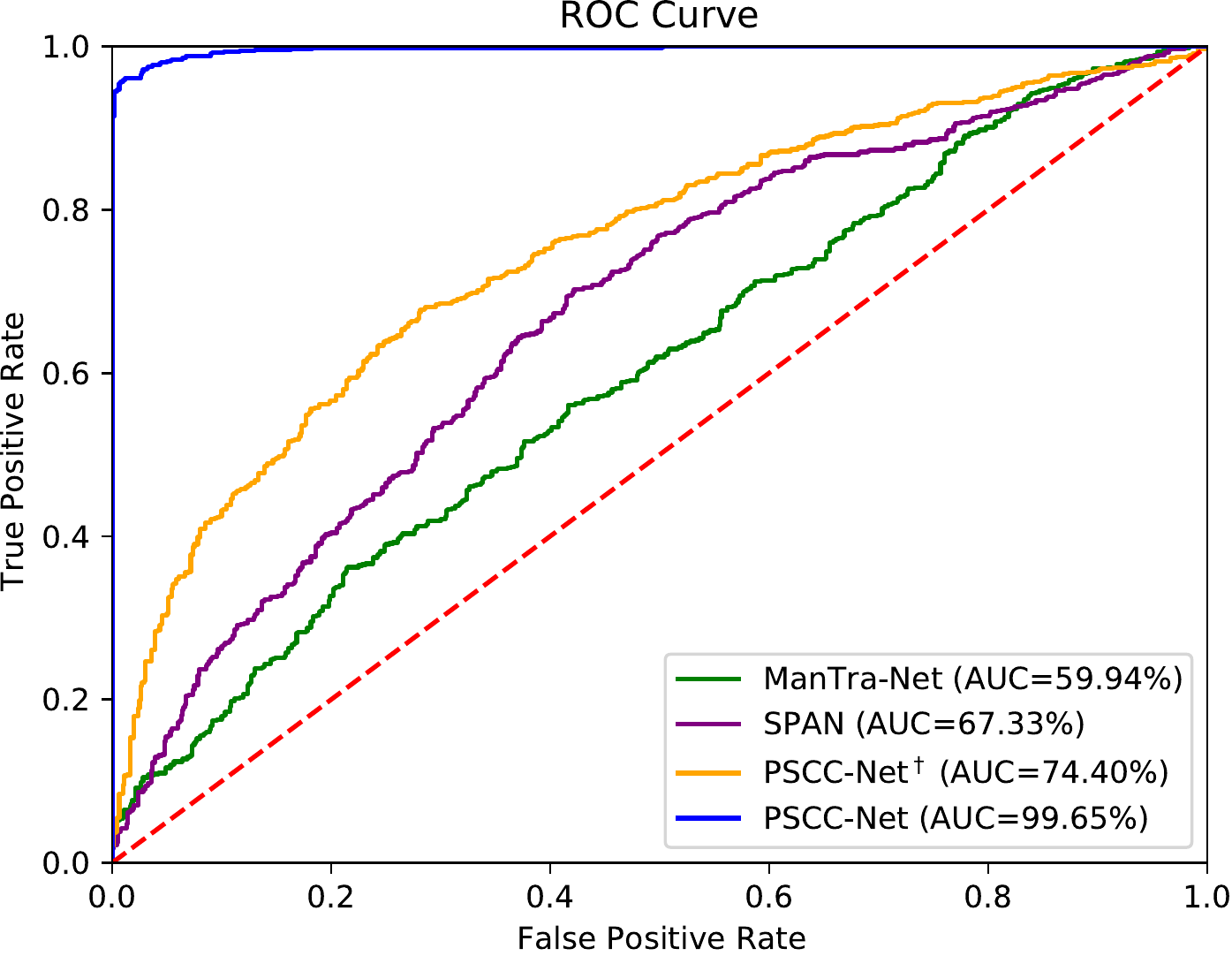}
	\caption{ROC of different methods for detection. Our detection head successfully alleviates the influence of false alarms in pristine images, thus achieves the best result.}
	\label{fig:roc}
\end{figure}
\begin{figure*}[t]
	\centering
	\begin{minipage}[h]{0.137\linewidth}
		\centering
		\includegraphics[width=\linewidth]{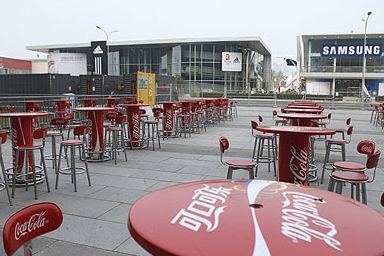}
	\end{minipage}
	\vspace{1mm}
	\begin{minipage}[h]{0.137\linewidth}
		\centering
		\includegraphics[width=\linewidth]{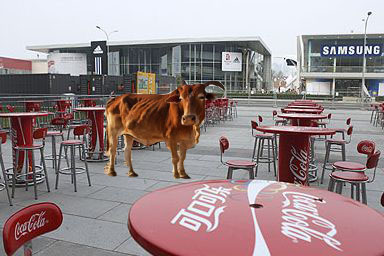}
	\end{minipage}
	\begin{minipage}[h]{0.137\linewidth}
		\centering
		\includegraphics[width=\linewidth]{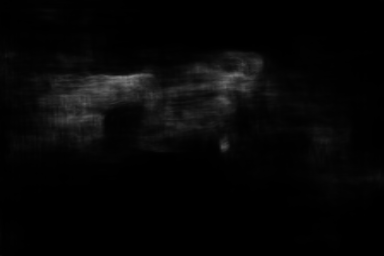}
	\end{minipage}
	\begin{minipage}[h]{0.137\linewidth}
		\centering
		\includegraphics[width=\linewidth]{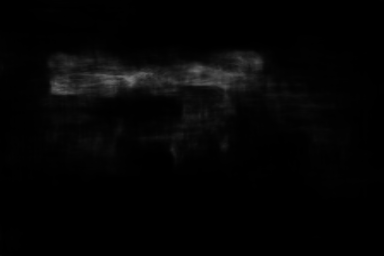}
	\end{minipage}
	\begin{minipage}[h]{0.137\linewidth}
		\centering
		\includegraphics[width=\linewidth]{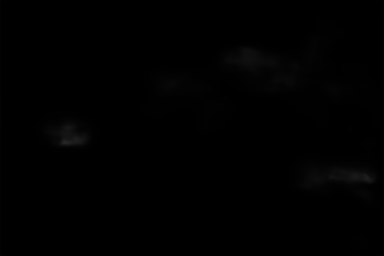}
	\end{minipage}
	\begin{minipage}[h]{0.137\linewidth}
		\centering
		\includegraphics[width=\linewidth]{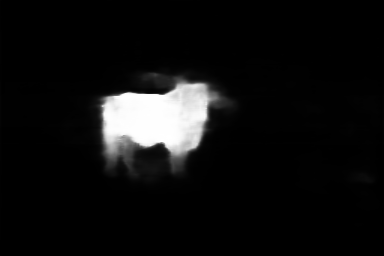}
	\end{minipage}
	\begin{minipage}[h]{0.137\linewidth}
		\centering
		\includegraphics[width=\linewidth]{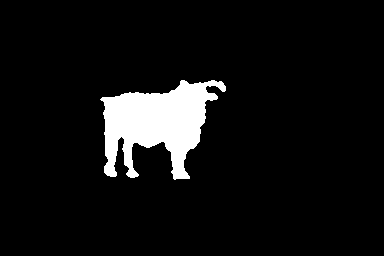}
	\end{minipage}
	\begin{minipage}[h]{0.137\linewidth}
		\centering
		\includegraphics[width=\linewidth]{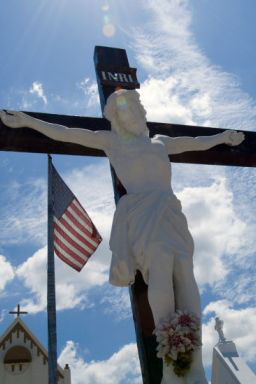}
	\end{minipage}
	\vspace{1mm}
	\begin{minipage}[h]{0.137\linewidth}
		\centering
		\includegraphics[width=\linewidth]{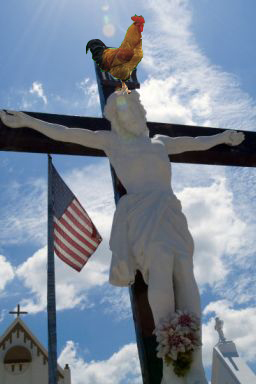}
	\end{minipage}
	\begin{minipage}[h]{0.137\linewidth}
		\centering
		\includegraphics[width=\linewidth]{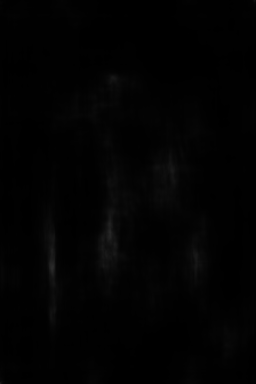}
	\end{minipage}
	\begin{minipage}[h]{0.137\linewidth}
		\centering
		\includegraphics[width=\linewidth]{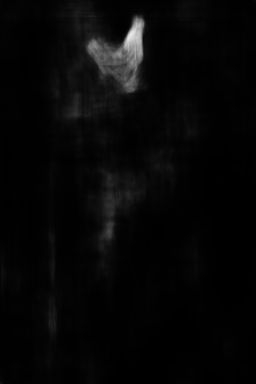}
	\end{minipage}
	\begin{minipage}[h]{0.137\linewidth}
		\centering
		\includegraphics[width=\linewidth]{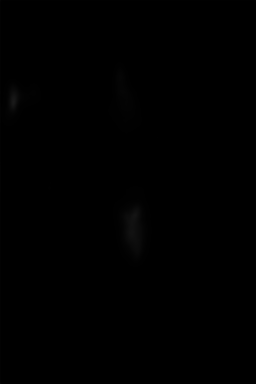}
	\end{minipage}
	\begin{minipage}[h]{0.137\linewidth}
		\centering
		\includegraphics[width=\linewidth]{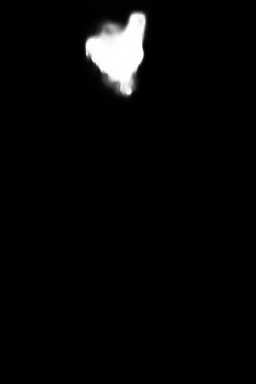}
	\end{minipage}
	\begin{minipage}[h]{0.137\linewidth}
		\centering
		\includegraphics[width=\linewidth]{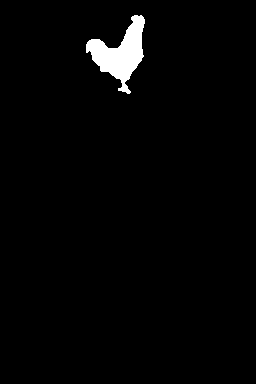}
	\end{minipage}
	\begin{minipage}[h]{0.137\linewidth}
		\centering
		\includegraphics[width=\linewidth]{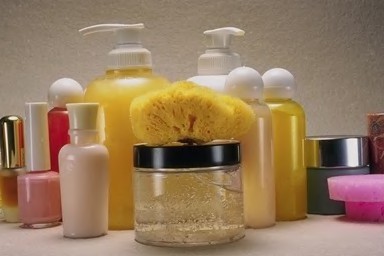}
	\end{minipage}
	\vspace{1mm}
	\begin{minipage}[h]{0.137\linewidth}
		\centering
		\includegraphics[width=\linewidth]{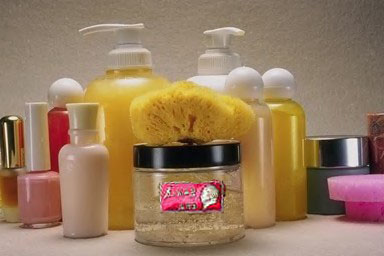}
	\end{minipage}
	\begin{minipage}[h]{0.137\linewidth}
		\centering
		\includegraphics[width=\linewidth]{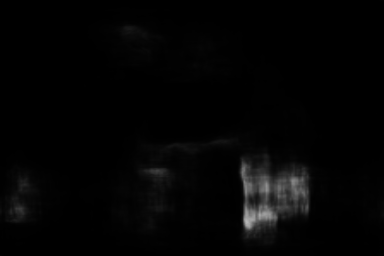}
	\end{minipage}
	\begin{minipage}[h]{0.137\linewidth}
		\centering
		\includegraphics[width=\linewidth]{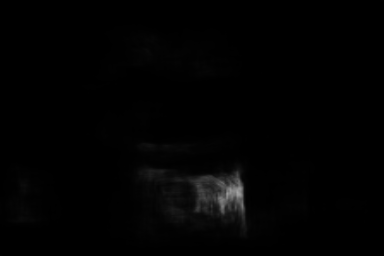}
	\end{minipage}
	\begin{minipage}[h]{0.137\linewidth}
		\centering
		\includegraphics[width=\linewidth]{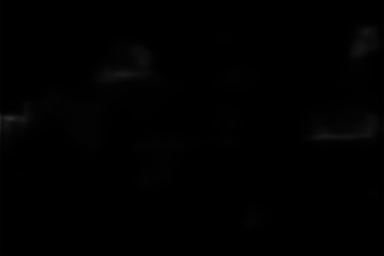}
	\end{minipage}
	\begin{minipage}[h]{0.137\linewidth}
		\centering
		\includegraphics[width=\linewidth]{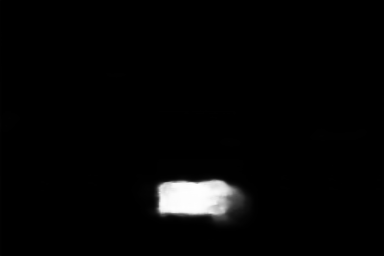}
	\end{minipage}
	\begin{minipage}[h]{0.137\linewidth}
		\centering
		\includegraphics[width=\linewidth]{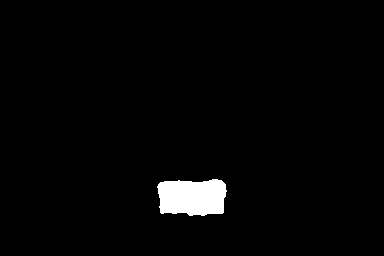}
	\end{minipage}
	\begin{minipage}[h]{0.137\linewidth}
		\centering
		\includegraphics[width=\linewidth]{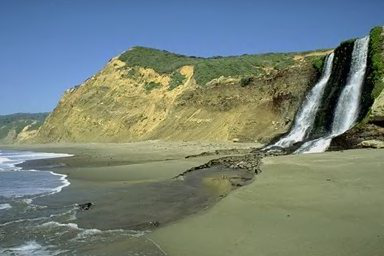}
	\end{minipage}
	\vspace{1mm}
	\begin{minipage}[h]{0.137\linewidth}
		\centering
		\includegraphics[width=\linewidth]{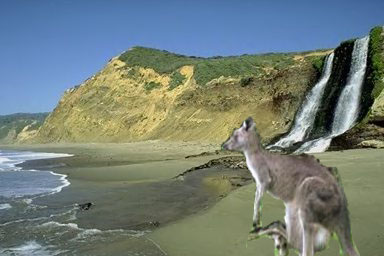}
	\end{minipage}
	\begin{minipage}[h]{0.137\linewidth}
		\centering
		\includegraphics[width=\linewidth]{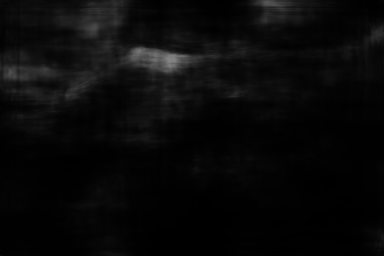}
	\end{minipage}
	\begin{minipage}[h]{0.137\linewidth}
		\centering
		\includegraphics[width=\linewidth]{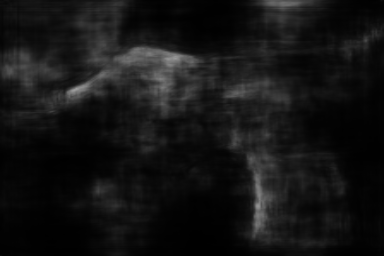}
	\end{minipage}
	\begin{minipage}[h]{0.137\linewidth}
		\centering
		\includegraphics[width=\linewidth]{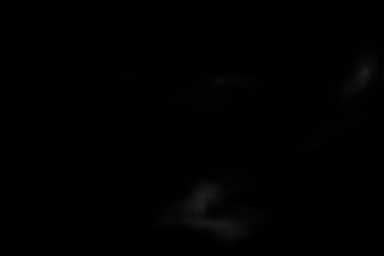}
	\end{minipage}
	\begin{minipage}[h]{0.137\linewidth}
		\centering
		\includegraphics[width=\linewidth]{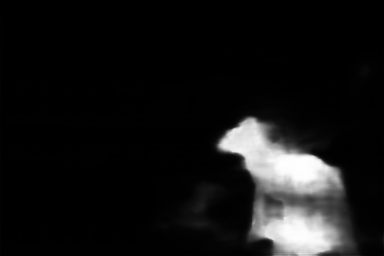}
	\end{minipage}
	\begin{minipage}[h]{0.137\linewidth}
		\centering
		\includegraphics[width=\linewidth]{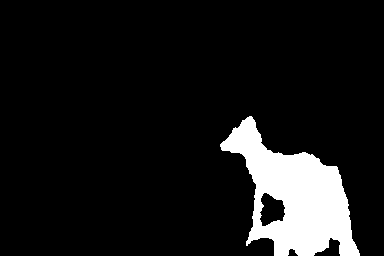}
	\end{minipage}
	\begin{minipage}[h]{0.137\linewidth}
		\centering
		\includegraphics[width=\linewidth]{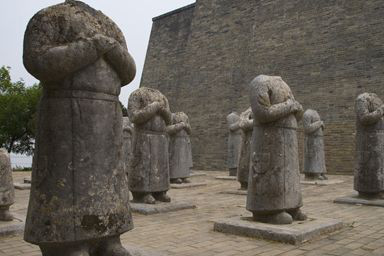}
	\end{minipage}
	\vspace{1mm}
	\begin{minipage}[h]{0.137\linewidth}
		\centering
		\includegraphics[width=\linewidth]{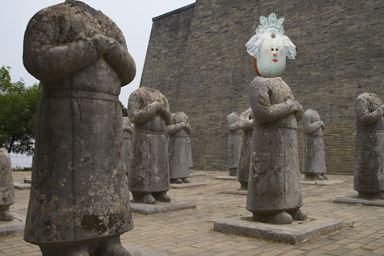}
	\end{minipage}
	\begin{minipage}[h]{0.137\linewidth}
		\centering
		\includegraphics[width=\linewidth]{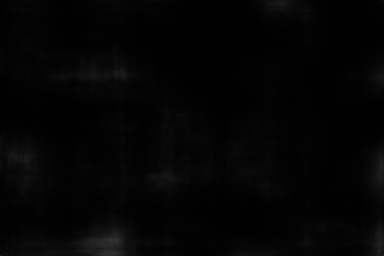}
	\end{minipage}
	\begin{minipage}[h]{0.137\linewidth}
		\centering
		\includegraphics[width=\linewidth]{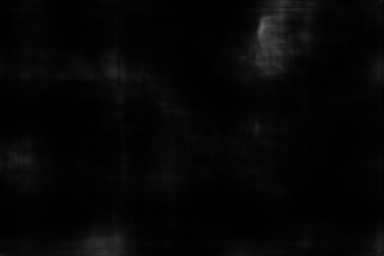}
	\end{minipage}
	\begin{minipage}[h]{0.137\linewidth}
		\centering
		\includegraphics[width=\linewidth]{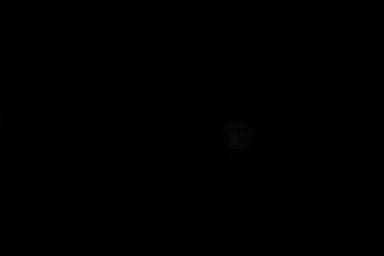}
	\end{minipage}
	\begin{minipage}[h]{0.137\linewidth}
		\centering
		\includegraphics[width=\linewidth]{detection/6/Sp_D_NRN_C_art0051_art0012_0565_ours.png}
	\end{minipage}
	\begin{minipage}[h]{0.137\linewidth}
		\centering
		\includegraphics[width=\linewidth]{detection/6/Sp_D_NRN_C_art0051_art0012_0565_gt.png}
	\end{minipage}
	\begin{minipage}[h]{0.137\linewidth}
		\centering
		\includegraphics[width=\linewidth]{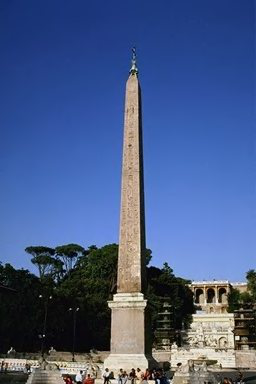}
		\scriptsize{(a) Pristine image}
	\end{minipage}
	\begin{minipage}[h]{0.137\linewidth}
		\centering
		\includegraphics[width=\linewidth]{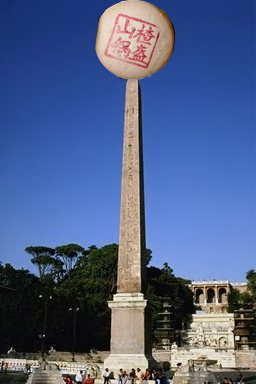}
		\scriptsize{(b) Manipulated image}
	\end{minipage}
	\begin{minipage}[h]{0.137\linewidth}
		\centering
		\includegraphics[width=\linewidth]{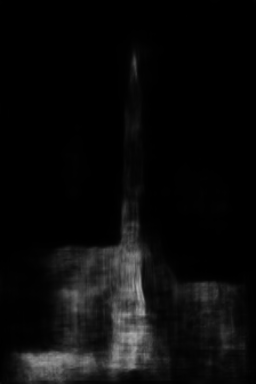}
		\scriptsize{(c) Pristine mask~\cite{hu2020span}}
	\end{minipage}
	\begin{minipage}[h]{0.137\linewidth}
		\centering
		\includegraphics[width=\linewidth]{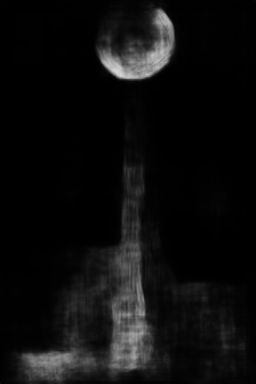}
		\scriptsize{(d) Manip. mask~\cite{hu2020span}}
	\end{minipage}
	\begin{minipage}[h]{0.137\linewidth}
		\centering
		\includegraphics[width=\linewidth]{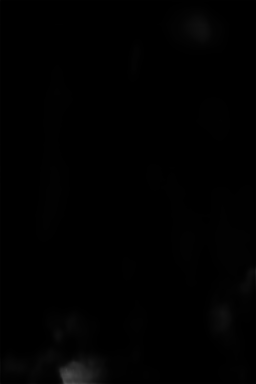}
		\scriptsize{(e) Our pristine mask}
	\end{minipage}
	\begin{minipage}[h]{0.137\linewidth}
		\centering
		\includegraphics[width=\linewidth]{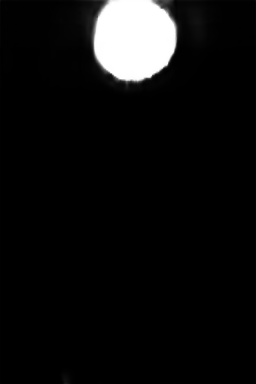}
		\scriptsize{(f) Our manip. mask}
	\end{minipage}
	\begin{minipage}[h]{0.137\linewidth}
		\centering
		\includegraphics[width=\linewidth]{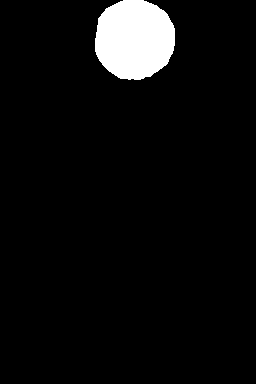}
		\scriptsize{(g) GT manip. mask}
	\end{minipage}
	
	\caption{Qualitative detection evaluations on CASIA-D. Since GT pristine masks are blank, they are not shown here for clarity.}
	\label{fig:detection}
\end{figure*}

\subsection{Comparisons on Detection}
 Since ManTra-Net and SPAN are the best performing baselines in the localization evaluation, and ManTra-Net does not develop the fine-tuned model,
 we choose to use the pre-trained model for detection evaluation, in order to make comparisons to both of them.
Although these two baselines make no direct attempt to perform detection, their estimated manipulation masks can be leveraged for this purpose. 
As such, we simply regard the average of the mask  as their scores. For fair comparisons, we build a variant that adopts the same averaging strategy to calculate this score, denoted as PSCC-Net$^\dagger$. In Tab.~\ref{table:detection}, owing to our well-predicted manipulation masks, the PSCC-Net$^\dagger$ achieves the best detection performance on all used metrics. Moreover, we depict the corresponding Receiver Operating Characteristic (ROC) curve in Fig.~\ref{fig:roc}. It is evident that the detection performance can be dramatically improved by introducing a tailored head. With a favorable detection, the IMDL methods can be more efficient. 
That is, detection is performed before localization, and only the detected forgery is passed for localization. 
Our network design is compatible with this efficiency consideration as the detection head is placed at the beginning of the bottom-up path. The qualitative evaluations of manipulation detection are demonstrated in Fig.~\ref{fig:detection}, where the predicted masks from SPAN~\cite{hu2020span} and our PSCC-Net on both pristine and manipulation images are compared. Without the \textit{existence-of-manipulation} assumption, for pristine images, the corresponding predicted masks from our PSCC-Net are nearly blank. However, the ones from SPAN suffer severe false alarms in most cases. As for the relevant manipulated images, the proposed method localizes the forged regions more accurately.

\begin{figure*}[t]
	\centering
	\begin{minipage}[h]{0.168\linewidth}
		\centering
		\includegraphics[width=\linewidth]{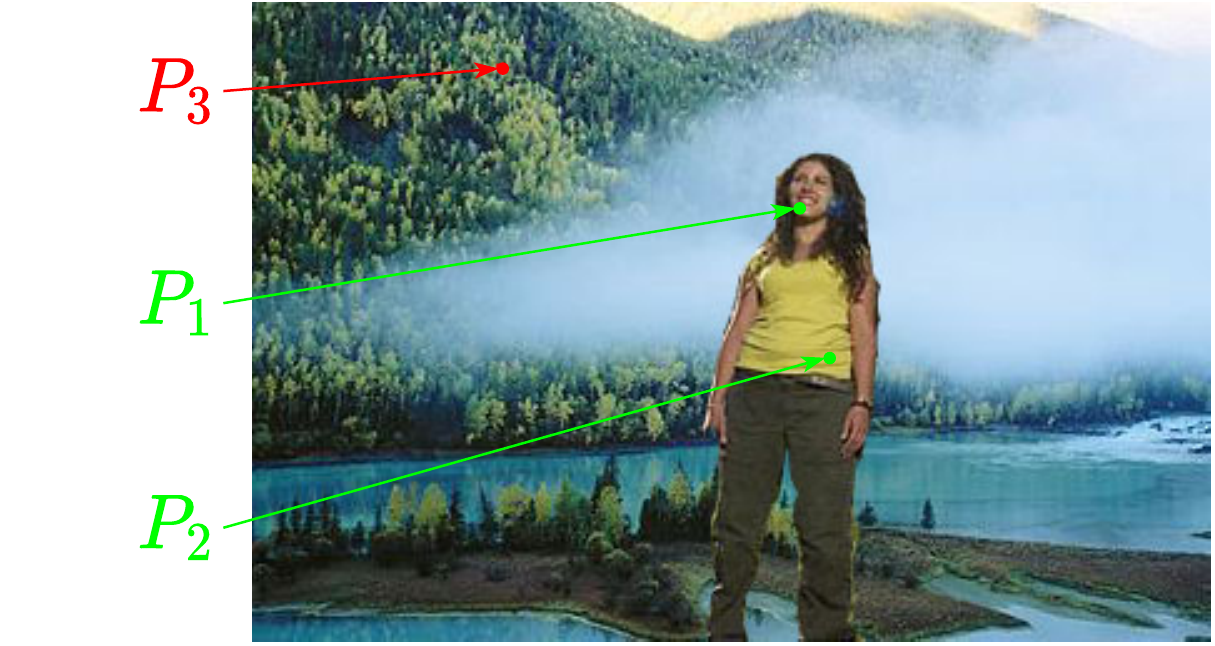}
	\end{minipage}
	\vspace{1mm}
	\begin{minipage}[h]{0.132\linewidth}
		\centering
		\includegraphics[width=\linewidth]{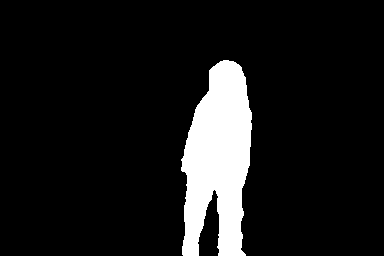}
	\end{minipage}
	\begin{minipage}[h]{0.132\linewidth}
		\centering
		\includegraphics[width=\linewidth]{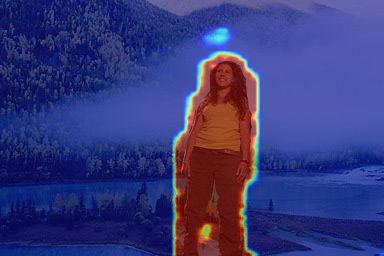}
	\end{minipage}
	\begin{minipage}[h]{0.132\linewidth}
		\centering
		\includegraphics[width=\linewidth]{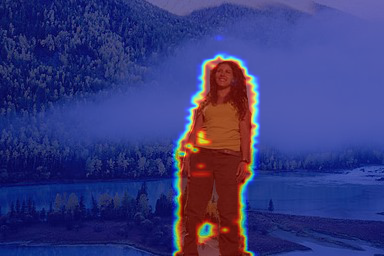}
	\end{minipage}
	\begin{minipage}[h]{0.132\linewidth}
		\centering
		\includegraphics[width=\linewidth]{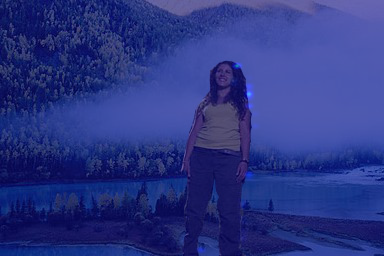}
	\end{minipage}
	\begin{minipage}[h]{0.132\linewidth}
		\centering
		\includegraphics[width=\linewidth]{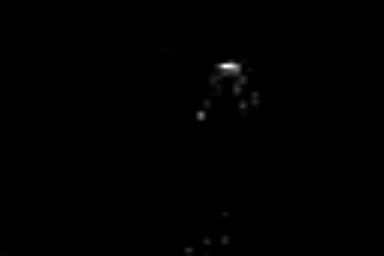}
	\end{minipage}
	\begin{minipage}[h]{0.132\linewidth}
		\centering
		\includegraphics[width=\linewidth]{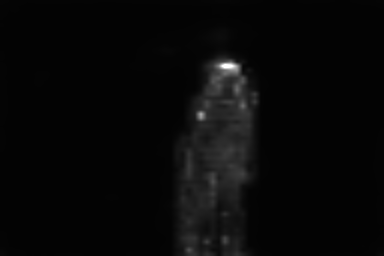}
	\end{minipage}
	\begin{minipage}[h]{0.168\linewidth}
		\centering
		\includegraphics[width=\linewidth]{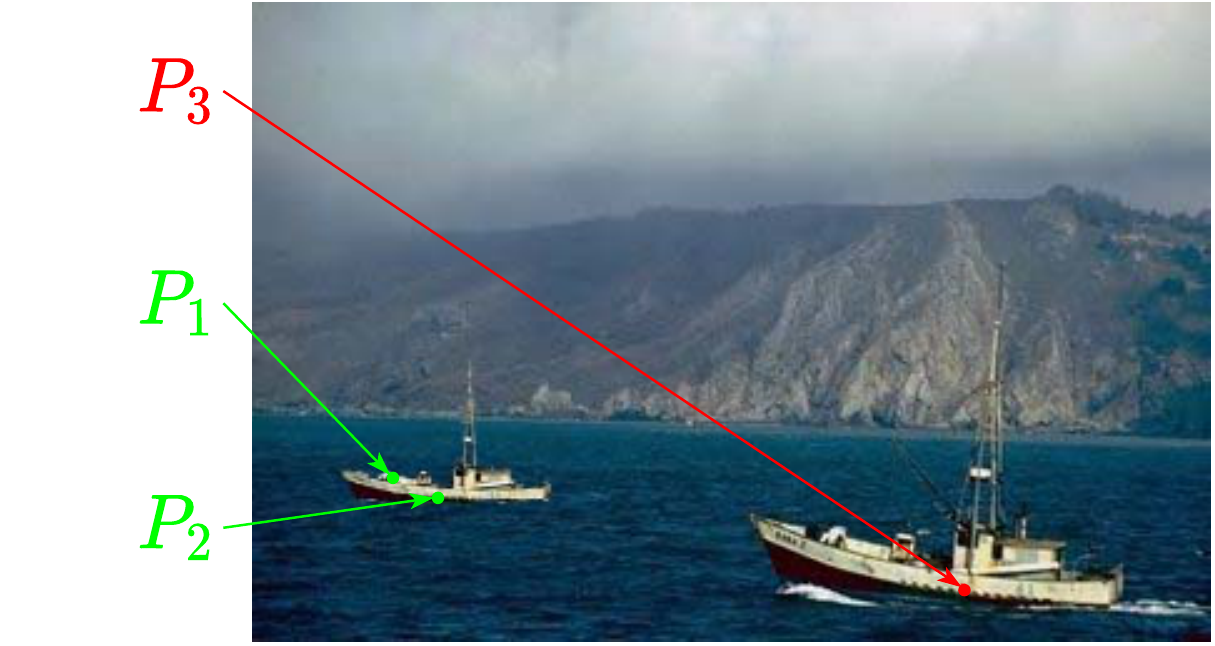}
	\end{minipage}
	\vspace{1mm}
	\begin{minipage}[h]{0.132\linewidth}
		\centering
		\includegraphics[width=\linewidth]{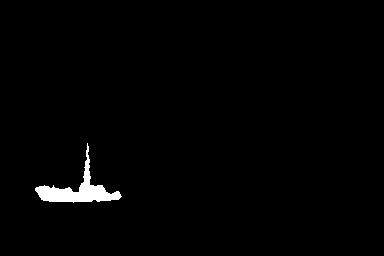}
	\end{minipage}
	\begin{minipage}[h]{0.132\linewidth}
		\centering
		\includegraphics[width=\linewidth]{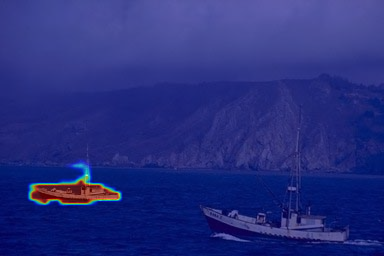}
	\end{minipage}
	\begin{minipage}[h]{0.132\linewidth}
		\centering
		\includegraphics[width=\linewidth]{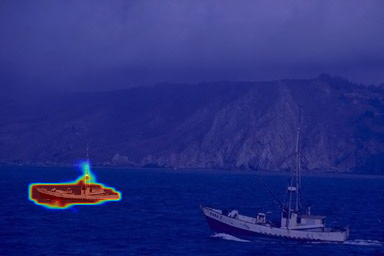}
	\end{minipage}
	\begin{minipage}[h]{0.132\linewidth}
		\centering
		\includegraphics[width=\linewidth]{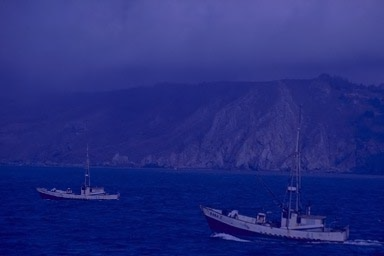}
	\end{minipage}
	\begin{minipage}[h]{0.132\linewidth}
		\centering
		\includegraphics[width=\linewidth]{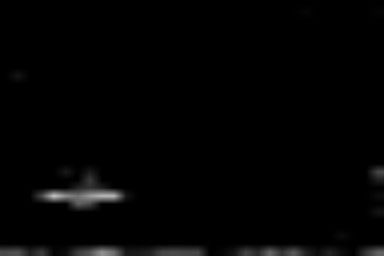}
	\end{minipage}
	\begin{minipage}[h]{0.132\linewidth}
		\centering
		\includegraphics[width=\linewidth]{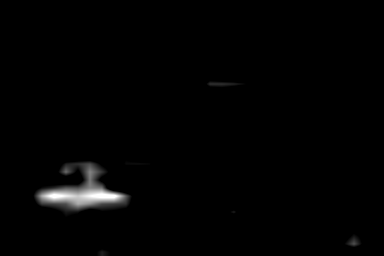}
	\end{minipage}
	\begin{minipage}[h]{0.168\linewidth}
		\centering
		\includegraphics[width=\linewidth]{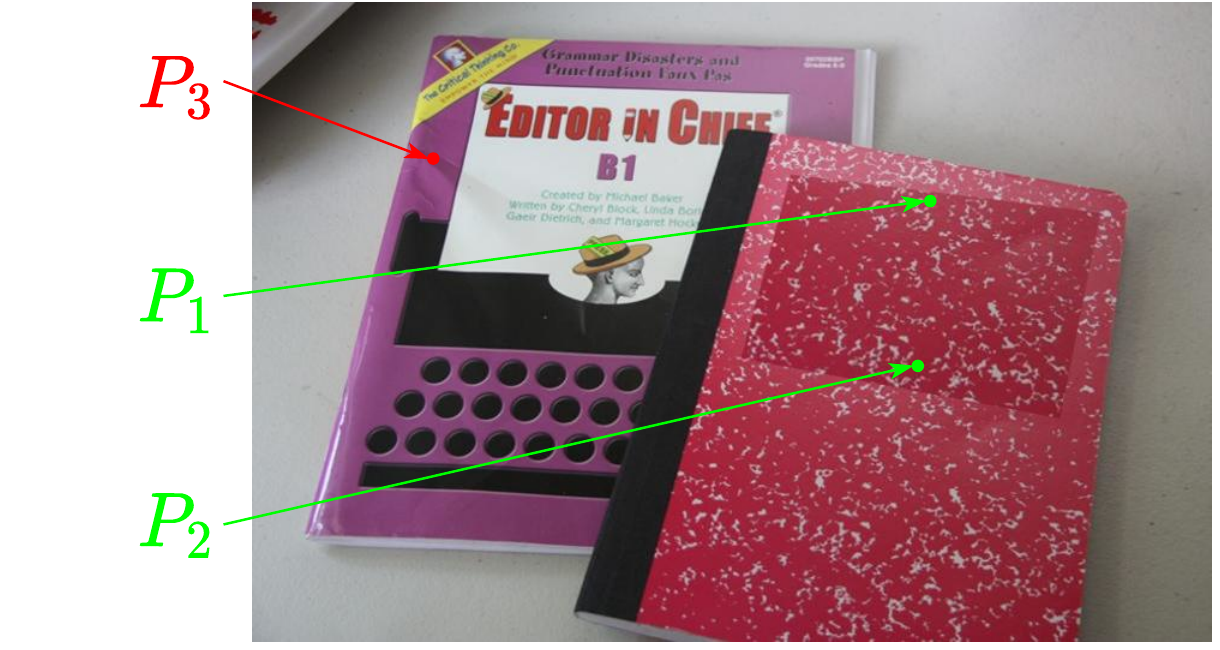}
	\end{minipage}
	\vspace{1mm}
	\begin{minipage}[h]{0.132\linewidth}
		\centering
		\includegraphics[width=\linewidth]{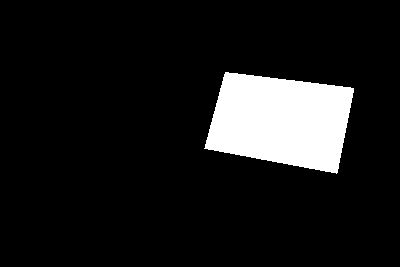}
	\end{minipage}
	\begin{minipage}[h]{0.132\linewidth}
		\centering
		\includegraphics[width=\linewidth]{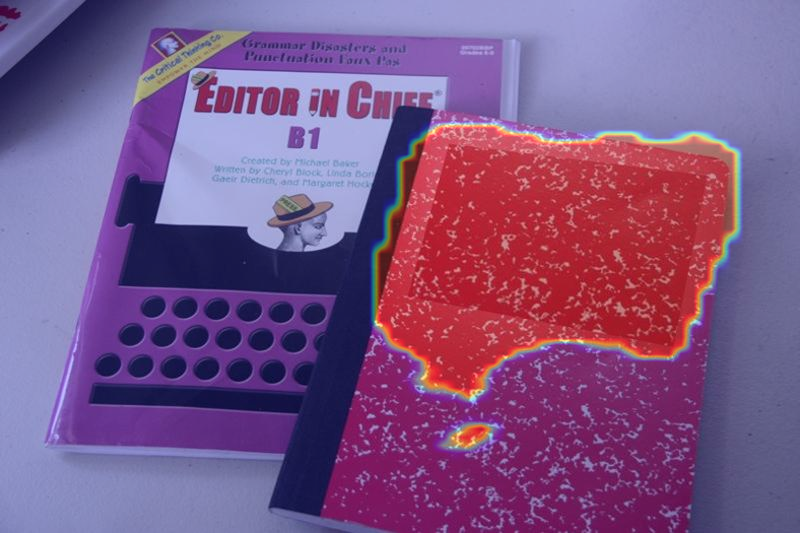}
	\end{minipage}
	\begin{minipage}[h]{0.132\linewidth}
		\centering
		\includegraphics[width=\linewidth]{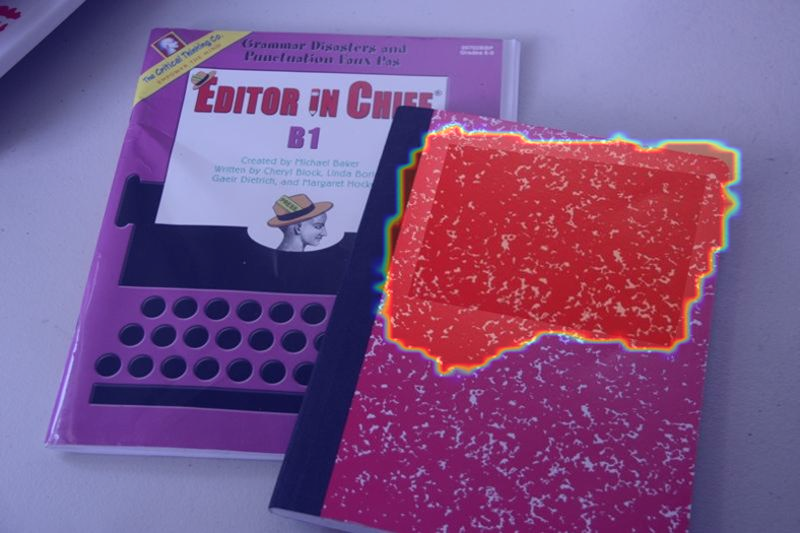}
	\end{minipage}
	\begin{minipage}[h]{0.132\linewidth}
		\centering
		\includegraphics[width=\linewidth]{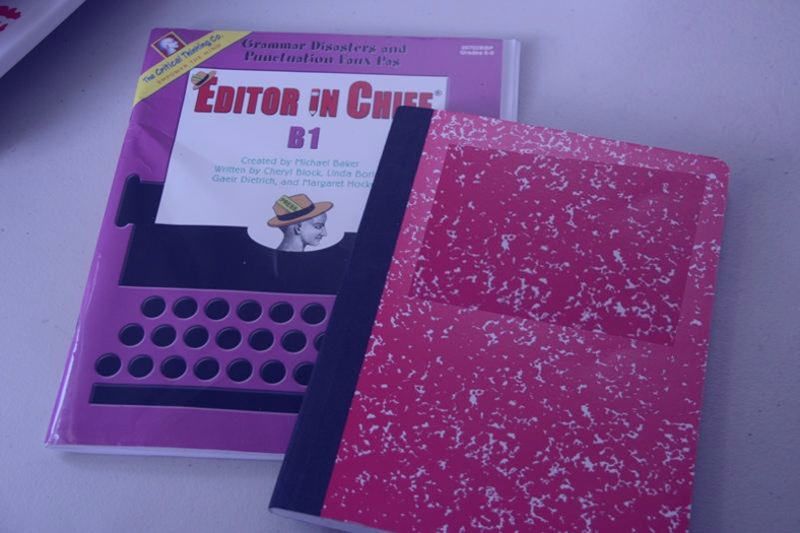}
	\end{minipage}
	\begin{minipage}[h]{0.132\linewidth}
		\centering
		\includegraphics[width=\linewidth]{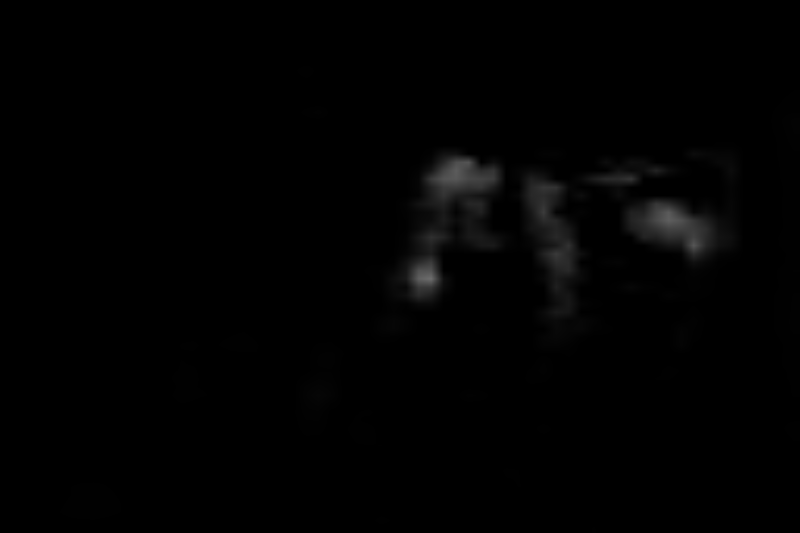}
	\end{minipage}
	\begin{minipage}[h]{0.132\linewidth}
		\centering
		\includegraphics[width=\linewidth]{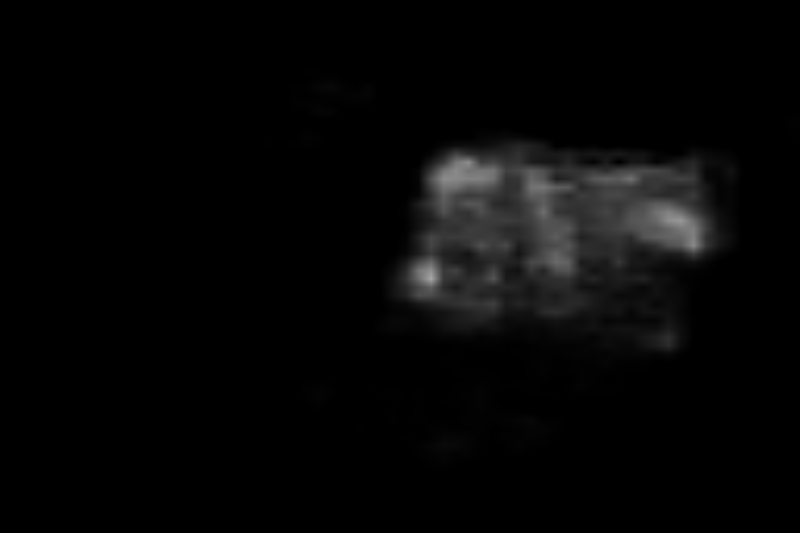}
	\end{minipage}
	
	\begin{minipage}[h]{0.168\linewidth}
		\centering
		\includegraphics[width=\linewidth]{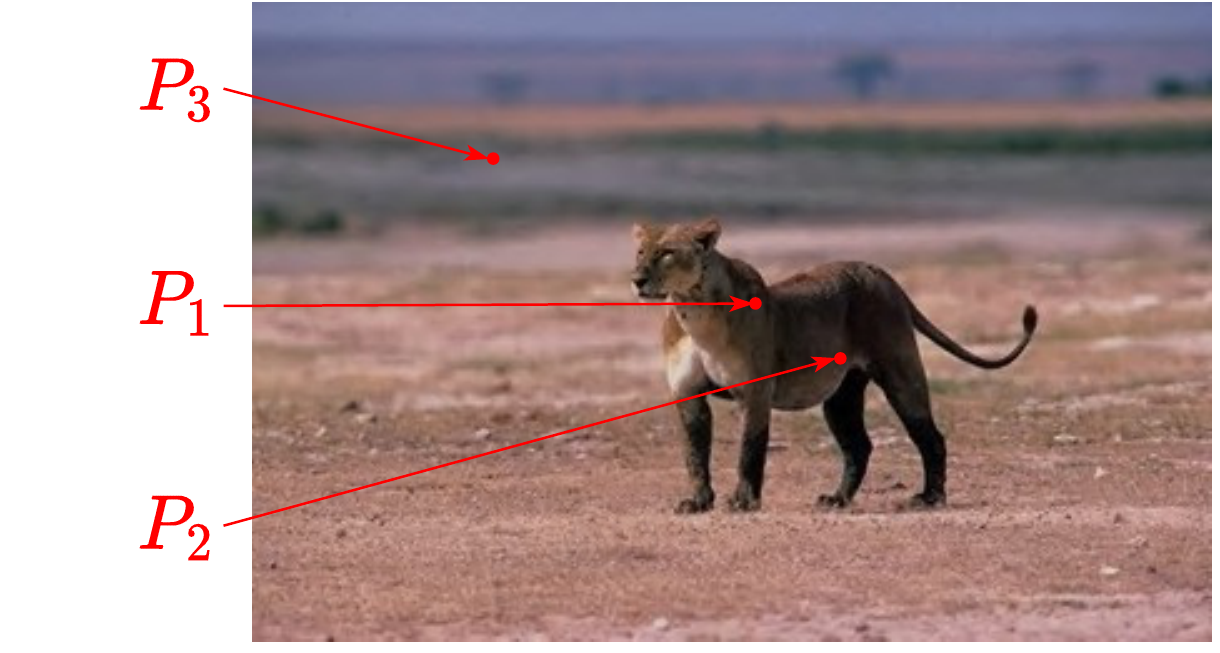}
		\scriptsize{(a) Test Image}
	\end{minipage}
	\begin{minipage}[h]{0.132\linewidth}
		\centering
		\includegraphics[width=\linewidth]{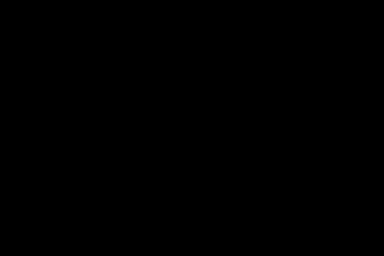}
		\scriptsize{(b) GT manip. mask}
	\end{minipage}
	\begin{minipage}[h]{0.132\linewidth}
		\centering
		\includegraphics[width=\linewidth]{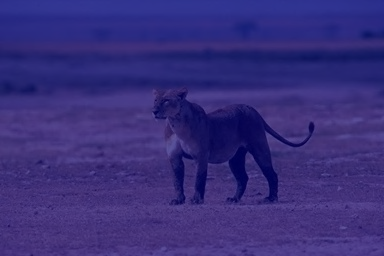}
		\scriptsize{(c) $P_1$ response}
	\end{minipage}
	\begin{minipage}[h]{0.132\linewidth}
		\centering
		\includegraphics[width=\linewidth]{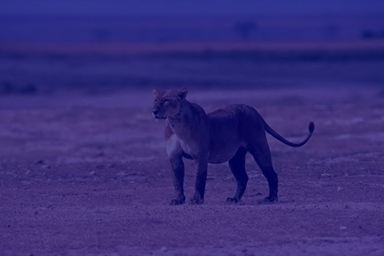}
		\scriptsize{(d) $P_2$ response}
	\end{minipage}
	\begin{minipage}[h]{0.132\linewidth}
		\centering
		\includegraphics[width=\linewidth]{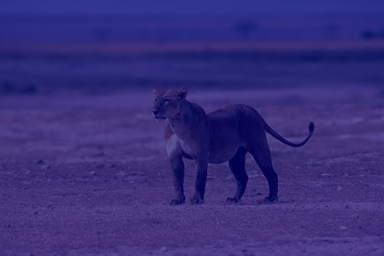}
		\scriptsize{(e) $P_3$ response}
	\end{minipage}
	\begin{minipage}[h]{0.132\linewidth}
		\centering
		\includegraphics[width=\linewidth]{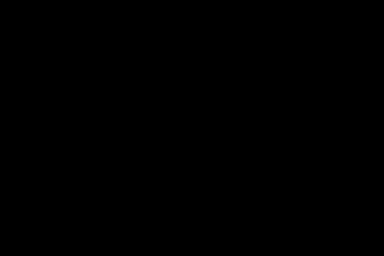}
		\scriptsize{(f) $1$st channel in $\mathbf{X}$}
	\end{minipage}
	\begin{minipage}[h]{0.132\linewidth}
		\centering
		\includegraphics[width=\linewidth]{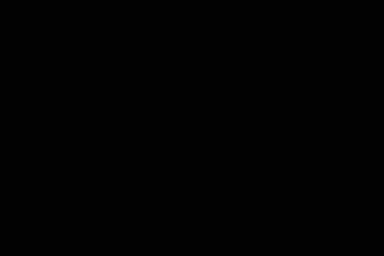}
		\scriptsize{(g) $1$st channel in $\mathbf{Y}_c$}
	\end{minipage}
	\caption{Visualization of spatial and channel-wise attentions in SCCM. From top to bottom, we show the spliced, copy-moved, inpainted, and authentic images respectively. For each test image, we show its GT manipulation mask, $3$ spatial response maps (one for each selected pixel), and the $1$st channel map in $\mathbf{X}$ and $\mathbf{Y}_c$. Zoom in for details.}
	\label{fig:visualization}
\end{figure*}
\subsection{Visualization of SCCM}
To provide insights into SCCM, we visualize the spatial response map for forged and pristine pixels in $\mathbf{M}_3$, by examining its spatial correlation represented in $\mathbf{A}_s$.
After interpolation, each row of $\mathbf{A}_s$ is associated with one pixel (\textit{e.g.}, $P_1$) in the test image, and its grayscale spatial response map can be obtained by reshaping this row vector from $1 \times HW$ to $H \times W$ (\textit{e.g.}, $P_1$ response). 
In Fig.~\ref{fig:visualization} (a), spliced, copy-moved, inpainted, and authentic images are shown from top to bottom respectively, each with one example. We select $3$ representative pixels for each image and annotate as $P_1$, $P_2$, and $P_3$. For manipulated images, $P_1$ and $P_2$ are from forged regions, and $P_3$ is from pristine regions; as for the authentic image, all pixels are pristine.
We project their grayscale spatial response maps into \textit{Jet} color map and overlay them on the manipulated image as in Figs.~\ref{fig:visualization} (c-e). It can be seen that for manipulated images, the spatial response maps of $P_1$ and $P_2$ have high values in forged regions and low values in pristine regions at most cases, but the map of $P_3$ retains low values in all regions including the one providing the copied content (\textit{e.g.}, the $P_3$ response in the $2$nd row of Fig.~\ref{fig:visualization} (e)). As for the authentic image, the spatial response maps of all selected pixels retain low values consistently. This visualization indicates that the features in forged regions are successfully clustered together, thus justifies the effectiveness of spatial attention in SCCM.

For channel-wise correlation $\mathbf{A}_c$, it is hard to provide a comprehensible visualization. Instead, we choose to visualize one channel of $\mathbf{Y}_c$ and compare it to the same channel of $\mathbf{X}$ to see if any region is enhanced. We visualize the $1$st channel of $\mathbf{X}$ and $\mathbf{Y}_c$ in Figs.~\ref{fig:visualization} (f,g). Indeed, the forged region in $\mathbf{Y}_c$ is consolidated compared to the one in  $\mathbf{X}$, and if the forged region does not exist (\textit{i.e.,} in the case of authentic images), no region is enhanced. This proves the effectiveness of channel-wise attention in SCCM.

\begin{figure*}[t!]
	\centering
	\begin{minipage}[h]{0.155\linewidth}
		\centering
		\includegraphics[width=\linewidth]{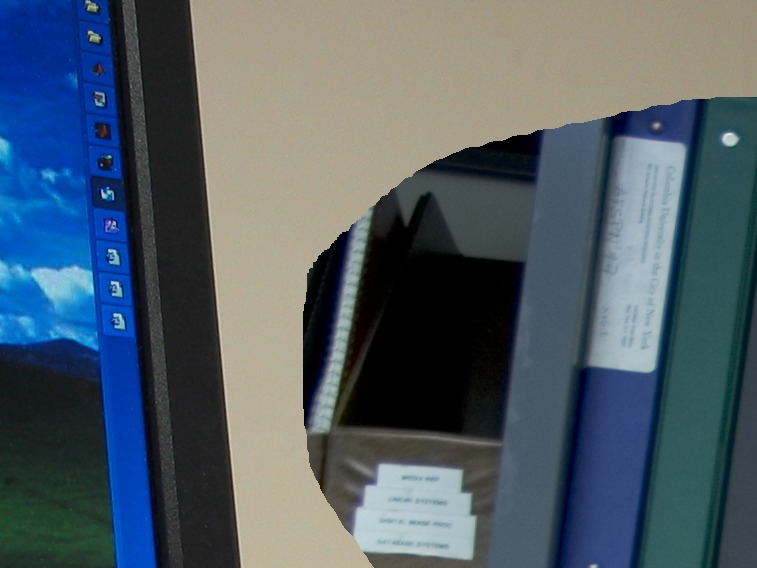}
	\end{minipage}
	\vspace{1mm}
	\begin{minipage}[h]{0.155\linewidth}
		\centering
		\includegraphics[width=\linewidth]{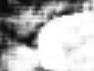}
	\end{minipage}
	\begin{minipage}[h]{0.155\linewidth}
		\centering
		\includegraphics[width=\linewidth]{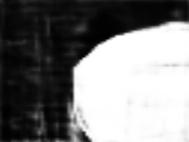}
	\end{minipage}
	\begin{minipage}[h]{0.155\linewidth}
		\centering
		\includegraphics[width=\linewidth]{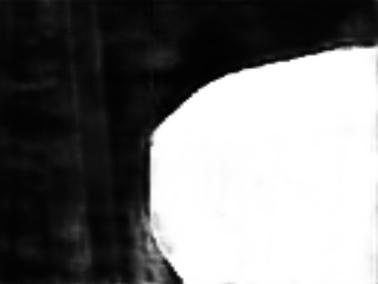}
	\end{minipage}
	\begin{minipage}[h]{0.155\linewidth}
		\centering
		\includegraphics[width=\linewidth]{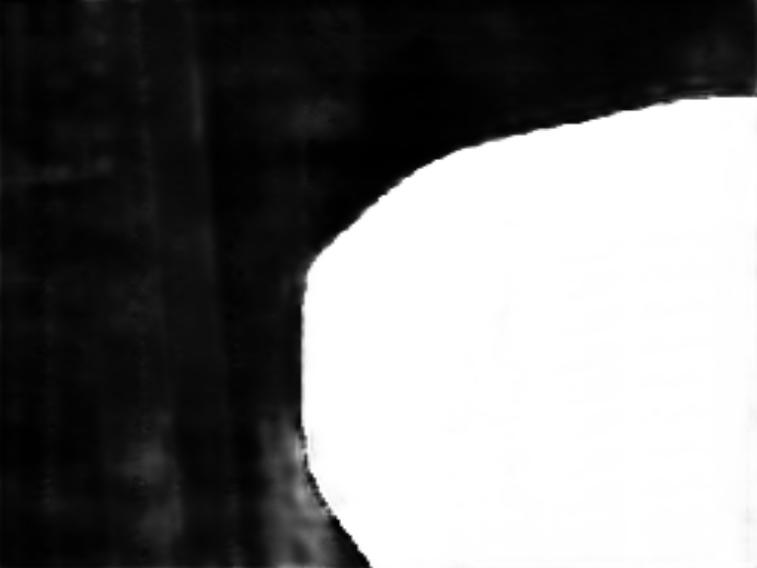}
	\end{minipage}
	\begin{minipage}[h]{0.155\linewidth}
		\centering
		\includegraphics[width=\linewidth]{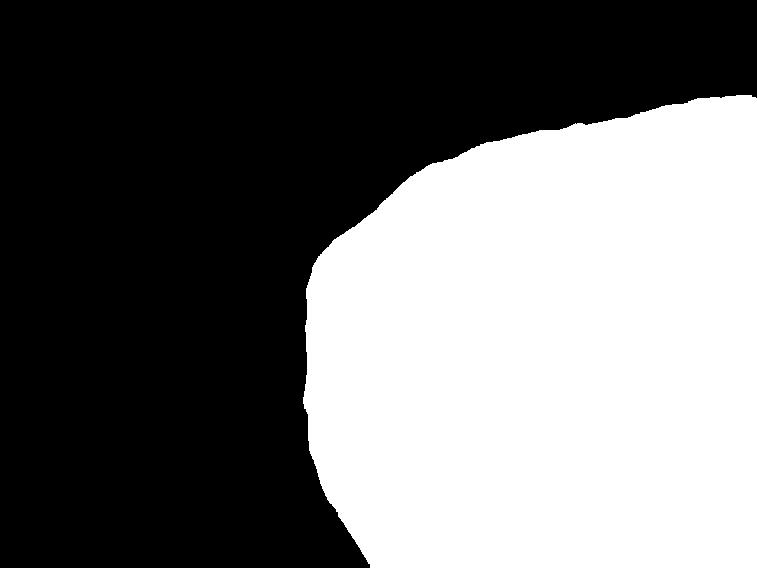}
	\end{minipage}
	\begin{minipage}[h]{0.155\linewidth}
		\centering
		\includegraphics[width=\linewidth]{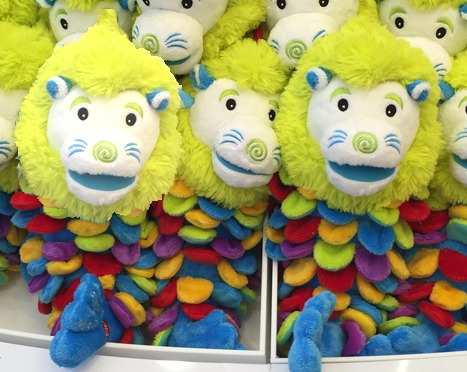}
	\end{minipage}
	\vspace{1mm}
	\begin{minipage}[h]{0.155\linewidth}
		\centering
		\includegraphics[width=\linewidth]{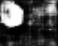}
	\end{minipage}
	\begin{minipage}[h]{0.155\linewidth}
		\centering
		\includegraphics[width=\linewidth]{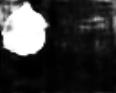}
	\end{minipage}
	\begin{minipage}[h]{0.155\linewidth}
		\centering
		\includegraphics[width=\linewidth]{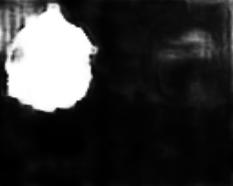}
	\end{minipage}
	\begin{minipage}[h]{0.155\linewidth}
		\centering
		\includegraphics[width=\linewidth]{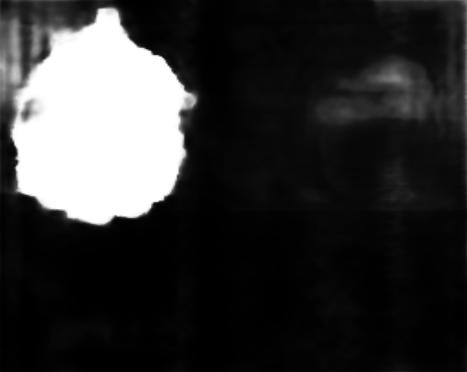}
	\end{minipage}
	\begin{minipage}[h]{0.155\linewidth}
		\centering
		\includegraphics[width=\linewidth]{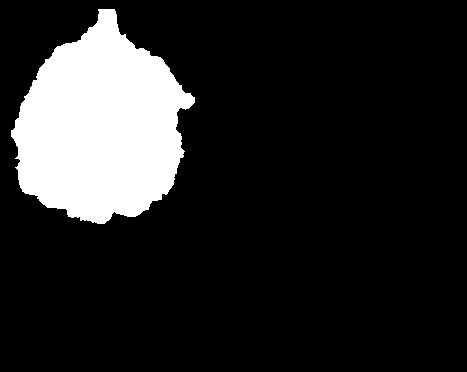}
	\end{minipage}
	\begin{minipage}[h]{0.155\linewidth}
		\centering
		\includegraphics[width=\linewidth]{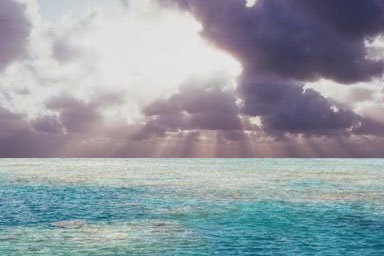}
	\end{minipage}
	\vspace{1mm}
	\begin{minipage}[h]{0.155\linewidth}
		\centering
		\includegraphics[width=\linewidth]{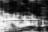}
	\end{minipage}
	\begin{minipage}[h]{0.155\linewidth}
		\centering
		\includegraphics[width=\linewidth]{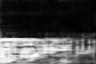}
	\end{minipage}
	\begin{minipage}[h]{0.155\linewidth}
		\centering
		\includegraphics[width=\linewidth]{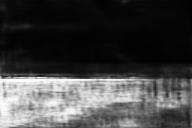}
	\end{minipage}
	\begin{minipage}[h]{0.155\linewidth}
		\centering
		\includegraphics[width=\linewidth]{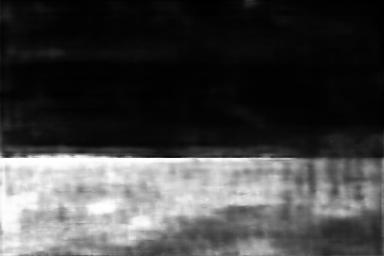}
	\end{minipage}
	\begin{minipage}[h]{0.155\linewidth}
		\centering
		\includegraphics[width=\linewidth]{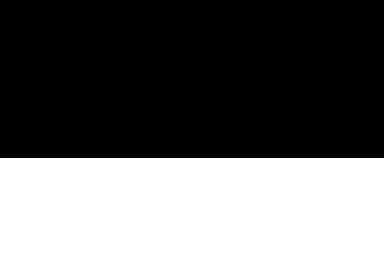}
	\end{minipage}
	\begin{minipage}[h]{0.155\linewidth}
		\centering
		\includegraphics[width=\linewidth]{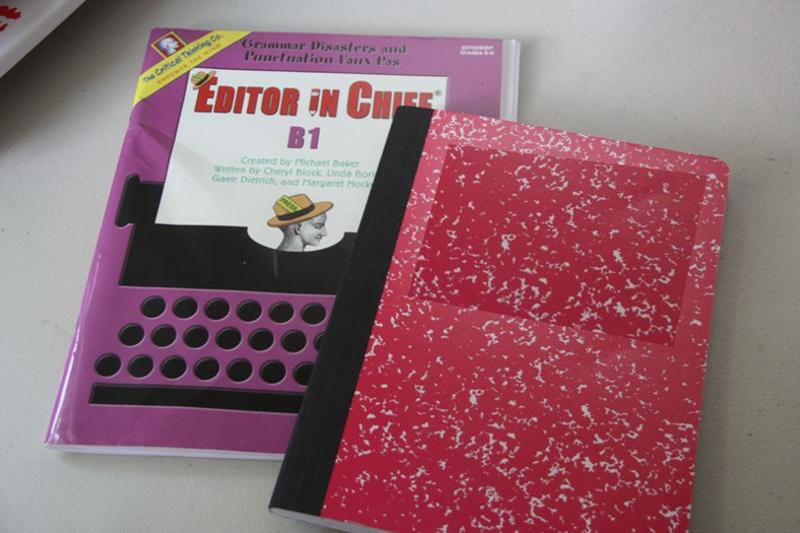}
	\end{minipage}
	\vspace{1mm}
	\begin{minipage}[h]{0.155\linewidth}
		\centering
		\includegraphics[width=\linewidth]{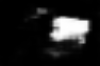}
	\end{minipage}
	\begin{minipage}[h]{0.155\linewidth}
		\centering
		\includegraphics[width=\linewidth]{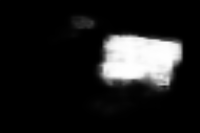}
	\end{minipage}
	\begin{minipage}[h]{0.155\linewidth}
		\centering
		\includegraphics[width=\linewidth]{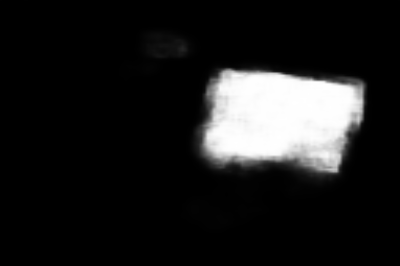}
	\end{minipage}
	\begin{minipage}[h]{0.155\linewidth}
		\centering
		\includegraphics[width=\linewidth]{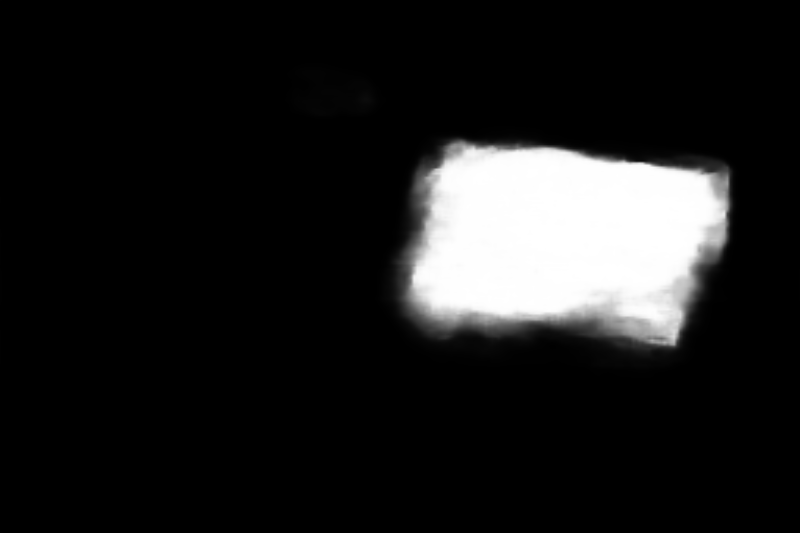}
	\end{minipage}
	\begin{minipage}[h]{0.155\linewidth}
		\centering
		\includegraphics[width=\linewidth]{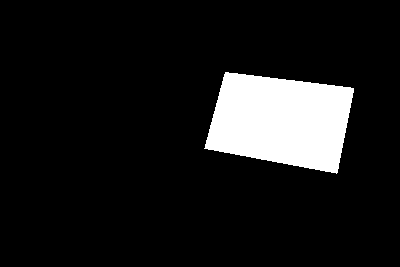}
	\end{minipage}
	\begin{minipage}[h]{0.155\linewidth}
	\centering
	\includegraphics[width=\linewidth]{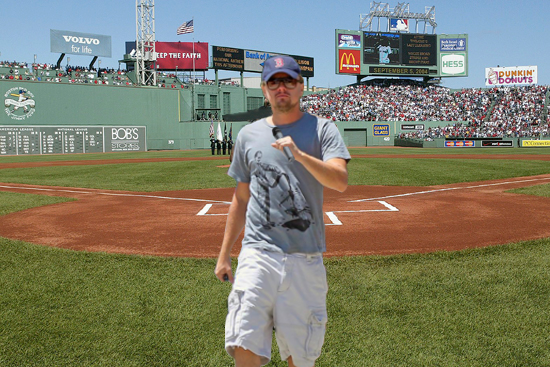}
	\scriptsize{(a) Manipulated image}
	\end{minipage}
	\begin{minipage}[h]{0.155\linewidth}
		\centering
		\includegraphics[width=\linewidth]{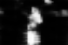}
		\scriptsize{(b) Mask $4$}
	\end{minipage}
	\begin{minipage}[h]{0.155\linewidth}
		\centering
		\includegraphics[width=\linewidth]{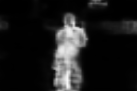}
		\scriptsize{(c) Mask $3$}
	\end{minipage}
	\begin{minipage}[h]{0.155\linewidth}
		\centering
		\includegraphics[width=\linewidth]{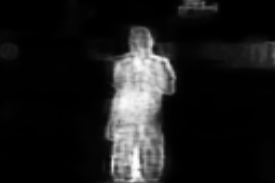}
		\scriptsize{(d) Mask $2$}
	\end{minipage}
	\begin{minipage}[h]{0.155\linewidth}
		\centering
		\includegraphics[width=\linewidth]{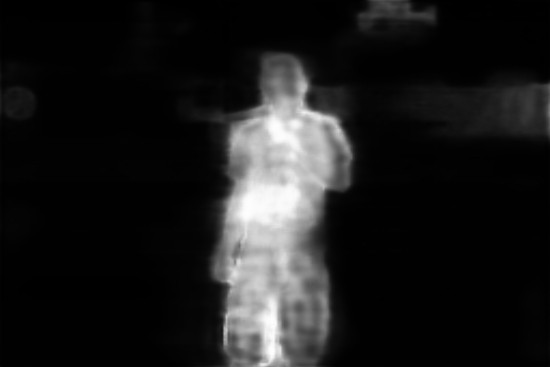}
		\scriptsize{(e) Mask $1$}
	\end{minipage}
	\begin{minipage}[h]{0.155\linewidth}
		\centering
		\includegraphics[width=\linewidth]{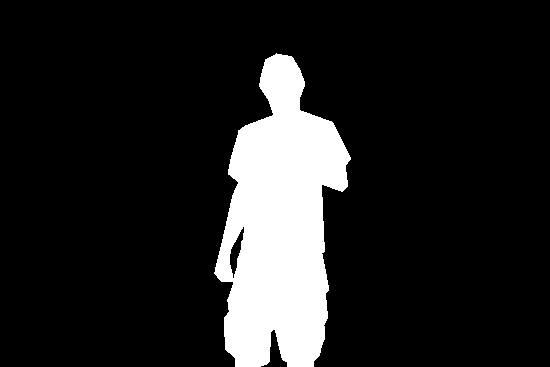}
		\scriptsize{(f) GT manip. mask}
	\end{minipage}
	\caption{Visualization of predicted manipulation masks from \textit{Scale} $4$ to \textit{Scale} $1$. From top to bottom, manipulated images are from Columbia, Coverage, CASIA, NIST16, and IMD20 respectively. All predicted masks are from our pre-trained model.}
	\label{fig:S_mask}
\end{figure*}
\begin{table*}[t!]
	\centering
	\renewcommand{\arraystretch}{1.4}
	
	\caption{Ablation study of PSCC-Net (AUC/F1 in $\%$). The runtime is reported in proportion to that of original PSCC-Net.}
	\begin{adjustbox}{width=0.75\linewidth}
		\begin{tabu}{llllll}
			\toprule
			Variants & Columbia & Coverage & CASIA & NIST16 & Runtime\\\hline
			 Mask 4 & $93.34$ / $79.22$ & $82.99$ / $44.23$ & $81.49$ / $31.69$ & $84.15$ / $30.55$ & $\mathbf{0.63}$ \\
			 Mask 3 & $98.08$ / $92.41$ & $83.48$ / $47.29$ & $82.55$ / $34.64$ & $85.25$ / $33.55$ & $0.75$ \\
			 Mask 2 & $98.18$ / $93.32$ & $84.44$ / $49.08$ & $82.78$ / $35.59$ & $85.38$ / $34.94$ & $0.88$ \\\hline
			 w/o CA+SA & $85.78$ / $70.32$ & $79.95$ / $43.27$ & $79.26$ / $31.06$ & $79.58$ / $31.73$ & $0.84$\\
			 w/o SA & $90.70$ / $75.68$ & $80.56$ / $43.50$ & $79.51$ / $31.08$ & $83.49$ / $32.34$ & $0.92$\\
			 w/o CA & $94.50$ / $85.34$ & $82.16$ / $45.04$ & $82.63$ / $35.97$ & $84.65$ / $33.42$ & $0.92$ \\
			 w/o FS & $97.54$ / $91.30$ & $84.06$ / $48.23$ &$81.88$ / $34.33$ &$84.45$ / $34.21$ & $1.04$ \\\hline
			 PSCC-Net & $\mathbf{98.19}$ / $\mathbf{93.45}$ & $\mathbf{84.65}$ / $\mathbf{49.78}$ & $\mathbf{82.93}$ / $\mathbf{36.27}$ & $\mathbf{85.47}$ / $\mathbf{35.73}$ & $1.00$ \\
			\bottomrule
		\end{tabu}
	\end{adjustbox}
	\label{table:ablation}
\end{table*}

\begin{table*}[t!]
	\begin{center}
		\renewcommand{\arraystretch}{1.4}
		\caption{Robustness analysis of localization with respect to various distortions. Pixel-level AUCs are reported (in $\%$).}
		\label{table:robustness}
		\begin{adjustbox}{width=\linewidth}
			\begin{tabular}{ccccccccccc}
				\toprule
				\multirow{2}{*}{Distortion} & Resize & Resize & GSBlur & GSBlur & GSNoise & GSNoise & JPEGComp & JPEGComp & \multirow{2}{*}{Mixed} & \multirow{2}{*}{ w/o distortion} \\ & $0.78\times$ & $0.25\times$ & $k = 3$ & $k = 15$ & $\sigma = 3$ & $\sigma = 15$ &  $q = 100$ & $q = 50$ \\ \hline
				\multicolumn{11}{c}{Columbia} \\ \hline
				ManTraNet~\cite{wu2019mantra} & $71.66$ & $68.64$ & $67.72$ & $62.88$ & $68.22$ & $54.97$ & $75.00$ & $59.37$ & $60.47$ & $77.95$ \\
				SPAN~\cite{hu2020span} & $89.99$ & $69.08$ & $78.97$ & $67.70$ & $75.11$ & $65.80$ & $93.32$ & $74.62$ & $62.54$ & $93.60$ \\
				PSCC-Net & $\mathbf{93.40}$ & $\mathbf{78.41}$ & $\mathbf{84.18}$ & $\mathbf{73.24}$ & $\mathbf{82.64}$ & $\mathbf{74.35}$ & $\mathbf{97.97}$ & $\mathbf{89.11}$ & $\mathbf{72.69}$ & $\mathbf{98.19}$ \\ \hline
				\multicolumn{11}{c}{NIST16} \\ \hline
				ManTraNet~\cite{wu2019mantra} & $77.43$ & $75.52$ & $77.46$ & $74.55$ & $67.41$ & $58.55$ & $77.91$ & $74.38$ & $64.82$ & $78.05$ \\
				SPAN~\cite{hu2020span} & $83.24$ & $80.32$ & $83.10$ & $79.15$ & $75.17$ & $67.28$ & $83.59$ & $80.68$ & $68.36$ & $83.95$ \\
				PSCC-Net & $\mathbf{85.29}$ & $ \mathbf{85.01}$ & $\mathbf{85.38}$ & $\mathbf{79.93}$ & $\mathbf{78.42}$ & $\mathbf{76.65}$ & $\mathbf{85.40}$ & $\mathbf{85.37}$ & $\mathbf{73.93}$ & $\mathbf{85.47}$ \\
				\bottomrule
			\end{tabular}
		\end{adjustbox}
	\end{center}
\end{table*}

\subsection{Visualization of Predicted Manipulation Masks on Different Scales}
The proposed PSCC-Net utilizes a progressive mechanism to reduce the prediction difficulty by avoiding generating the mask from the finest scale directly. Instead, the mask on the coarsest scale is first predicted to locate the regions that are potentially forged based on the current available information. The subsequent prediction on the finer scale can leverage the previous mask and pay more attention to those selected regions. This process repeatedly performs until generating the manipulation mask at the finest scale as our final prediction.

Here, we visualize the performance improvement of manipulation localization from the Scale 4 to Scale 1. In Fig.~\ref{fig:S_mask}, \textit{Mask 4}, \textit{Mask 3}, and \textit{Mask 2} are the variants that truncate the original model after generating manipulation masks on the $4$th, $3$rd, and $2$nd scales, and \textit{Mask 1} is the output of the original model. It can be seen that benefiting to the proposed progressive mechanism, the localization performance is gradually improved from \textit{Mask 4} to \textit{Mask 1} in terms of lower false alarms and clearer boundaries. More discussions about quantitative comparisons and terminating PSCC-Net earlier for runtime saving can be found in Sec.~\ref{sec:ablation}.

\subsection{Runtime Analysis and Ablation Study} \label{sec:ablation}
In Tab.~\ref{table:ablation}, we test several variants of PSCC-Net to justify the network design, where all variants are pre-trained on our synthetic dataset. Average AUC/F$1$s are reported (in $\%$), and the runtime (in proportion) is relative to that of PSCC-Net. Our full model takes $0.019s$ to process one $1,080$P image, whereas ManTra-Net and SPAN take $0.208s$ and $0.161s$, respectively. Moreover, as shown in Fig.~\ref{fig:PDANet}, terminating the PSCC-Net earlier on \textit{Mask 4}, \textit{Mask 3} or \textit{Mask 2} in inference time is feasible and will not interfere the prediction of manipulation mask at that scale. From our experiments, terminating the prediction on \textit{Mask 4} can shorten the runtime to $0.012s$, \textit{i.e.}, $\sim37\%$ additional saving. Though nonessential for research datasets, this time-saving is significant and economical in practical applications, \textit{e.g.}, $14.6$ million photos
are uploaded to Facebook \textit{per hour}\footnote{https://www.pingdom.com/blog/social-media-in-2017/}.

The comparisons of \textit{Mask 4}, \textit{Mask 3}, \textit{Mask 2}, and the original PSCC-Net demonstrate the  gradual improvement in performance, 
which is a clear manifestation of our progressive mechanism. Since \textit{Mask 3} already performs well under AUC and F1 scores, it is a good stopping point for mask prediction.

We also build several variants for SCCM, including the ones without spatial and channel-wise attentions (\textit{w/o SA+CA}), without spatial attention (\textit{w/o SA}),  without channel-wise attention (\textit{w/o CA}), and without feature sharing (\textit{w/o FS}), which obtains embeddings from different $\theta$ and $\phi$ functions to compute spatial and channel-wise similarities. The comparisons illustrate that both SA and CA outperform the baseline (\textit{w/o SA+CA}), and the performance gain acquired from SA is more than that from CA. In addition, feature sharing not only slightly reduces the runtime, but also enables mutual accommodation between these two attentions to help SCCM achieve better results than the one employing different features (\textit{i.e., w/o FS}).

\begin{table}[t!]
	\centering
	\caption{Robustness analysis of detection for PSCC-Net with respect to various distortions on CASIA-D. Image-level AUCs and F1 scores are reported (in $\%$).}
	\renewcommand{\arraystretch}{1.4}
	\begin{adjustbox}{width=0.65\linewidth}
		\begin{tabu}{lll}
			\toprule
			Distortion & AUC & F1 \\ \hline
			Resize $0.78\times$ & $95.48$ & $91.46$ \\
			Resize $0.25\times$ & $74.86$ & $70.47$ \\
			GSBlur $k = 3$ & $92.93$ & $87.55$ \\
			GSBlur $k = 15$ &  $88.59$ & $83.56$ \\
			GSNoise $\sigma = 3$  & $89.78$ & $83.37$ \\
			GSNoise $\sigma = 15$  & $85.50$ & $80.87$\\
			JPEGComp $q = 100$ & $99.44$ & $96.47$ \\ 
			JPEGComp $q = 50$ & $99.45$  & $96.47$ \\ 
			Mixed & $87.55$ & $83.51$ \\ \hline
			w/o distortion & $\mathbf{99.65}$ & $\mathbf{97.12}$ \\
			\bottomrule
		\end{tabu}
	\end{adjustbox}
	\label{table:robustness_detection}
\end{table}

\subsection{Robustness Analysis} 
To analyze the robustness of PSCC-Net for localization, we follow the distortion settings in~\cite{hu2020span} to degrade the raw manipulated images from Columbia and NIST16. These distortions include resizing images to a different scale (\textit{Resize}), applying Gaussian blur with kernel size $k$ (\textit{GSBlur}), adding Gaussian noise with standard deviation $\sigma$ (\textit{GSNoise}), and performing JPEG compression with quality factor $q$ (\textit{JPEGComp}). In addition, we 
introduce a mixed version of the aforementioned distortions (\textit{Mixed}), where the resizing scale, kernel size $k$, standard deviation $\sigma$, quality factor $q$ are randomly selected from the intervals $[0.25, 0.78]$, $[3, 15]$, $[3, 15]$, and $[50, 100]$, respectively. Tab.~\ref{table:robustness} shows the robustness analysis of localization under pixel-level AUC with pre-trained models. The PSCC-Net is more robust than ManTra-Net and SPAN under all distortions. It is worth noting that resizing is commonly performed when uploading images to social media. Indeed, benefiting from the operation that resamples the manipulation features into the fixed sizes, the impact of resizing to PSCC-Net is the least as compared to the others.

We also analyze the detection robustness of PSCC-Net with respect to various distortions on CASIA-D. In Tab.~\ref{table:robustness_detection}, it can be seen that our PSCC-Net is quite robust for detection, especially in the case where the JPEG compression is performed.

\subsection{Limitations}
PSCC-Net enables us to detect and localize various types of manipulations. As compared to image-level detection, the pixel-level localization is more challenging, especially while dealing with real-life manipulated images. Here we demonstrate some failure cases on IMD20~\cite{novozamsky2020imd2020}.

\begin{figure}[t]
	\centering
	\begin{minipage}[t]{0.185\linewidth}
		\centering
		\includegraphics[width=\linewidth]{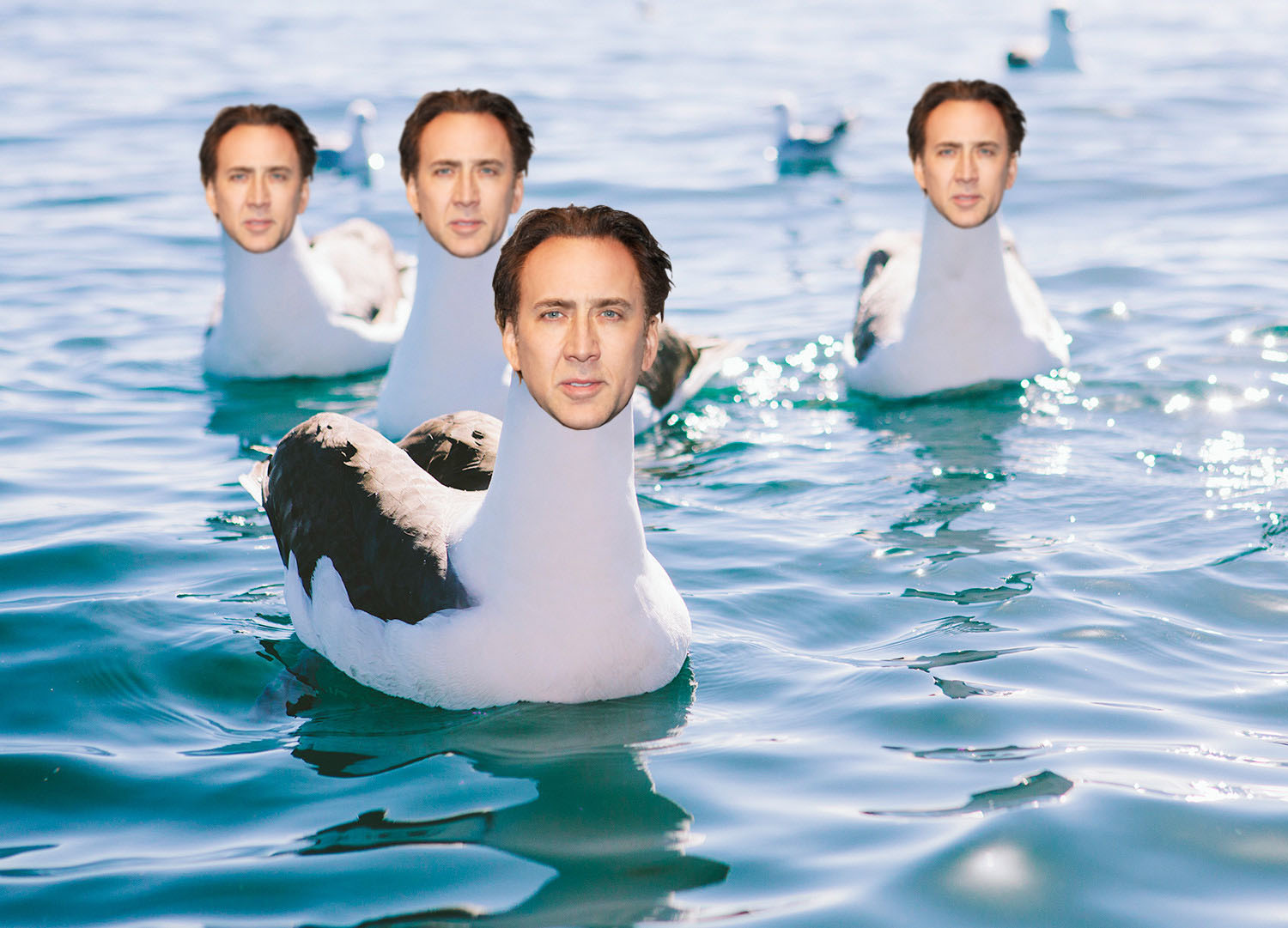}
	\end{minipage}
	\vspace{1mm}
	\begin{minipage}[t]{0.185\linewidth}
		\centering
		\includegraphics[width=\linewidth]{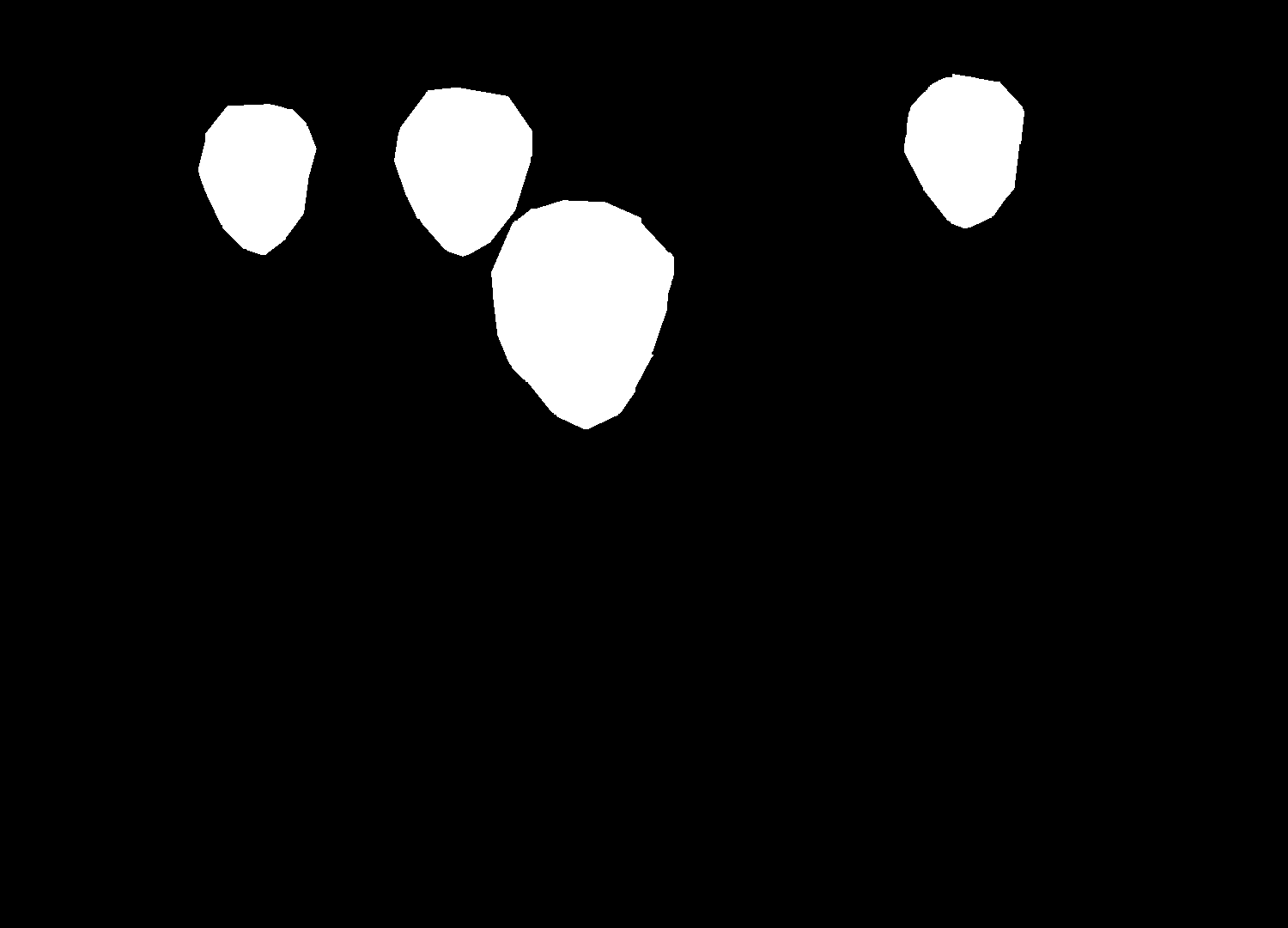}
	\end{minipage}
	\begin{minipage}[t]{0.185\linewidth}
		\centering
		\includegraphics[width=\linewidth]{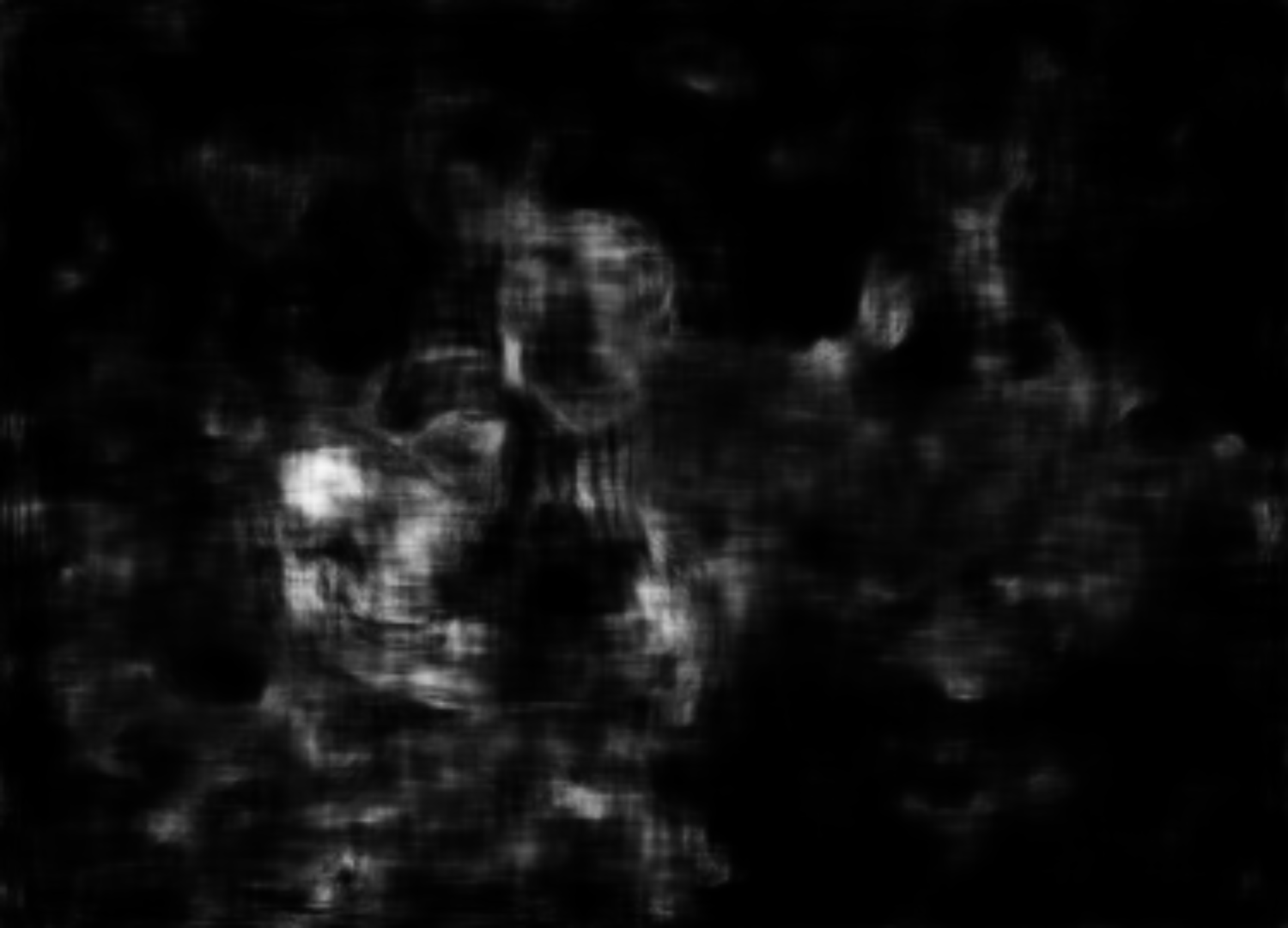}
	\end{minipage}
	\begin{minipage}[t]{0.185\linewidth}
		\centering
		\includegraphics[width=\linewidth]{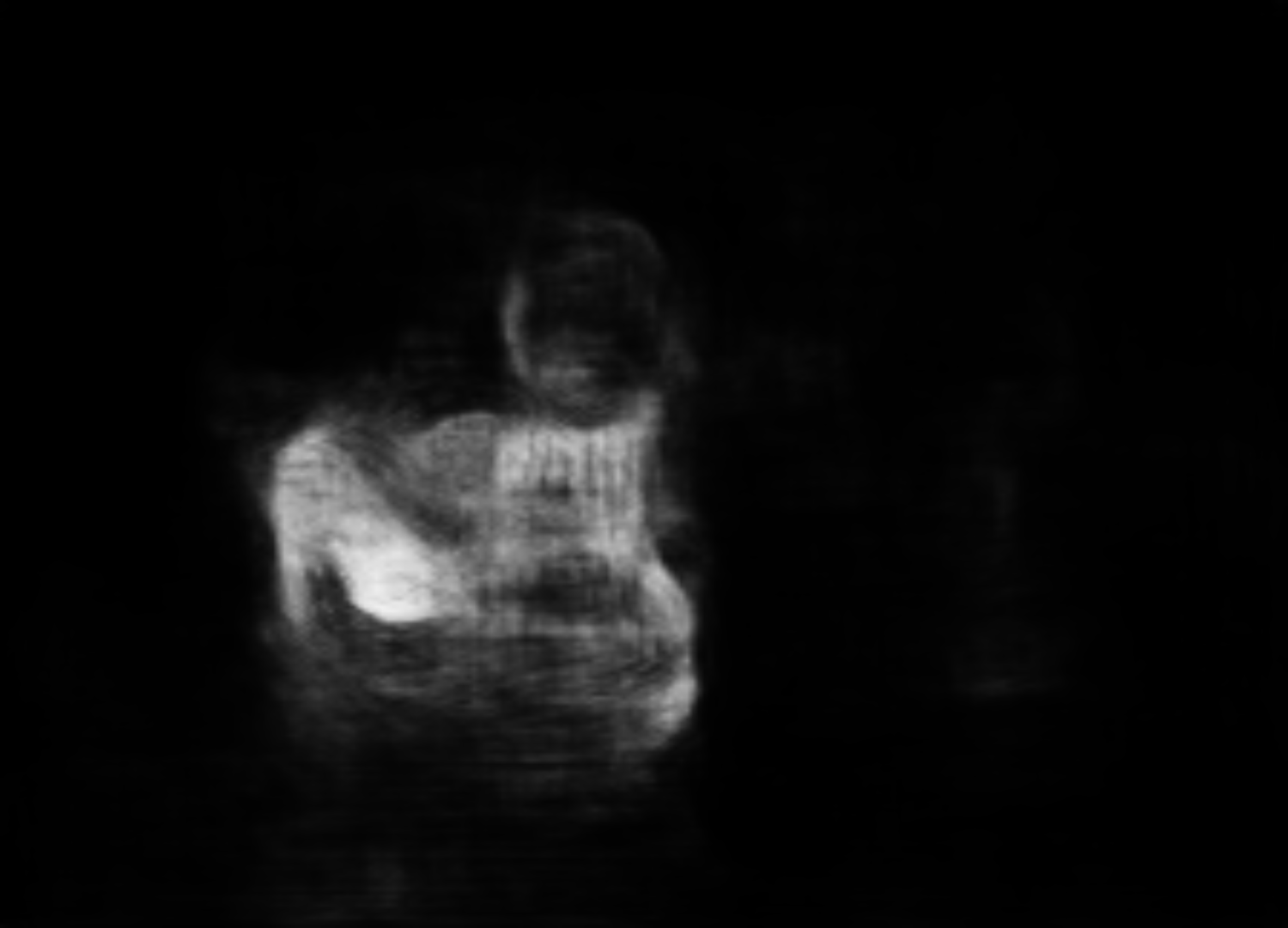}
	\end{minipage}
	\begin{minipage}[t]{0.185\linewidth}
		\centering
		\includegraphics[width=\linewidth]{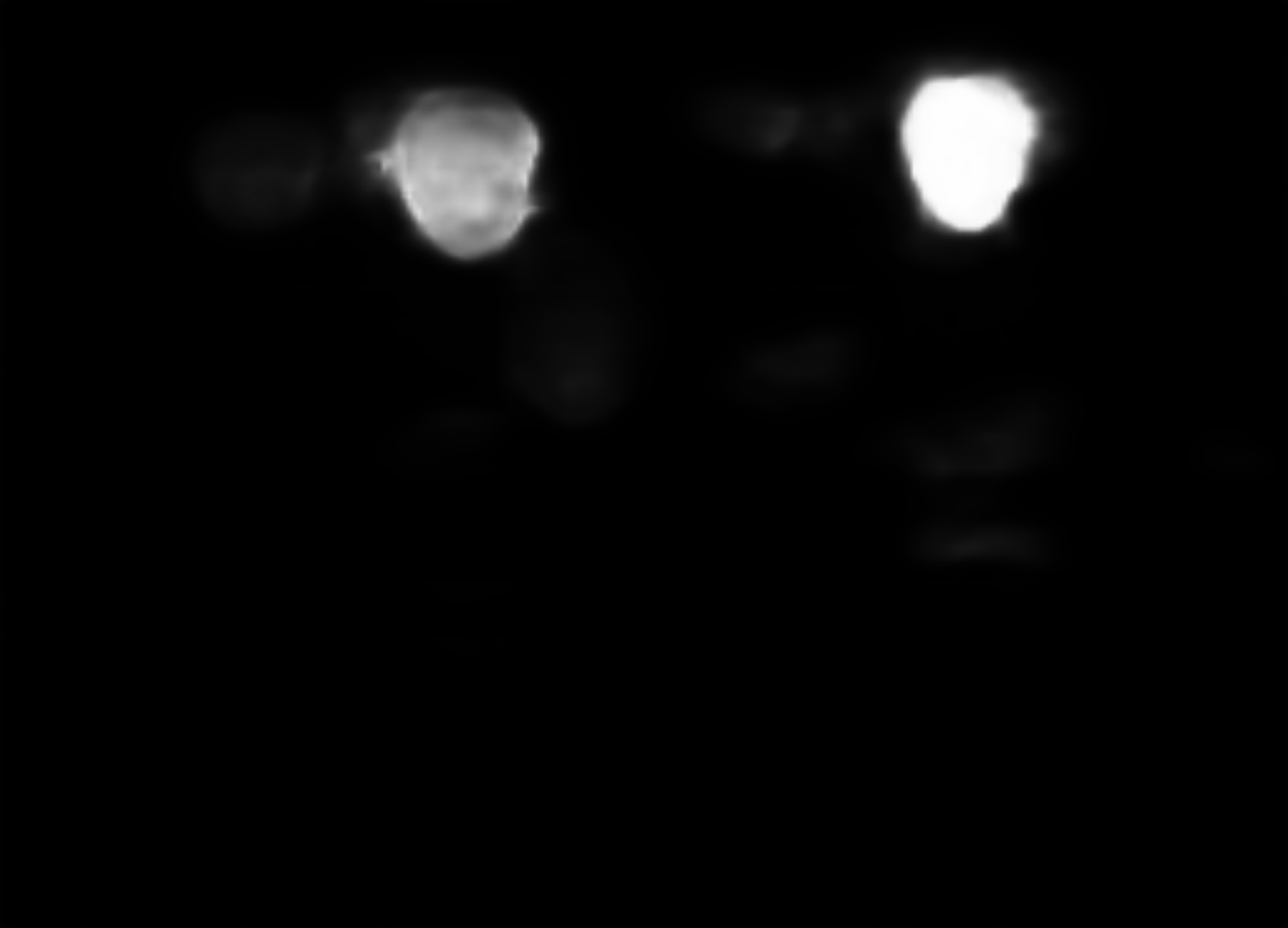}
	\end{minipage}
	\begin{minipage}[t]{0.185\linewidth}
	\centering
	\includegraphics[width=\linewidth]{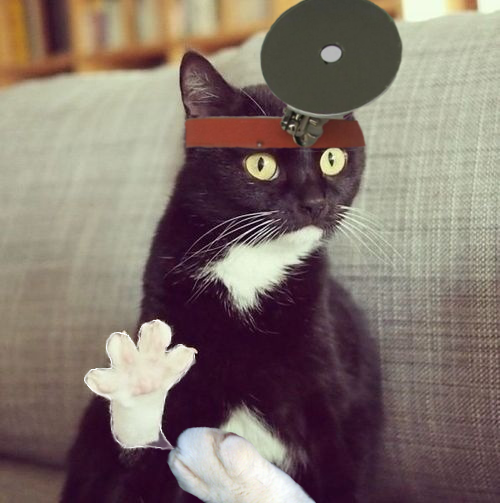}
	\end{minipage}
	\vspace{1mm}
	\begin{minipage}[t]{0.185\linewidth}
		\centering
		\includegraphics[width=\linewidth]{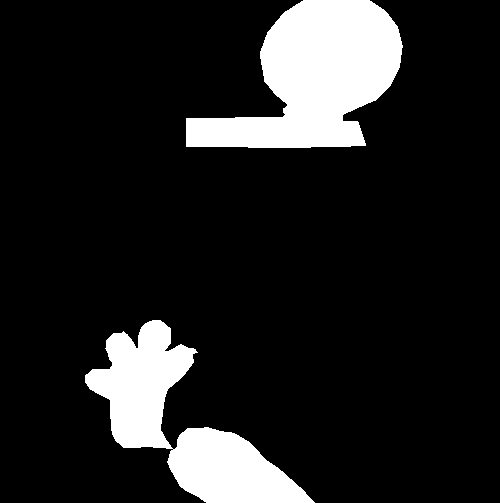}
	\end{minipage}
	\begin{minipage}[t]{0.185\linewidth}
		\centering
		\includegraphics[width=\linewidth]{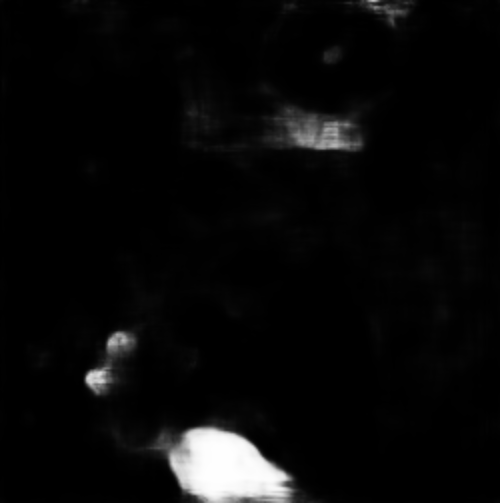}
	\end{minipage}
	\begin{minipage}[t]{0.185\linewidth}
		\centering
		\includegraphics[width=\linewidth]{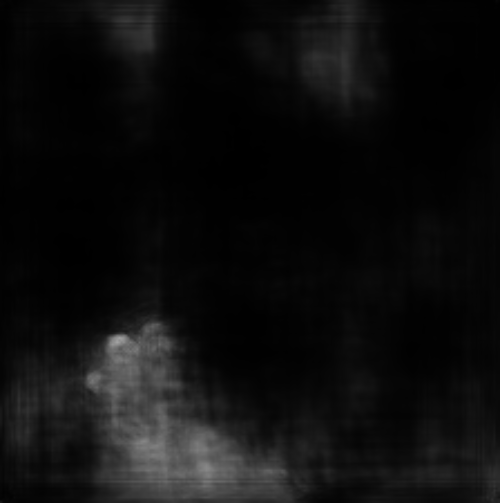}
	\end{minipage}
	\begin{minipage}[t]{0.185\linewidth}
		\centering
		\includegraphics[width=\linewidth]{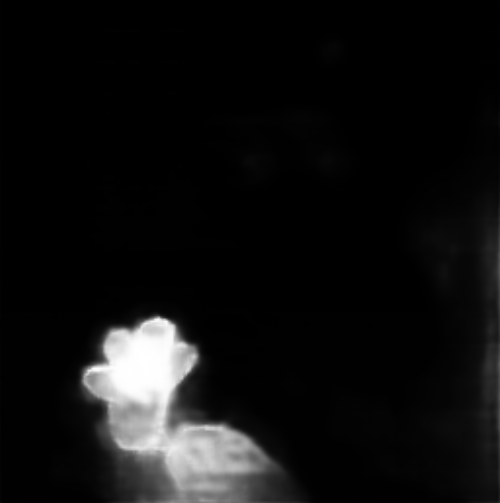}
	\end{minipage}
	\begin{minipage}[t]{0.185\linewidth}
		\centering
		\includegraphics[width=\linewidth]{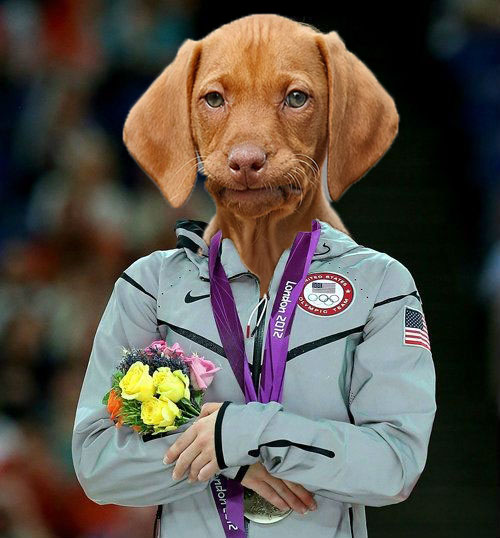}
		\scriptsize{(a) Manipulated}
	\end{minipage}
	\begin{minipage}[t]{0.185\linewidth}
		\centering
		\includegraphics[width=\linewidth]{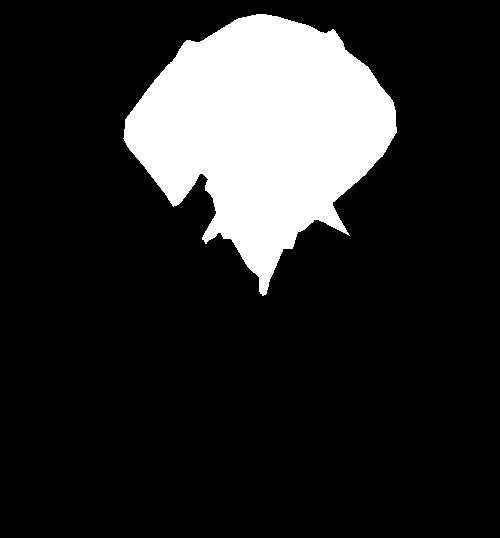}
		\scriptsize{(b) GT}
	\end{minipage}
	\begin{minipage}[t]{0.185\linewidth}
		\centering
		\includegraphics[width=\linewidth]{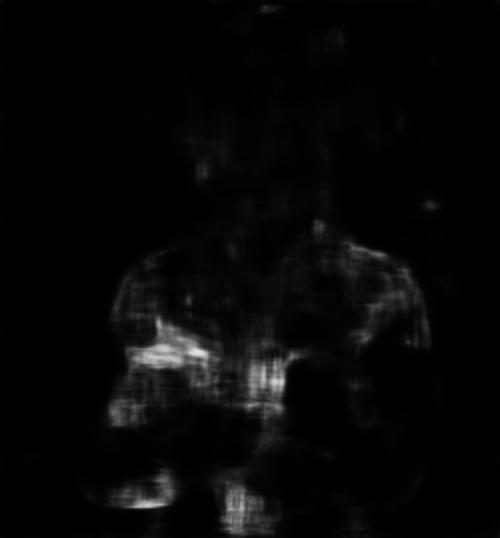}
		\scriptsize{(c) ManTra-Net}
	\end{minipage}
	\begin{minipage}[t]{0.185\linewidth}
		\centering
		\includegraphics[width=\linewidth]{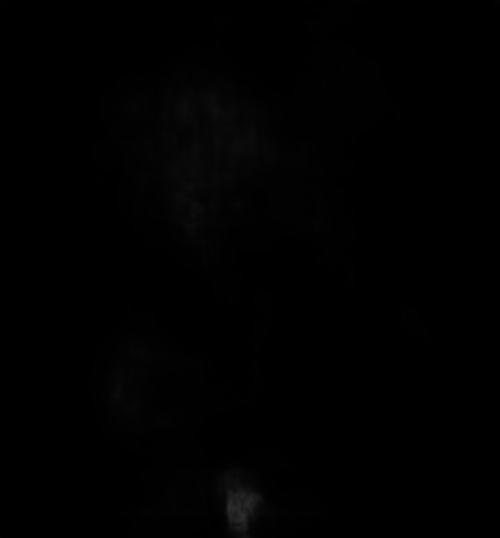}
		\scriptsize{(d) SPAN}
	\end{minipage}
	\begin{minipage}[t]{0.185\linewidth}
		\centering
		\includegraphics[width=\linewidth]{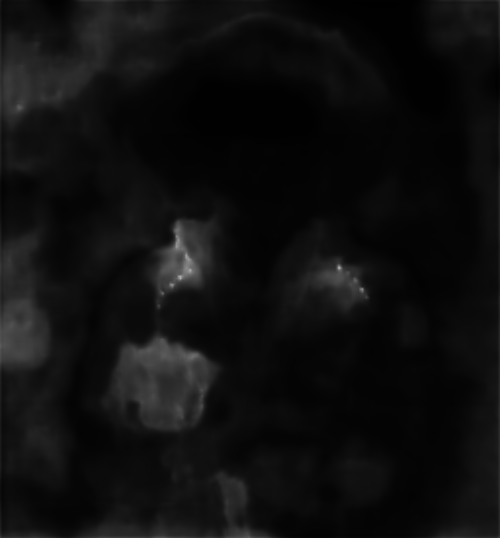}
		\scriptsize{(e) PSCC-Net}
	\end{minipage}
	\caption{\small Failure cases. Zoom in for details.}
	\label{fig:S_limitation}
\end{figure}

In Fig.~\ref{fig:S_limitation}, it is clear that for real-life manipulated images, the forged regions may have diverse sizes and shapes. In the first row, we show a specific case where the same pattern is copied several times but with different scales. Despite our method fails to localize all forged regions, it is less sensitive to scale variation as compared to ManTra-Net~\cite{wu2019mantra} and SPAN~\cite{hu2020span}, owing to our tailored network design. In addition, our method may fail to localize the whole forged regions or only localize part of them in some cases (\textit{e.g.}, the last two rows). One possible reason is that some manipulation traces are elaborately removed by fabricators. Indeed, the compared IMDL methods also have difficulty to tackle these manipulated images. Note that even in theses cases, our PSCC-Net still performs relatively better than the SOTAs~\cite{wu2019mantra,hu2020span} for image manipulation localization (\textit{e.g.}, see the $2$rd row).

\section{Conclusion}
In this work, a novel PSCC-Net is proposed to meet the challenge of advanced image manipulation techniques.
We employ a progressive mechanism to predict the manipulation mask on all backbone scales, where each mask serves as a prior to help predict the next-scale mask. Moreover, a SCCM is designed to perform spatial and channel-wise attentions on extracted features, which provides holistic information to make our model more generalized to manipulation attacks. Extensive experiments demonstrate that our PSCC-Net outperforms the SOTA methods on both detection and localization. For future work, we will develop techniques for estimating the uncertainty of predicted manipulation masks to further improve the IMDL performance.


%

%
%
%
%
%

\ifCLASSOPTIONcaptionsoff
\newpage
\fi



%
%
%
\bibliographystyle{IEEEtran}
\bibliography{egbib}

%




\begin{IEEEbiography}[{\includegraphics[width=1in,height=1.25in,clip,keepaspectratio]{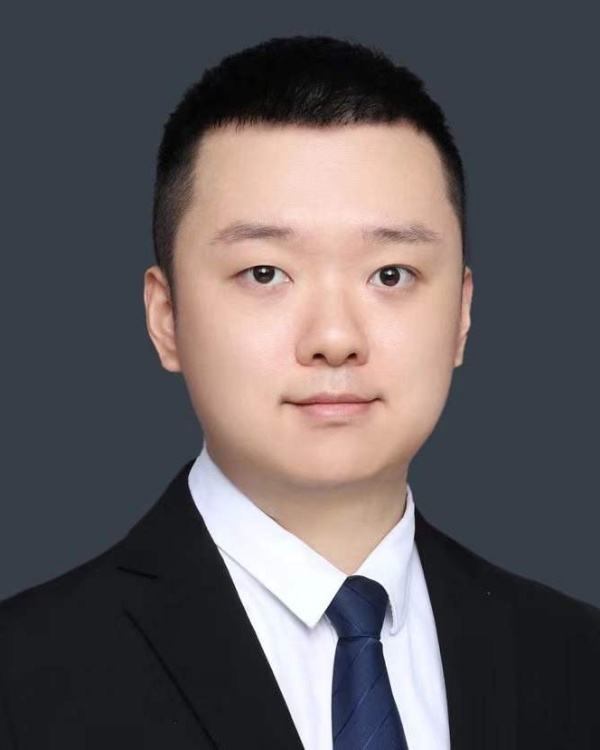}}]{Xiaohong Liu}
	received the  Ph.D. degree in electrical and computer engineering from McMaster University, Hamilton, ON, Canada, in 2021, the M.A.Sc. degree in electrical and computer engineering from University of Ottawa, Ottawa, ON, Canada, in 2016, and the B.E. degree in communication engineering from Southwest Jiaotong University, Chengdu, China, in 2014. He is currently a tenure-track Assistant Professor with John Hopcroft Center, Shanghai Jiao Tong University, Shanghai, China. His research interests include image/video restoration and image segmentation. He was the recipient of the Ontario Graduate Scholarship in 2019, NSERC Alexander Graham Bell Canada Graduate Scholarship-Doctoral and Borealis AI Global Fellowship award in 2020. He is a reviewer of several IEEE journals, including \textsc{IEEE Transactions on Pattern Analysis and Machine Intelligence, IEEE Transactions on Multimedia, IEEE Transactions on Circuits and Systems for Video Technology,} and \textsc{IEEE Transactions on Intelligent Transportation Systems.}
\end{IEEEbiography}


\begin{IEEEbiography}[{\includegraphics[width=1in,height=1.25in,clip,keepaspectratio]{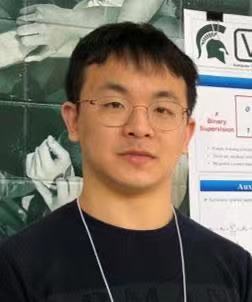}}]{Yaojie Liu}
	is a research scientist at Google Research. He received the Ph.D. degree in Computer Science and Engineering from Michigan State University in 2021. He received the M.S. in Computer Science from the Ohio State
	University in 2016 and the B.S. in Communication Engineering from University of Electronic
	Science and Technology of China in 2014. His research areas of interest are security of face biometric systems (e.g., face anti-spoofing, digital manipulation attack, adversarial attack), 3D face modeling, face representation \& analysis, XAI, image systhesis, and multi-model modeling.
\end{IEEEbiography}

\begin{IEEEbiography}[{\includegraphics[width=1in,height=1.25in,clip,keepaspectratio]{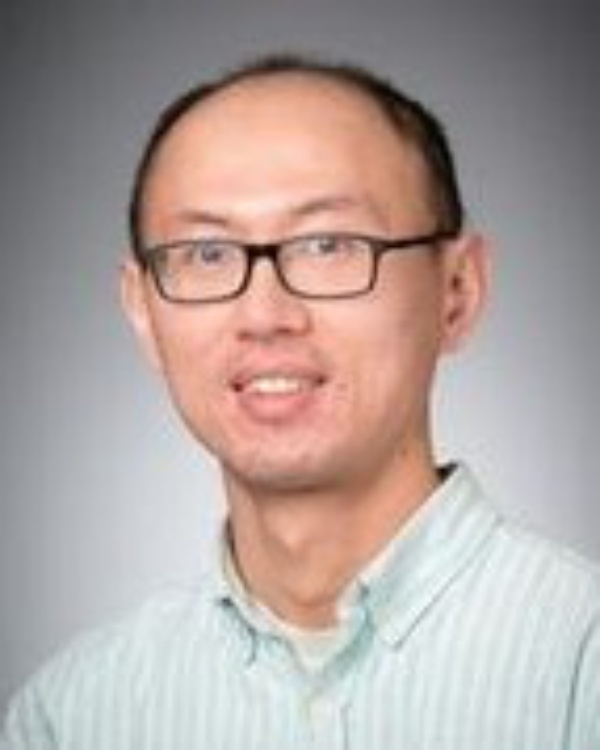}}]{Jun Chen}
(Senior Member, IEEE) received the B.E. degree in communication engineering from Shanghai Jiao Tong University, Shanghai, China, in 2001, and the M.S. and Ph.D. degrees in electrical and computer engineering from Cornell University, Ithaca, NY, USA, in 2004 and 2006, respectively.

From September 2005 to July 2006, he was a Post-Doctoral Research Associate with the Coordinated Science Laboratory, University of Illinois at Urbana–Champaign, Urbana, IL, USA, and a Post-Doctoral Fellow with the IBM Thomas J. Watson Research Center, Yorktown Heights, NY, USA, from July 2006 to August 2007. Since September 2007, he has been with the Department of Electrical and Computer Engineering, McMaster University, Hamilton, ON, Canada, where he is currently a Professor. His research interests include information theory, machine learning, wireless communications, and signal processing.

Dr. Chen was a recipient of the Josef Raviv Memorial Postdoctoral Fellowship in 2006, the Early Researcher Award from the Province of Ontario in 2010, the IBM Faculty Award in 2010, the ICC Best Paper Award in 2020, and the JSPS Invitational Fellowship in 2021. He held the title of the Barber-Gennum Chair of information technology from 2008 to 2013 and the title of the Joseph Ip Distinguished Engineering Fellow from 2016 to 2018. He served as an Editor for \textsc{IEEE Transactions on Green Communications and Networking} from 2020 to 2021. He is currently an Associate Editor of \textsc{IEEE Transactions on Information Theory}.
\end{IEEEbiography}

\begin{IEEEbiography}[{\includegraphics[width=1in,height=1.25in,clip,keepaspectratio]{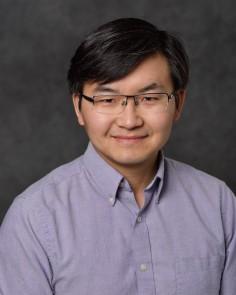}}]{Xiaoming Liu}
	(Senior Member, IEEE) received the PhD degree in electrical and computer engineering from Carnegie Mellon University, Pittsburgh, Pennsylvania, in 2004. He is currently a MSU Foundation Professor with the Department of Computer Science and Engineering, Michigan State University, East Lansing, Michigan. Before joining MSU, in Fall 2012, he was a research scientist at General Electric (GE) Global Research. His research interests include computer vision, machine learning, and biometrics. As a coauthor, he is a recipient of Best Industry Related Paper Award Runner-up at ICPR 2014, Best Student Paper Award at WACV 2012 and 2014, and Best Poster Award at BMVC 2015. He has been the area chair for numerous conferences, including CVPR, ECCV, ICCV, NeurIPS, and ICLR. He is the program chair of WACV 2018, BTAS 2018, AVSS 2022, and general chair of FG 2023. He is an associate editor of the \textsc{Pattern Recognition}, and the \textsc{IEEE Transactions on Image Processing}. He has authored more than 160 scientific publications, and has filed 29 U.S. patents.
\end{IEEEbiography}

\end{document}